%% file: main.tex
\documentclass[12points]{article}

\input{config/packages}

\input{config/header.tex}

\title{Robust Policy Optimization to Prevent Catastrophic Forgetting}
\author[1]{Mahdi Sabbaghi\thanks{Correspondence can be made to: smahdi@seas.upenn.edu}}
\author[1]{George Pappas}
\author[2]{Adel Javanmard}
\author[1]{Hamed Hassani}

\affil[1]{University of Pennsylvania}
\affil[2]{University of Southern California}

\date{}

\begin{document}
\maketitle

\begin{abstract}
Large language models are commonly trained through multi-stage post-training: first via RLHF, then fine-tuned for other downstream objectives. Yet even small downstream updates can compromise earlier learned behaviors (e.g., safety), exposing a brittleness known as catastrophic forgetting. This suggests standard RLHF objectives do not guarantee robustness to future adaptation. To address it, most prior work designs downstream-time methods to preserve previously learned behaviors. We argue that preventing this requires pre-finetuning robustness: the base policy should avoid brittle high-reward solutions whose reward drops sharply under standard fine-tuning.

We propose Fine-tuning Robust Policy Optimization (FRPO), a robust RLHF framework that optimizes reward not only at the current policy, but across a KL-bounded neighborhood of policies reachable by downstream adaptation. The key idea is to ensure reward stability under policy shifts via a max-min formulation.
By modifying GRPO, we develop an algorithm with no extra computation, and empirically show it substantially reduces safety degradation across multiple base models and downstream fine-tuning regimes (SFT and RL) while preserving downstream task performance. We further study a math-focused RL setting, demonstrating that FRPO preserves accuracy under subsequent fine-tuning.
\begin{center}
    \faGithub~\url{https://github.com/Helloworld10011/FRPO}
\end{center}
\end{abstract}

\input{files_arxiv/intro_arxiv}
\input{files_arxiv/related_arxiv}

\section{Preliminaries: RLHF and Fine-tuning}\label{sec:prelim}
We study the scenario where a base model is fine-tuned on a downstream task. 
Let $x\sim p(x)$ denote a prompt and $y_i = (y_{i, 1},\dots,y_{i, |y_i|}) \sim \pi(\cdot\mid x)$ a full generated sample, where $y_{i, t} \sim \pi (\cdot \mid  [x,y_{i, <t}])$. We denote the reference policy with $\pi_{\mathrm{ref}}$ as the model before the RLHF stage. At training time, we aim to optimize the policy $\pi_\theta$ as the base policy we want to be robust to downstream fine-tuning. At the downstream stage, we wish to optimize a policy $Q$ initialized from $\pi_\theta$. Therefore, the chain of models is: $\pi_{\mathrm{ref}} \to \pi_\theta \to Q$. 

\vspace{-0.1in}

\paragraph{RLHF objective and policy optimization.}
Following standard RLHF and given a reward signal, we define the trajectory return $r(x,y)=\sum_{t=1}^{|y_i|} r_t(x,y_{\le t})$ (often implemented as an outcome reward, i.e., one reward for the entire response). The RLHF objective then maximizes \citep{ouyang2022training}: 
\[
\max_\theta\;\; \E_{x\sim p}\E_{y\sim \pi_\theta(\cdot|x)}\!\big[ r(x,y)\big]\;-\;\beta\,\E_{x\sim p}\!\big[\KL(\pi_\theta(\cdot|x)\,\|\,\pi_{\mathrm{ref}}(\cdot|x))\big]
\]
In practice, this objective is optimized with PPO-style policy gradients using on-policy samples from a lagged policy $\pi_{\mathrm{old}}$ \citep{schulman2017proximal}; we adopt GRPO \citep{shao2024deepseekmath} as an efficient algorithm for optimizing this objective, which we summarize next. 

\vspace{-0.1in}
\paragraph{GRPO for policy optimization.}
For each prompt $x$, GRPO samples a group of responses ${y_i}\big|_{i=1}^G \sim \pi_{\mathrm{old}}(\cdot\mid x)$. It then updates the trainable policy $\pi_\theta$ using a PPO-style clipped importance ratio. Without a learned critic, advantages are obtained from within-group reward statistics, and a KL penalty to a fixed reference policy is added directly to the loss:
\begin{align}\label{eq:GRPO_main}
J(\theta)=& \E_{x\sim p}\!\Bigg[\left(\frac{1}{G}\sum_{i=1}^G \frac{1}{|y_i|}\sum_{t=1}^{|y_i|} \min\left\{\frac{\pi_{\theta}(y_{i,t}|x,y_{i,<t})}{\pi_{{\mathrm{old}}}(y_{i,t}|x,y_{i,<t})} A_{i, t}, \left[\frac{\pi_{\theta}(y_{i,t}|x,y_{i,<t})}{\pi_{{\mathrm{old}}}(y_{i,t}|x,y_{i,<t})}\right]_{1-\eps}^{1+\eps} A_{i, t} \right\}\right) \Bigg] \nonumber \\
& \hspace{0.65 \textwidth} -\beta \KL(\pi_\theta \| \pi_{\mathrm{ref}})
\end{align}
where $A_{i,t} \equiv A_i = r_i - \frac{1}{G}\sum_{j=1}^G r_j$ as we only work with an outcome-based reward. Unlike the approach of \citep{shao2024deepseekmath}, we do not normalize the advantages by the standard deviation to avoid the difficulty bias reported in \citep{liu2025understanding}. The KL divergence is estimated with the following approximately unbiased estimator \citep{ouyang2022training}:
\begin{equation}\label{eq:KL-approx}
    KL(\pi_\theta(\cdot \mid x)\,\|\,\pi_{\mathrm{ref}}(\cdot \mid x)) \approx \frac{1}{G}\sum_{i=1}^G \frac{1}{|y_i|}\sum_{t=1}^{|y_i|} \left(\frac{\pi_{\mathrm{ref}}(y_{i,t}|x, y_{i,<t})}{\pi_{\theta}(y_{i,t}|x, y_{i,<t})}- \log\frac{\pi_{\mathrm{ref}}(y_{i,t}|x, y_{i,<t})}{\pi_{\theta}(y_{i,t}|x, y_{i,<t})} -1 \right)
\end{equation}
\paragraph{KL-bounded is bounded during fine-tuning.}
In standard downstream adaptation, the fine-tuned policy $Q$ does not move arbitrarily far from its initialization $\pi_\theta$. In RL-based methods, this is enforced explicitly via trust-region updates \citep{schulman2015trust} or the KL regularizer (the RLHF penalty to a reference policy) \citep{schulman2017proximal,ouyang2022training}. In SFT, this is controlled implicitly through conservative optimization choices like small learning rates and early stopping \citep{mosbach2020stability,dodge2020fine}, or parameter-efficient updates such as LoRA \citep{hu2022lora}. 

\begin{figure}
    \centering
    \begin{subfigure}[b]{0.4\textwidth}
        \includegraphics[width=\linewidth]{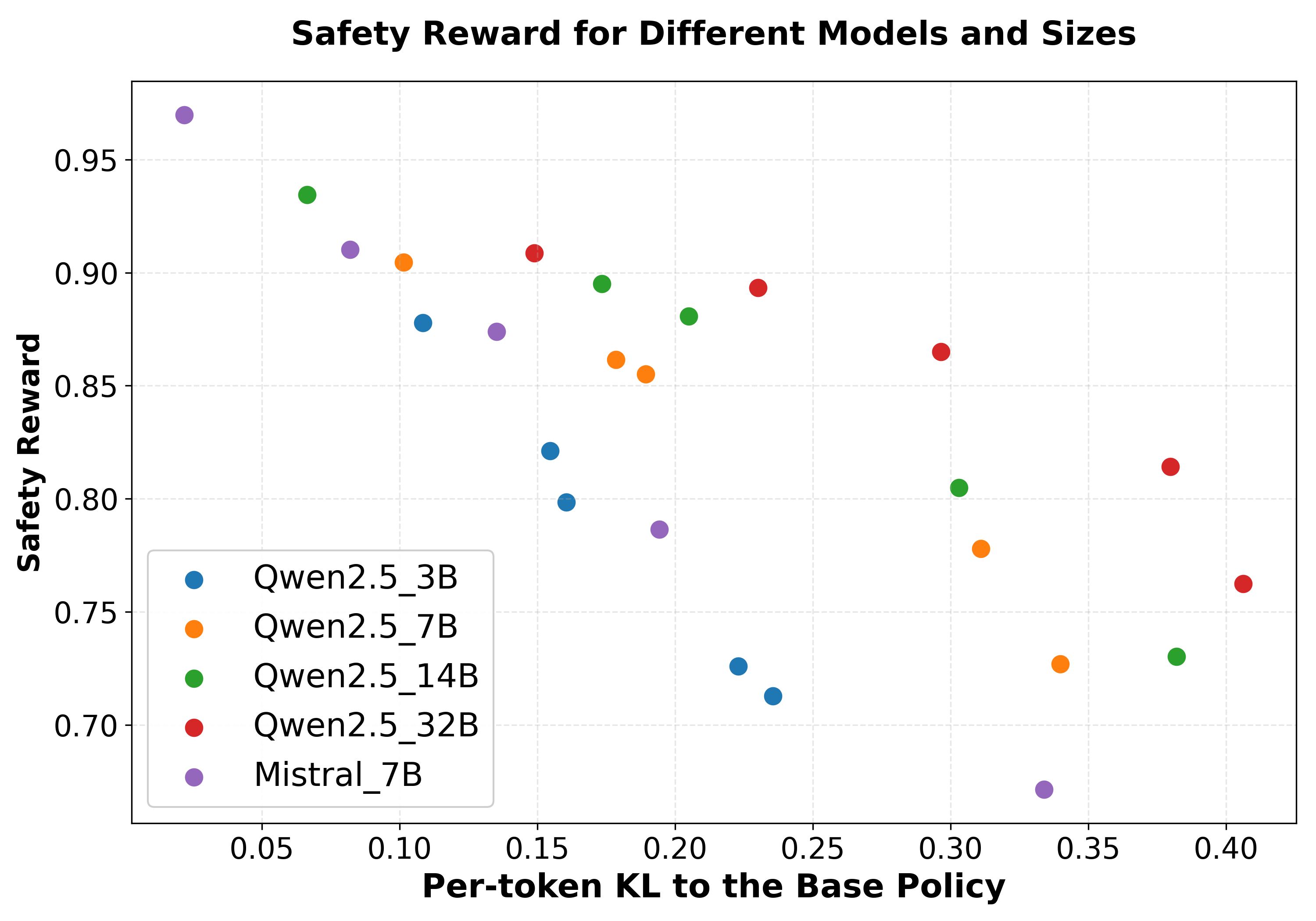}
    \end{subfigure}
    \hspace{0.02in}
    \begin{subfigure}{0.4\linewidth}
        \includegraphics[width=\linewidth]{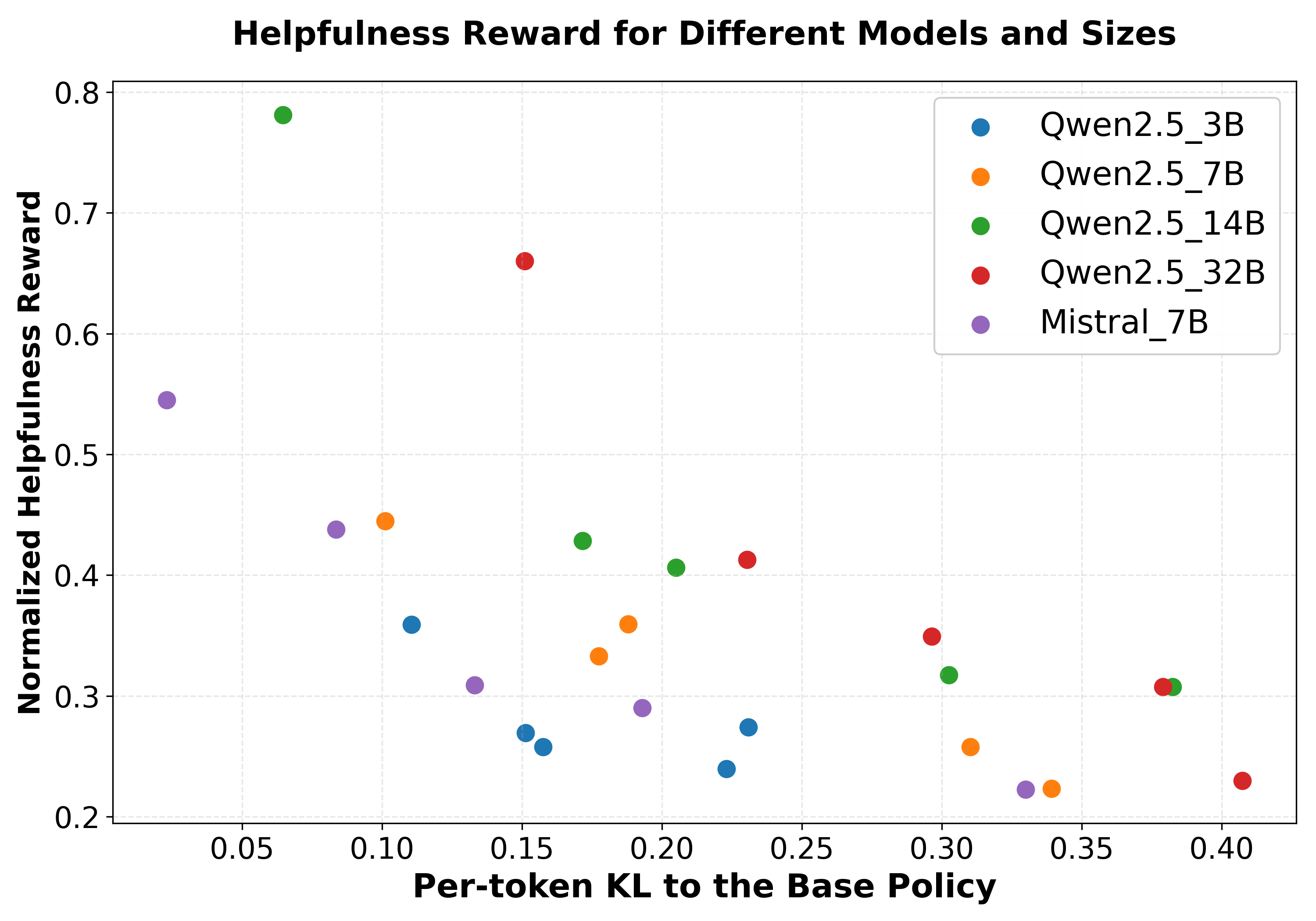}
    \end{subfigure}
    \vspace{-2pt}
     \caption{Safety rewards \textbf{(left)} and normalized helpfulness rewards \textbf{(right)} for models of different sizes trained with GRPO for safety and then fine-tuned on GSM8K. Helpfulness is normalized by each model's initial reward. Both rewards decrease as the per-token KL from the base model grows. Around per-token KL $\approx 0.3$, helpfulness drops consistently across models and overfitting begins; we therefore use this value as the main KL budget in \Cref{sec:experiments}.}
     \label{fig:reward_vs_KL}
     \vspace{-8pt}
\end{figure}

We further validate this without any regularizer in \Cref{fig:reward_vs_KL}. We first train Qwen2.5-3B, 7B, 14B, 32B, and Mistral with GRPO and with our safety training recipe explained in \Cref{sec:exp-safety}. We then fine-tune these models on GSM8k \citep{cobbe2021training} across a wide range of hyper-parameters, from less to more aggressive. We do this with and without LoRA, $\mathrm{lr \in \{1e-6, 6e-6, 6e-5\}}$, and for 1 or 2 epochs. We plot the safety and helpfulness rewards after finetuning against the KL divergence to their base model (before fine-tuning). \Cref{fig:reward_vs_KL} shows that both of these rewards decrease as the KL grows, but we can see that larger models are less susceptible to forgetting and remain higher \citep{ramasesh2021effect,liu2025conditions}. Most of the training settings fall within the KL ball: per-token-KL $\leq0.3$. More importantly, as the right figure shows, around per-token-KL=0.3, that helpfulness reward for all the models nearly collapses and they approach overfitting. We conclude that the same KL-ball describes the fine-tuning dynamics before overfitting across different models and sizes. 
Motivated by this, we model downstream adaptation as a policy $Q$ that remains within a KL neighborhood of $\pi_\theta$.

\section{Problem Formulation: Robust RLHF}\label{sec:formulation}
Our primary objective is to train a policy $\pi_\theta$ that preserves its expected reward across all fine-tuned variants $Q$ lying within a specified distance of $\pi_\theta$.
More concretely, let $p(x)$ be a distribution over contexts, let $\pi_\theta(y\mid x)$ denote the trainable base policy, let $\pi_{\mathrm{ref}}(y\mid x)$ be a fixed reference policy (e.g., the pre-RLHF model), and let $r(x,y)$ be a scalar reward.
We consider any arbitrary $Q(y | x)$ that is an adaptation of $\pi_\theta(\cdot | x)$  in downstream fine-tuning, but that must stay within an average
$\KL$ ball of radius $\rho>0$:
\begin{align}\label{eq:primal}
\max_{\pi_\theta} & \;
\underbrace{\inf_{Q}\; \E_{x\sim p}\E_{y\sim Q(\cdot\mid x)}[\,r(x,y)\,]}_{\text{robust reward}}
-\beta\,\KL\!\big(\pi_\theta \| \pi_{\mathrm{ref}}\big)\\
\text{s.t. } \ &\ \ 
\E_{x\sim p}\Big[\KL\!\big(Q(\cdot\mid x)\,\big\|\,\pi_\theta(\cdot\mid x)\big)\Big]\ \le\ \rho,
\qquad 
\forall x: \int Q(\mathrm{d}y\mid x)=1. \nonumber
\end{align}
Crucially, the constraint is imposed  in expectation over the prompt distribution $p(x)$, rather than point-wise for each $x$. This allows the downstream policy to change significantly on task-relevant prompts (where fine-tuning occurs), but requires it to stay close on average. 
We use the following lemma from \citep{shapiro2017,duchi2021learning} modified to handle the additional expectation over $x$. The proof can be found in \Cref{app:lemma_proof}.

\begin{lemma}[Inner optimization under a general $f$–divergence]\label{lem:optQ_fdiv}

Let $f:\R \!\to\R_+ \cup\{+\infty\}$ be a convex function with $f(1)=0$. Define the likelihood ratio $L(y\mid x):=\frac{Q(y\mid x)}{\pi(y\mid x)}$. Then, the inner problem in \Cref{eq:primal} under the average constraint:
\[
\E_{x\sim p}\E_{y\sim\pi(\cdot\mid x)}[\,f(L(y\mid x))\,]\ \le\ \rho,
\qquad
\E_{y\sim\pi(\cdot\mid x)}[\,L(y\mid x)\,]=1\ \ \forall x
\]
admits the following dual form:
\begin{align}
\inf_{Q}\ \E_{x\sim p}\E_{y\sim Q(\cdot\mid x)}[\,r(x,y)\,]
=
\sup_{\lambda\ge 0,\ \eta:\calX\to\R}
\Big\{ & -\lambda\rho - \E_{x\sim p}\eta(x) \nonumber \\
& - \lambda\,
\E_{x\sim p}\E_{y\sim\pi(\cdot\mid x)}
\!\Big[f^*\!\Big(\frac{-\,r(x,y)-\eta(x)}{\lambda}\Big)\Big]\Big\},
\label{eq:general-dual}
\end{align}
where $f^*(s):=\sup_{t\ge 0}\{st-f(t)\}$ is the Fenchel conjugate. In addition, if the supremum is finite, it is attained at some $(\lambda^*,\eta^*)$. 
\end{lemma}

\noindent In our setting, we take $f(x) = x \log (x)$ that leads to the forward KL divergence between $Q$ and $\pi_\theta$. Then, it is easy to see that the conjugate is $f^*(y) = \exp(y-1)$, and the corresponding optimal likelihood ratio is $L(y | x) = \frac {\exp(-r(x, y) /\lambda)} {\E_{y\sim\pi_\theta(\cdot\mid x)}\!\left[e^{-\,r(x,y)/\lambda}\right]}$. The optimization problem thus becomes:

\begin{equation*}
    \sup_{\lambda,\ \eta} \Big\{ \E_{x\sim p}\E_{y\sim\pi_\theta(\cdot\mid x)}
\Big[ -\lambda \,\exp \big(-\frac{1}{\lambda} (r(x,y) + \eta(x) + \lambda) \big) - \eta(x) \Big] - \lambda\rho \Big\} 
\end{equation*}

\noindent For each $x$, \Cref{eq:general-dual} can be optimized w.r.t.\ $\eta(x)$ separately. After plugging in, we obtain:
\begin{align}\label{eq:all_forms}
\min_L \max_{\lambda \geq0, \eta} \calL = \max_{\lambda \geq 0} \max_{\eta(x)} \min_{L}\calL(L,\lambda,\eta)
= & \max_{\lambda \geq 0} \big \{ -\,\E_{x\sim p}\!\big[\lambda\log Z(x)\big]-\lambda\rho \big \}, \\ & Z(x):= \E_{y\sim\pi_\theta(\cdot\mid x)}\!\left[e^{-\,r(x,y)/\lambda}\right] \nonumber
\end{align}


\noindent Combining with \Cref{eq:primal}, the problem reduces to a joint max–max program, where we can always swap the order of the two maxes. We obtain the following:
\begin{align}\label{eq:with_rho}
\ \max_{\lambda\ge 0}\max_{\pi_\theta}
\left\{
-\E_{x\sim p}\!\left[\lambda\log\Big(\E_{y \sim \pi_\theta(\cdot\mid x)}e^{-\,r(x,y)/\lambda}\Big)\right]
-\lambda\rho
-\beta\,\KL\!\big(\pi_\theta \|\pi_{\mathrm{ref}}\big)
\right\}
\end{align}

\noindent Eventually, we view $\lambda$ as a tunable hyperparameter in the policy optimization. Therefore, the final optimization is: 
\begin{align}
\boxed{\max_{\pi_\theta}
\left\{
\underbrace{
-\E_{x\sim p}\!\left[\lambda\log\Big(\E_{y \sim \pi_\theta(\cdot\mid x)}e^{-\,r(x,y)/\lambda}\Big)\right]
-\beta\,\KL\!\big(\pi_\theta \|\pi_{\mathrm{ref}}\big)}_{J_\lambda(\theta)}
\right\}}
\label{eq:final_opt}
\end{align}

 \begin{remark}\label{remark:lambda-rho}
 If $\rho=0$ (unperturbed objective), then the optimal $\lambda$ is $\lambda^*=\infty$, which recovers GRPO and the term $\E[r(x,y)]$ in the objective as $\lambda\to\infty$. To see this, we start from $\rho = 0$ and write \eqref{eq:all_forms}  as 
 $-\min_{\lambda \geq 0} [\E_{x\sim p}\!\big[\lambda\log Z(x)\big] + \lambda \rho]$. Now, by the Jensen's inequality and the concavity of the log function, we have $\log Z(x)\ge \E_{y\sim\pi_\theta(\cdot|x)}[-r(x,y)/\lambda]$, and also $\E_{x\sim p}\!\big[\lambda\log Z(x)\big] \ge \E[-r(x,y)]$. Furthermore, by taking $\lambda\to\infty$ we can achieve this minimum. Therefore, the optimal value is $\lambda^*=\infty$.
\end{remark}

\begin{remark} \label{remark:variance}
The max-min formulation yields the \emph{entropic risk}
$-\lambda\log \E_{\pi_\theta}e^{-r/\lambda}$, where smaller $\lambda$ yields a more risk-averse objective.
For large $\lambda$, by a Taylor expansion in $1/\lambda$ and keeping $O(1/\lambda)$, the first two terms in \eqref{eq:with_rho} become:
\begin{align*}
&\max_{\lambda\ge 0}\max_{\pi_\theta}
\left\{
\E[r(x,y)] - \frac{1}{2\lambda}{{\rm Var}}(r(x,y))
-\lambda\rho
\right\}  = \max_{\pi_\theta}
\left\{ \E[r(x,y)] - \sqrt{2\rho {\rm Var}(r(x,y))} \right\}
\end{align*}
The right side is optimized at $\lambda = \frac{\mathrm{std}(r(x,y))}{\sqrt{2\rho}}$. This shows a trade-off between mean reward and variance in the objective, favoring more consistent policies.
\end{remark}

\section{FRPO Algorithm} \label{sec:algorithm}
{
\begin{algorithm}[t]
\caption{Fine-tuning Robust Policy Optimization (FRPO)}\label{alg:frpo}
\begin{algorithmic}[1]
\State Initialize policy parameters $\theta$.
\For{iteration $k=1,2,\ldots$}
    \State $\pi_{\text{old}} \leftarrow \pi_\theta$ \Comment{lagged sampling policy; updated every iteration (or every $K$ iterations)}
    \ForAll{$x$ in minibatch}
        \State Sample $\{y_i\}_{i=1}^G \sim \pi_{\text{old}}(\cdot\mid x)$
        \State $r_i \leftarrow r(x,y_i)$;\;\; $A_i \leftarrow r_i - \frac{1}{G}\sum_{j=1}^G r_j$ \Comment{centered group advantages}
        \State Compute $\hat Z_\lambda(x)$ using clipped ratios $\frac{\pi_\theta}{\pi_{\text{old}}}$ (Eq.~\ref{eq:algorithm_GRPO}) 
        \Comment{partition function}
        \State Compute leave-one-out $\hat Z_{\lambda,-j}(x)$ for $j=1,\ldots,G$
        \State $\log \tilde Z_\lambda(x) \leftarrow
        G \log \hat Z_\lambda(x) - \frac{G-1}{G}\sum_{j=1}^G \log \hat Z_{\lambda,-j}(x)$
        \Comment{jackknife: reduces $O(1/G)$ bias}
    \EndFor
    \State $J(\theta) \leftarrow \mathbb{E}_x\Big[-\lambda \log \tilde Z_\lambda(x) - \beta \KL(\pi_\theta\|\pi_{\text{ref}})\Big]
    + \text{baseline (Eq.~\ref{eq:low_var})}$
    \Comment{add the baseline and the KL}
    \State Update $\theta \leftarrow \theta + \eta \nabla_\theta J(\theta)$.
\EndFor
\end{algorithmic}
\end{algorithm}
\noindent
}
We derived the robust objective $J_\lambda(\theta)$ in \Cref{eq:final_opt}. Following the discussion in \Cref{sec:prelim}, we focus on the outcome-based setting where $r_{i, t} = r_i = r(x, y_i)$, and use the centered advantages: $A_i = r_i - \frac{1}{G}\sum_{j=1}^G r_j$. We now explain how to optimize this objective using a GRPO-style policy gradient. 
First, we rewrite the inner expectation under $\pi_{\mathrm{old}}$ using importance sampling, as samples are drawn from the lagged policy. The objective becomes: 

\begin{align*}
J_\lambda (\theta) = \E_{x\sim p}\!\left[-\lambda\log\Big(\E_{y\sim\pi_{{\mathrm{old}}}(\cdot\mid x)}\frac{\pi_{\theta}(y|x)}{\pi_{\mathrm{old}}(y|x)}e^{-\,r(x,y)/\lambda}\Big)\right]
-\beta\,\KL\!\big(\pi_\theta\|\pi_{\mathrm{ref}}\big)
\end{align*}
We replace rewards with advantages in the equation above; we show in \Cref{app:baseline} that subtracting the group average does not affect the gradient. Using token-wise ratios and clipping in \Cref{eq:GRPO_main}, the Monte Carlo estimate of the objective becomes:
\begin{align}\label{eq:algorithm_GRPO}
& J_\lambda(\theta) = \E_{x\sim p}\!\Big[-\lambda\log\big(\hat Z_\lambda(x) \big) 
                   - \KL(\pi_\theta \,\|\,\pi_{\mathrm{ref}})\Big], \\
&\hat Z_\lambda(x) := \frac{1}{G}\sum_{i=1}^G \frac{1}{|y_i|}\sum_{t=1}^{|y_i|} 
                     e^{-A_{i,t}/\lambda}
                     \left\lceil \frac{\pi_{\theta}(y_{i,t}|x,y_{i,<t})}
                     {\pi_{\mathrm{old}}(y_{i,t}|x,y_{i,<t})}\right\rceil^{1+\eps} \nonumber
\end{align}
Here we use that $e^{A / \lambda} \geq 0$, and $\min\{u, u\big|_{1-\eps}^{1+\eps}\} = u\big|^{1+\eps}$ for $u \geq 0$, and KL is plugged in from \Cref{eq:KL-approx}. So if $A$ is highly negative the clipping avoids instability, while bad actions  are freely pushed toward zero probability because the lower bound  $1 - \epsilon$ drops out. 

\vspace{-0.1in}
\paragraph{Baseline.} In \Cref{app:baseline}, we show that when $\lambda$ is large, the gradient of the log-partition function estimate $\log(Z)$  has a high variance, leading to convergence issues as discussed in several prior works \citep{chung2021beyond,mei2022role}.
We address this by adding a baseline to the objective that completely cancels the leading drift term in the gradient for large $\lambda$, but does not change the expected gradient. This extra term makes our algorithm converge to GRPO as $\lambda \to \infty$ (see \Cref{remark:lambda-rho}).

\begin{equation}
\begin{split}
    J_{\text{low-variance}}(\theta) ={} J(\theta) +
     \underbrace{ \lambda \; \mathbb{E}_{x\sim p} \left( \frac{1}{G}\sum_{i=1}^G \frac{1}{|y_i|}\sum_{t=1}^{|y_i|}
    \left[ \frac{\pi_{\theta}(y_{i,t}|x,y_{i,<t})}{\pi_{\text{old}}(y_{i,t}|x,y_{i,<t})}\right]^{1+\epsilon} \right)}_{\text{baseline}}
\end{split}
\label{eq:low_var}
\end{equation}
The term inside the sum is independent of the advantages and only adds a constant to the objective in expectation. 

\vspace{-0.1in}
\paragraph{Bias reduction.} In \Cref{app:jackknife},
we show that the Monte Carlo approximation of $\log Z(x)$ introduces a bias of order $O(1/G)$ in $\nabla_\theta J_\lambda (\theta)$. This bias is problematic when the group size $G$ is small (the standard group size is typically between 4 and 16). Then, we show that the jackknife technique reduces this bias to $O(1/G^2)$.
Skipping the KL regularizer and the baseline term for readability, we write the corrected objective as:
\begin{align}\label{eq:jackknife_obj}
J_{\text{corrected}}(\theta)
= \E_{x\sim p}\!\Bigg[
&-\lambda G \, \log\!\left(\hat Z_\lambda(x) \right)
+ \lambda \frac{G-1}{G}\sum_{j=1}^G \log\!\left(\hat Z_{\lambda, -j}(x)\right)
\Bigg]
\end{align}
where $\hat Z_{\lambda, -j}(x)$ is the leave-one-out estimate with index $j$ removed.
The full algorithm with the baseline and the jackknife correction is summarized in \Cref{alg:frpo}.

\input{files_arxiv/exp_arxiv2}
\input{files_arxiv/conclusion_arxiv}

\section*{Acknowledgment} 
This research has been supported by Coefficient Giving and the UK AI Security Institute. AJ was supported in part by  the Sloan fellowship in mathematics, the NSF Award DMS-2311024, an Amazon
Faculty Research Award, an Adobe Faculty Research Award and an iORB grant form USC Marshall School of Business.

\newpage

\bibliography{bibliography}
\bibliographystyle{plainnat}

\newpage
\appendix

\input{files_arxiv/appendix_arxiv}

\end{document}

%% file: config/packages.tex
\usepackage[utf8]{inputenc}
\usepackage[T1]{fontenc}
\usepackage[hyphens]{url}
\usepackage{authblk}
\makeatletter
\renewcommand\AB@affilsepx{\quad\protect\Affilfont}
\makeatother
\renewcommand\Affilfont{\fontsize{10}{13}\selectfont}
\usepackage{booktabs}
\usepackage[
    colorlinks=true,
    linkcolor={red},
    citecolor={blue!50!black},
    urlcolor={blue!70!black}
]{hyperref}
\usepackage{amsfonts}
\usepackage{pifont}
\usepackage{nicefrac}
\usepackage{microtype}
\usepackage{amsmath, amsthm, amssymb}
\usepackage{cleveref}
\usepackage{tcolorbox}
\usepackage{fancybox}
\usepackage{graphicx}
\usepackage{listings} 
\usepackage{float}
\usepackage{algorithm}
\usepackage[noend]{algpseudocode}
\usepackage{color}
\usepackage{tikz}
\usepackage{pgfplots}
\usepackage{array}
\usepackage{caption}
\usepackage{subcaption}
\usepackage[textsize=scriptsize]{todonotes}
\usepackage{enumitem}
\usepackage{wrapfig}
\usepackage{etoolbox}
\usepackage{diagbox}
\usepackage{comment}
\usepackage{makecell}
\usepackage{multirow}
\usepackage{bbm, bm}
\usepackage{multicol}
\usepackage{mathpazo}
\usepackage{array, booktabs}
\usepackage{arydshln}
\usepackage{hyperref}
\usepackage{subcaption}
\usepackage{siunitx}
\usepackage{times}
\usepackage{setspace}
\usepackage[normalem]{ulem}
\usepackage[round]{natbib}
\usepackage{fontawesome5}
\usepackage[
  textwidth=6.0in,
  textheight=9.0in,
  left=1.25in,
  top=1in
]{geometry}

\setlist[itemize]{leftmargin=2.5em}
\setlist[itemize]{leftmargin=2.5em}
\setlist[enumerate]{leftmargin=2.5em}
\usepackage[framemethod=TikZ]{mdframed}
\mdfsetup{skipabove=5pt,
innertopmargin=0pt}

%% file: config/header.tex
\newtheoremstyle{slplain}
  {.4\baselineskip\@plus.1\baselineskip\@minus.1\baselineskip}
  {.3\baselineskip\@plus.1\baselineskip\@minus.1\baselineskip}
  {\itshape}
  {}
  {\bfseries}
  {.}
  { }
  {}
\theoremstyle{slplain} 

\newtheorem*{definition*}{Definition}
\newtheorem*{theorem*}{Theorem}
\newtheorem{theorem}{Theorem}[section]
\newtheorem{lemma}[theorem]{Lemma}

\makeatletter
\newtheorem*{rep@theorem}{\rep@title}
\newcommand{\newreptheorem}[2]{%
\newenvironment{rep#1}[1]{%
 \def\rep@title{#2 \ref{##1}}%
 \begin{rep@theorem}}%
 {\end{rep@theorem}}}
\makeatother

\newreptheorem{theorem}{Theorem}
\newreptheorem{lemma}{Lemma}
\newreptheorem{claim}{Claim}
\newreptheorem{corollary}{Corollary}
\newreptheorem{proposition}{Proposition}

\theoremstyle{definition}

\newtheorem{remark}[theorem]{Remark}

\theoremstyle{plain} 

\numberwithin{equation}{section}

\newcommand{\R}{\mathbb{R}}

\DeclareMathOperator{\E}{\mathbb{E}}
\DeclareMathOperator{\KL}{\mathrm{KL}}

\renewcommand\bar\overline

\newcolumntype{C}[1]{>{\centering\let\newline\\\arraybackslash\hspace{0pt}}m{#1}}

\newcommand{\calL}{\ensuremath{\mathcal{L}}}

\newcommand{\calX}{\ensuremath{\mathcal{X}}}

\def\nd/{\textsuperscript{nd}}
\def\rd/{\textsuperscript{rd}}
\def\th/{\textsuperscript{th}}

\newcommand{\eps}{\ensuremath{\epsilon}}

\makeatletter
\def\nnil{\nil}
\newcounter{prob}

\newcounter{dual}

\makeatother

\newenvironment{prob*}{%
	\csname equation*\endcsname%
	\aligned%
}{%
	\endaligned%
	\csname endequation*\endcsname%
}

%% file: files_arxiv/intro_arxiv.tex
\vspace{-10pt}
\section{Introduction}
\vspace{-4pt}

Large language models (LLMs) are becoming the driving force of agentic systems, ranging from everyday chat and mathematical reasoning to robotics~\citep{ahn2022can,schick2023toolformer,driess2023palm,achiam2023gpt}. Such broad deployment requires training a single base model capable of supporting diverse tasks, and, in some cases, fine-tuning for specific downstream applications.
However, optimizing for a downstream objective can compromise other capabilities
developed during earlier training 
\citep{de2019episodic,sun2019lamol}. 
Addressing this trade-off is the core challenge of \textit{continual learning}: training models on new tasks without losing prior knowledge  \citep{de2021continual,wang2022learning,wang2022dualprompt}.

\begin{figure*}[t]
    \centering
    \includegraphics[width=1.0\linewidth]{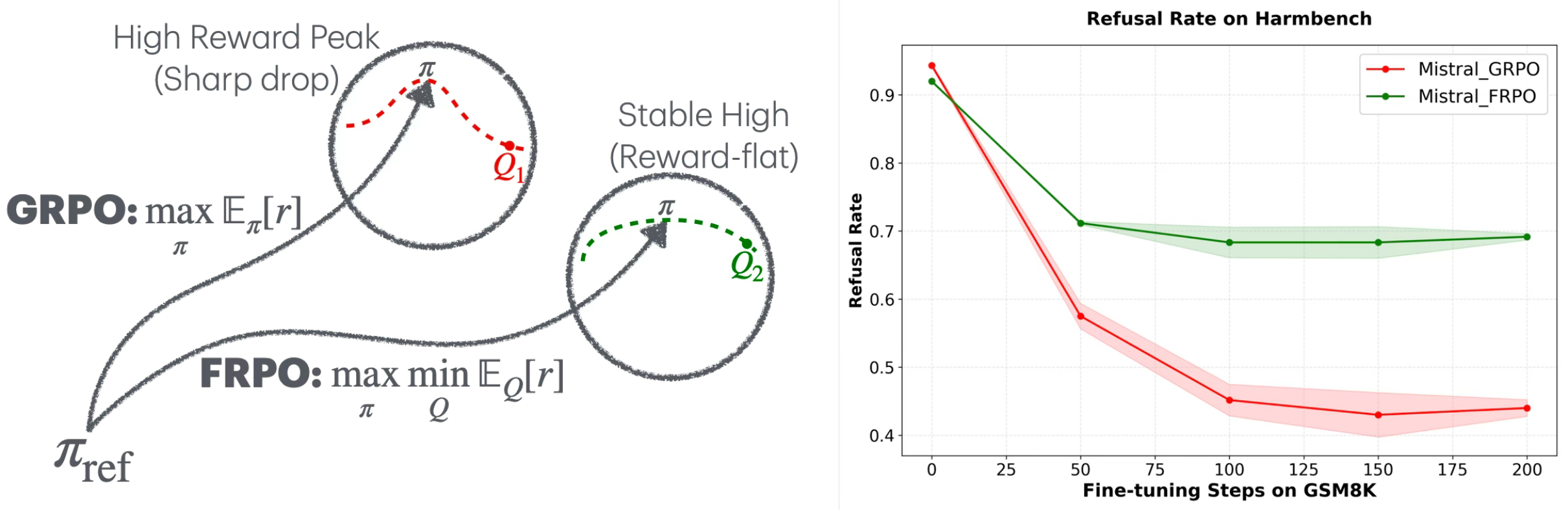}
    \vspace{-6pt}
    \caption{Illustration of FRPO. Standard RLHF finds high-reward policies that may lie in sharp regions, whereas our method optimizes for reward-flatness within a KL neighborhood, finding policies that maintain high reward after downstream adaptation.}
    \label{fig:intro}
    \vspace{-12pt}
\end{figure*}

We study a two-stage continual learning setting in LLM pipelines: (1) the model acquires a behavior such as safety guardrails or a specific capability; and then (2) it is adapted to a downstream task while maintaining the earlier behavior. Recent studies \citep{qi2023fine,zhan2023removing,qi2024safety} show that this often fails: downstream fine-tuning can degrade previously learned behaviors, a phenomenon known as \textit{catastrophic forgetting}. 
\Cref{fig:intro}(right) illustrates this failure mode: fine-tuning on a math task such as GSM8K inadvertently leads to ``forgetting'' the safety guardrails. 
To mitigate forgetting, many studies focus on developing methods at downstream time to preserve the model’s capabilities. 
Rehearsal methods augment downstream training with a subset of previous data 
\citep{rolnick2019experience,sun2019lamol,scialom2022fine,huang2024mitigating}. Parameter-efficient fine-tuning (PEFT) constrains downstream updates into separate modules (e.g., LoRA), thereby reducing parameter interference between the objectives \citep{hu2022lora, hsu2024safe, wang2024learn}.
Regularization strategies constrain  optimization through penalties that prevent the policy from drifting excessively \citep{li2017learning,schulman2017proximal,lee2019mixout, kirkpatrick2017overcoming}.
Finally, 
model merging methods aim to 
combine task-specific models post-hoc to retain all capabilities without expensive retraining \citep{ilharco2022editing, wortsman2022model, yi2024safety, djuhera2025safemerge}.

Downstream-time methods are effective only when used within their intended procedures. Consequently, any later ``ordinary'' fine-tuning that deviates from these procedures can still compromise previously learned behaviors. Yet, much of the literature continues to rely on standard approaches---supervised fine-tuning (SFT) or RL-based methods such as GRPO~\citep{ouyang2022training,shao2024deepseekmath}---which focus on optimizing the immediate downstream objective. However, the central issue remains: \emph{``The base model must be made robust to future fine-tuning, regardless of the downstream task or algorithm.''}

To this end, we propose \textbf{F}ine-tuning \textbf{R}obust \textbf{P}olicy \textbf{O}ptimization (FRPO), a new algorithm aimed at making the policy robust to downstream fine-tuning. Since downstream adaptations typically remain within a neighborhood of the current policy (e.g., bounded in KL divergence), achieving robustness requires avoiding high-reward solutions that reside in sharp regions of the policy space, where even small updates can lead to substantial reward degradation. Rather than maximizing reward solely at the current policy, our objective explicitly considers a neighborhood around the policy and seeks flatter regions, leading to policy stability. 

Formally, to account for future downstream adaptations, we consider the set of all policies reachable within a KL-bounded neighborhood around the current policy as illustrated in \Cref{fig:intro}.
 This formulation is \textit{agnostic} to the downstream fine-tuning, assuming only that the policy remains within this KL neighborhood---reflecting standard practices like KL regularization in RLHF \citep{ouyang2022training} and implicit constraints from limited learning rates or LoRA.
We then propose an objective that maximizes the minimum reward over this set. By deriving the dual form, we show that this robustness criterion is equivalent to penalizing low-reward rollouts under the current policy. This leads to an entropic-risk objective that emphasizes low-reward trajectories and discourages policies with high reward variance. Finally, we derive FRPO to optimize this objective within the standard GRPO framework, with no additional computation.

\textbf{Our contributions} are summarized as follows:
\begin{itemize}[leftmargin=0.7em]
    \item \textbf{RLHF framework for fine-tuning robustness.} 
    We consider a max-min optimization that accounts for all downstream shifts within a KL ball centered at the base policy. We show that this formulation is equivalent to optimizing an entropic-risk   objective,
    with a tunable parameter $\lambda$ controlling sensitivity to low-reward trajectories.

    \item \textbf{FRPO with no extra computation.}
    We present FRPO, a policy gradient method that integrates seamlessly with the GRPO framework. FRPO derives from a closed-form solution to the max-min problem, includes a baseline for stable optimization, and recovers GRPO as $\lambda {\scriptstyle \to} \infty$.

    \item \textbf{Experimental results.}
    We evaluate several fine-tuning schemes, including instruction-following and math (see \Cref{fig:alpaca,fig:GSM8K_lim}) to demonstrate that FRPO-trained models are significantly more effective at preserving safety guardrails and prior capabilities compared to other methods. 
    Lastly, we fine-tune math-trained models on code generation to show that FRPO maintains 22\% higher accuracy on MATH than GRPO.  
\end{itemize}

%% file: files_arxiv/related_arxiv.tex
\subsection{Related Work}
\paragraph{RLHF.}
RLHF optimizes a KL-regularized objective to align LLMs \citep{christiano2017deep,ouyang2022training,bai2022training,stiennon2020learning}. PPO \citep{schulman2017proximal} is the standard choice but requires a learned critic, whereas GRPO and RLOO \citep{shao2024deepseekmath, ahmadian2024back} remove this requirement via group-based advantages. DPO \citep{rafailov2023direct} 
bypasses reward modeling entirely. We modify GRPO to incorporate robustness while preserving its computational simplicity.

\vspace{-0.1in}
\paragraph{Flat minima and sharpness-aware optimization.}
Flat minima in the loss landscape correlate with better generalization \citep{keskar2016large,jiang2019fantastic,cha2021swad}. This motivates sharpness-aware minimization (SAM) \citep{foret2020sharpness} and variants \citep{kwon2021asam,zhuang2022surrogate,kim2022fisher}. SAM considers a min-max loss in the parameter space, seeking flat regions. Our method instead operates in \emph{policy space}: we seek ``reward flatness'' over a KL neighborhood. This yields policies stable under downstream perturbations---a different notion from parameter-space flatness.

\vspace{-0.1in}
\paragraph{Distributionally robust optimization and robust RL.}
Distributionally robust optimization (DRO) optimizes worst-case performance over uncertainty sets \citep{Shapiro01012002,kuhn2019wasserstein, duchi2021learning,shapiro2017}.
Applications include group distributionally robustness and domain adaptation \citep{sagawa2019distributionally,oren2019distributionally,sinha2017certifying}. In RL, robust MDPs consider adversarial dynamics \citep{vinitsky2020robust,pinto2017robust}, while risk-sensitive RL optimizes CVaR \citep{cvar2015chow,tamar2015optimizing} or entropic risk \citep{osogami2012robustness,fei2021risk,fei2021exponential}. Unlike robust MDPs \citep{smirnova2019distributionally,eysenbach2021maximum,zhang2020robust,derman2020distributional} that perturb environment dynamics and demand robust decisions, our method perturbs the \emph{policy itself} in an RLHF setting to stay robust to downstream fine-tuning. We apply DRO to the policy space with KL constraints, yielding an entropic risk objective over trajectories.

\vspace{-0.1in}
\paragraph{Adversarial training for alignment.}
Several methods train models to preserve alignment after adversarial fine-tuning. TAR \citep{tamirisa2024tamper} performs iterative adversarial fine-tuning to identify vulnerable points; Circuit Breaking \citep{zou2024improving} projects unsafe inputs to incoherent outputs, and Representation Noising \citep{rosati2024representation} drives harmful representations toward random noise. Another line of work \citep{huang2024vaccine,casper2024defending} finds more robust representation against downstream perturbations. 
These methods involve expensive inner-loop optimization (TAR) or modify internal representations (Representation Rerouting) that is primarily applicable to safety rather than a general objective. In contrast, we derive a closed-form solution for optimizing a general-purpose reward and demonstrate broader effectiveness on mathematical reasoning and continual learning as well.  Additional related work is provided in \Cref{additional}.

%% file: files_arxiv/exp_arxiv2.tex
\section{Experiments}\label{sec:experiments}
Our experimental investigation focuses on two scenarios: (i) \emph{Safety training with RLHF}, where we aim to mitigate the catastrophic forgetting of safety guardrails; and (ii) \emph{Math training with RL}, which represents a broader non-safety application where we study the preservation of mathematical accuracy after fine-tuning on code generation.

\subsection{Robustness in Safety Training}\label{sec:exp-safety}

We train two model families: Mistral-v0.1-Instruct \citep{jiang2023mistral7b} and Qwen2.5-7B-Instruct \citep{qwen2025qwen25technicalreport}. 
We include Mistral because its lack of built-in guardrails makes the contrast between our method and GRPO clearer. Additionally, Qwen is a more capable model and a standard baseline for GRPO in the literature and does not have strong guardrails.

\begin{figure*}
    \centering
    \begin{subfigure}{0.333\linewidth}
        \includegraphics[width=\linewidth]{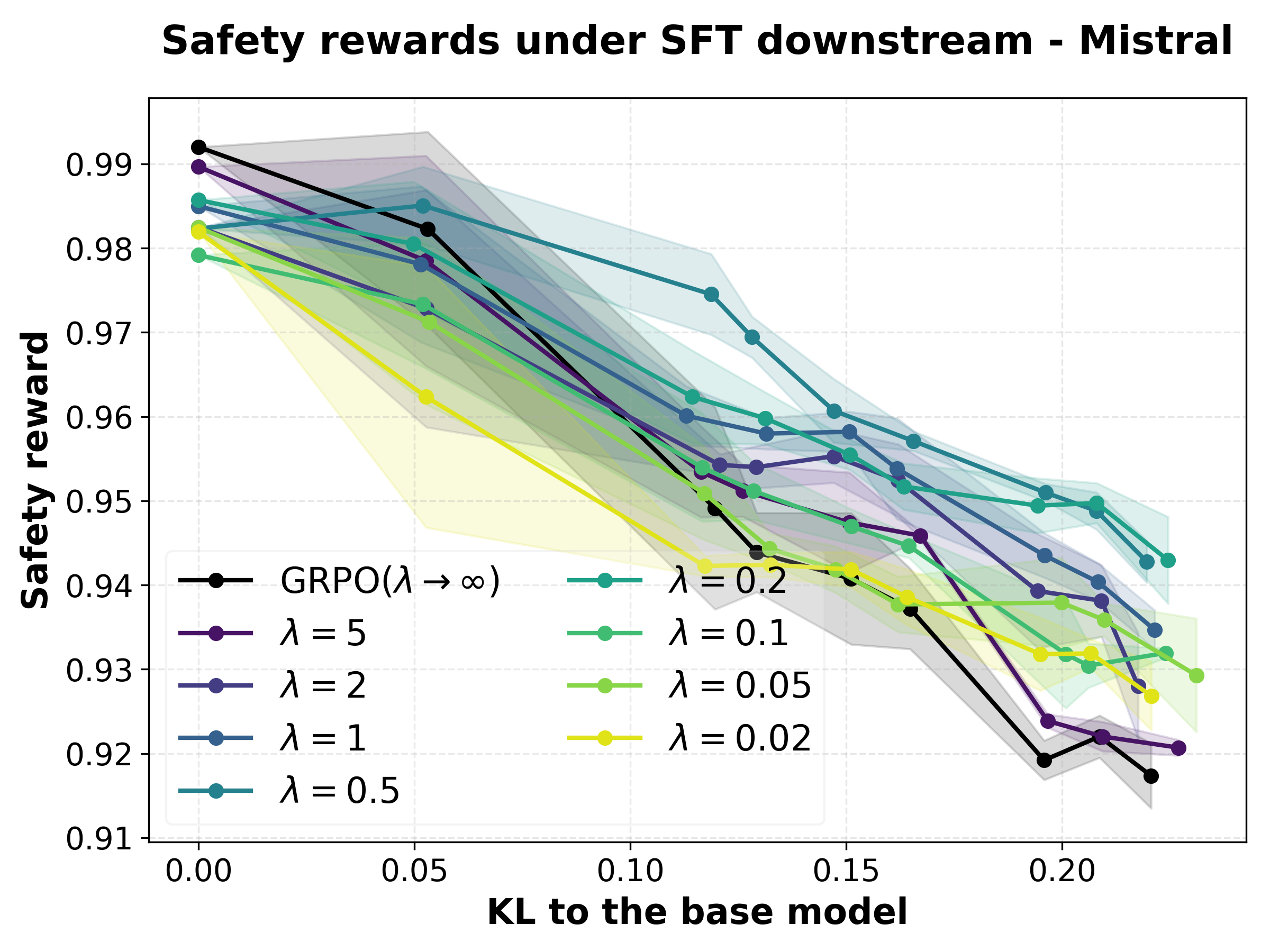}
    \end{subfigure}%
    \begin{subfigure}{0.333\linewidth}
        \includegraphics[width=\linewidth]{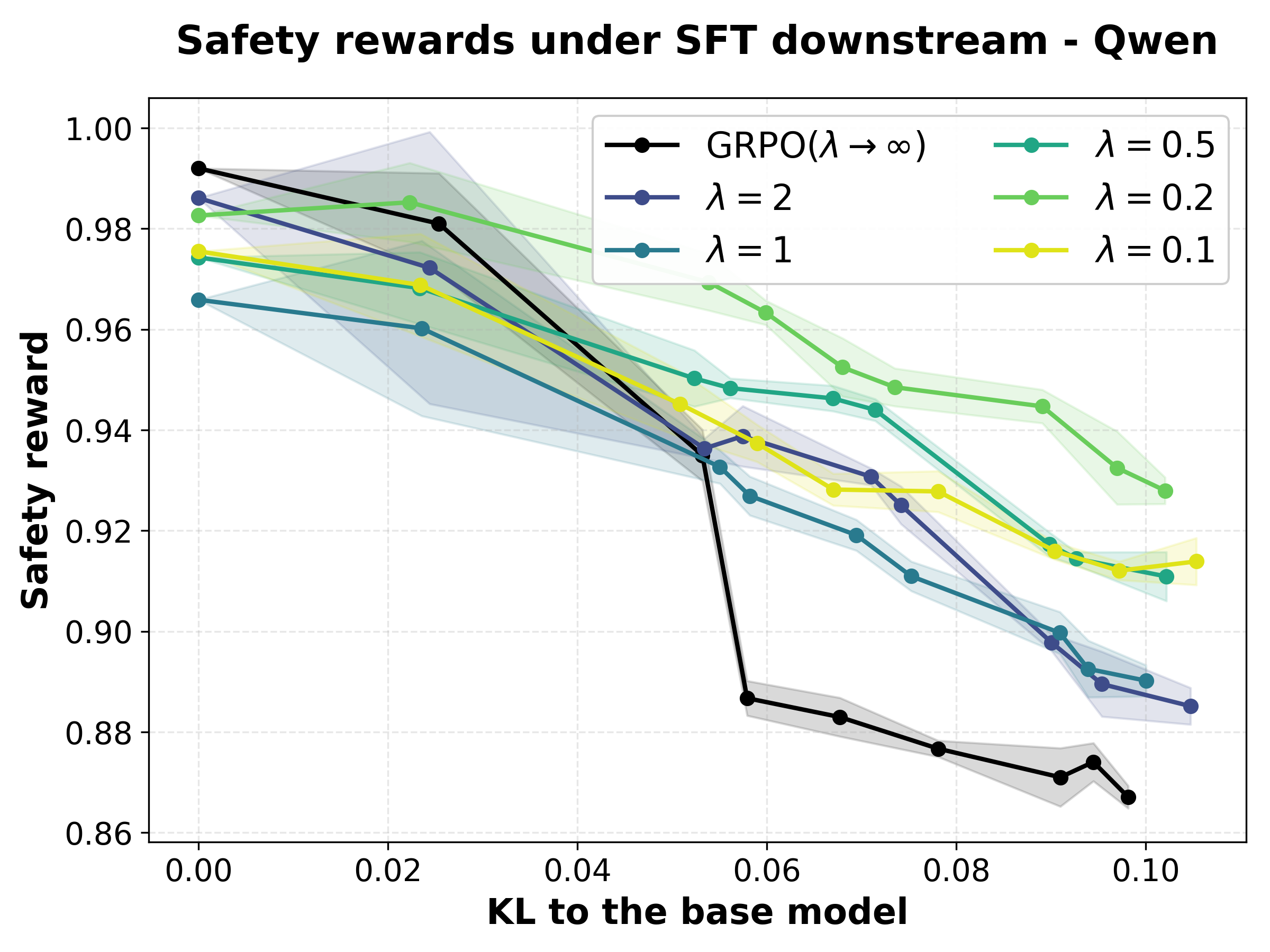}
    \end{subfigure}%
    \begin{subfigure}{0.333\linewidth}
        \includegraphics[width=\linewidth]{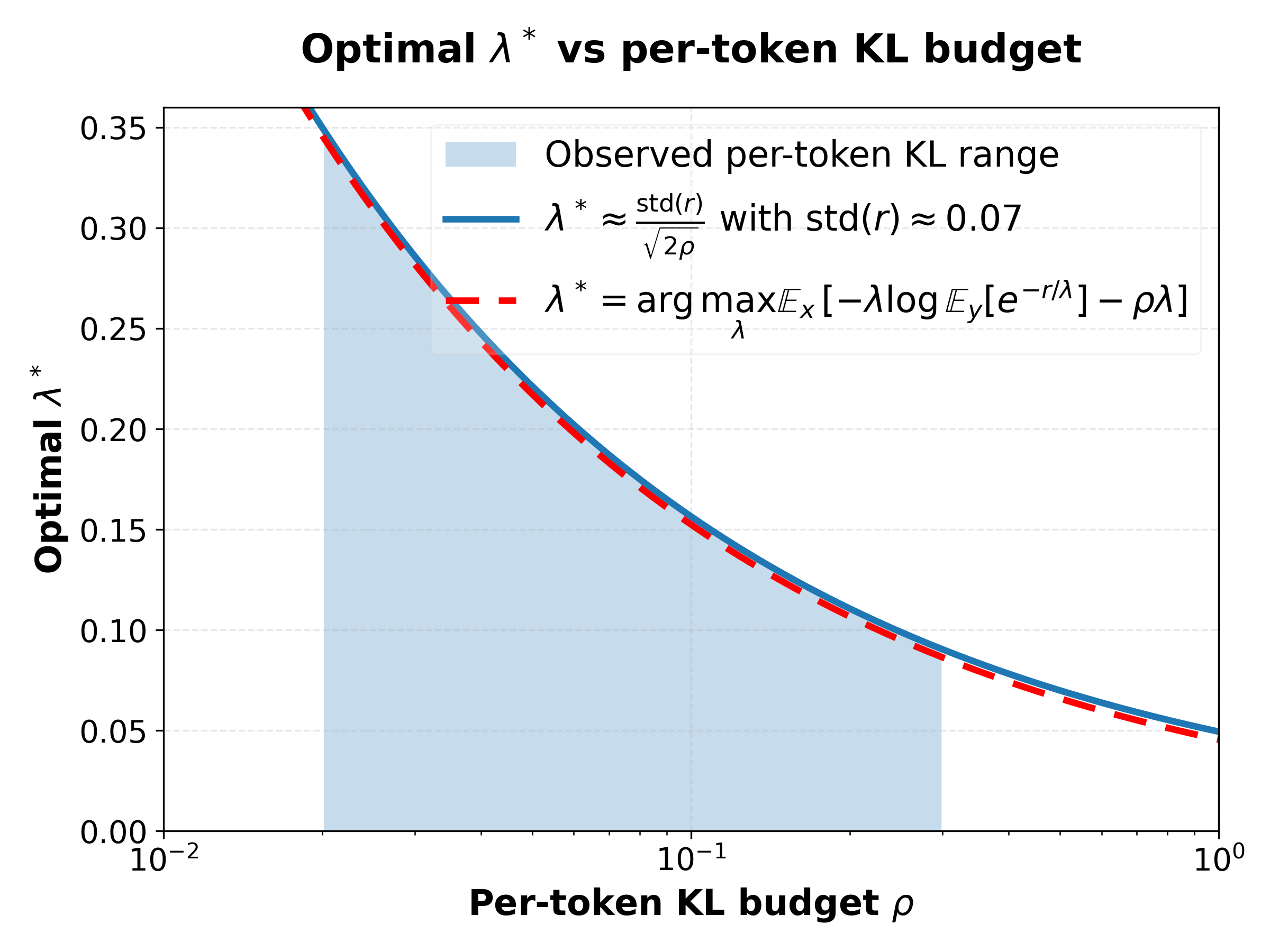}
    \end{subfigure}
    \vspace{-10pt}
     \caption{\textbf{(left/middle)} The safety reward for Mistral and Qwen as KL increases during fine-tuning on Alpaca, when sweeping $\lambda$, and evaluated on a split of the safety prompts; $\lambda = 0.2$ better preserves the safety reward for both models and yields the most flat landscape. 
     \textbf{(right)} The optimal $\lambda$ observed in the experiment fall in the predicted interval. We can choose $\lambda$ without any sweeping.}
     \vspace{-10pt}
     \label{fig:lambda}
\end{figure*}

\vspace{-0.1in}
\paragraph{Training Results.} 
Our safety dataset contains 1000 harmful and 1000 harmless prompts (to avoid over-refusal). We use separate rewards: a safety score for harmful prompts, and a helpfulness reward for harmless prompts. Details are in \Cref{app:training}. We show that FRPO achieves similar training results as GRPO (\Cref{fig:training}) with no difference in safety evaluations (see below). 

Additionally, in \Cref{ablation:higher_KL}, we show that allowing for a larger KL by reducing $\beta$ enables FRPO to find more reward-flat solutions that better preserve the reward under fine-tuning.

\vspace{-0.1in}
\paragraph{Evaluations.}
We report the refusal rate on HarmBench and the harmfulness score from StrongREJECT \citep{mazeika2024harmbench,souly2024strongreject} under three downstream fine-tuning settings:
\textbf{(i)} SFT on Alpaca \citep{alpaca}, which prior work \citep{qi2023fine} reports can harm the guardrails;
\textbf{(ii)} SFT on GSM8K \citep{cobbe2021training}, testing vulnerability under math fine-tuning similar to \citep{qi2024safety};
\textbf{(iii)} RL on UltraFeedback \citep{cui2023ultrafeedback}, probing the helpfulness-harmlessness tradeoff \citep{bai2022training, tan2025equilibrate}.
All results are averaged over 3 runs.

\vspace{-0.1in}
\paragraph{Baselines.} 
Our primary baselines are GRPO-trained models 
using the same base model and safety dataset, which ensures a controlled comparison. 
For broader comparison, we include: \textbf{(i)} Llama-3.1-8B \citep{dubey2024llama3herdmodels}, a model with standard guardrails; \textbf{(ii)} Llama-3-TAR-refusal \citep{tamirisa2024tamper}, 
which runs iterative adversarial fine-tuning to find vulnerable policies;
\textbf{(iii)} Llama-3-RR and Mistral-RR \citep{zou2024improving} models, which use an unlearning approach;
\textbf{(iv)} Llama-3-derta \citep{yuan2025refuse}, which improves robustness to jailbreaking attacks. We also compare with GRPO + SAM \citep{foret2020sharpness,watts2026sharpness} in \Cref{sec:SAM} trained with the same setting.


\subsubsection{SFT on Alpaca and GSM8K}\label{sec:alpaca}
We evaluate the robustness of guardrails to instruction-tuning via full-parameter SFT on Alpaca \citep{alpaca} and GSM8K \citep{cobbe2021training} for mathematical reasoning. We follow the experimental setup of \citep{qi2023fine,qi2024safety}, which showed that fine-tuning on these datasets increases the attack success rate on Llama-2 models. We use $\mathrm{lr}=10^{-6}$ for Alpaca and $\mathrm{lr}=6\times10^{-6}$  for GSM8K, training for one epoch in both cases. We first discuss how we select the optimal $\lambda$ for our experiments. We then compare the results with those of other baselines. 


\vspace{-0.1in}
\paragraph{Role of $\lambda$ in the reward landscape.} 
For illustrating the role of $\lambda$, we train both Mistral and Qwen with $\lambda$ in $[0.05, 5]$. We then fine-tune all models on Alpaca and measure the safety reward on the safety-training data. For different values of $\lambda$, \Cref{fig:lambda} (left/middle) shows the trajectory of the safety reward on the landscape during fine-tuning as the KL increases. 
When decreasing $\lambda$ from GRPO ($\lambda\to\infty$), the final reward improves up to the optimal point and decreases afterwards. Early in fine-tuning, when the KL is small, GRPO has the highest reward because it optimizes the average reward. But as KL grows, its reward falls below other $\lambda$ values. 
For Mistral, $\lambda  \in \{0.2,0.5\}$ perform best in the explored KL range. Similarity, for Qwen, $\lambda = 0.2$ better preserves the reward.

\vspace{-0.1in}
\paragraph{Selecting $\lambda$.}
According to Remark \ref{remark:variance}, $\lambda$ is approximately optimal at $\frac{\mathrm{std}(r(x,y))}{\sqrt{2\rho}}$. \Cref{fig:training} (g, h) demonstrate that reward statistics remain consistent across various $\lambda$ values, with $\mathrm{std}(r(x,y)) \sim 0.07$ by the end of training for both Qwen and Mistral. Using this estimate, we plot the theoretical $\lambda$ as a function of the divergence budget $\rho$. In the same figure, we also present the optimal $\lambda$ derived from \Cref{eq:with_rho} by collecting rewards on a validation set and numerically optimizing for each $\rho$. 


To determine a suitable target range for \( \rho \), we note that the per-token KL values consistently fall within \([0.02, 0.3]\) across all checkpoints (see \Cref{fig:lambda}, left and middle panels). Moreover, as shown in \Cref{fig:reward_vs_KL}, helpfulness begins to decline once \( \text{KL} \gtrsim 0.3 \), indicating the onset of overfitting across models. This interval is highlighted as the blue region in \Cref{fig:lambda} (right panel).

The corresponding optimal values \( \lambda^* \) for this KL range lie in \([0.09, 0.35]\), which includes the empirically selected value \( \lambda = 0.2 \), thereby aligning our analysis with the experimental observations. Additionally, performance is relatively insensitive to the precise choice of \( \lambda \): values such as \( \lambda \in \{0.1, 0.5\} \) still outperform other settings (see left and middle panels in \Cref{fig:lambda}). Based on this, we fix \( \lambda = 0.2 \) for the main comparisons.

\vspace{-0.1in}
\paragraph{Main results.} 
Results for fine-tuning on Alpaca are shown in \Cref{fig:alpaca}. The first two columns demonstrate gains over GRPO and Mistral-RR, even though Mistral-RR unlearns unsafe content. The right column compares against broader baselines; Models trained with $\lambda=0.2$ maintain the highest refusal rates on HarmBench and match Llama-3-RR on StrongREJECT. Moreover, \Cref{sec:normal_alpaca} confirms that this improved safety is not a byproduct of overfitting or performance degradation: results on Alpaca are consistent with GRPO and general capabilities remain close across models.

\begin{figure}
    \centering
    \begin{subfigure}{0.41\textwidth}
        \includegraphics[width=\linewidth]{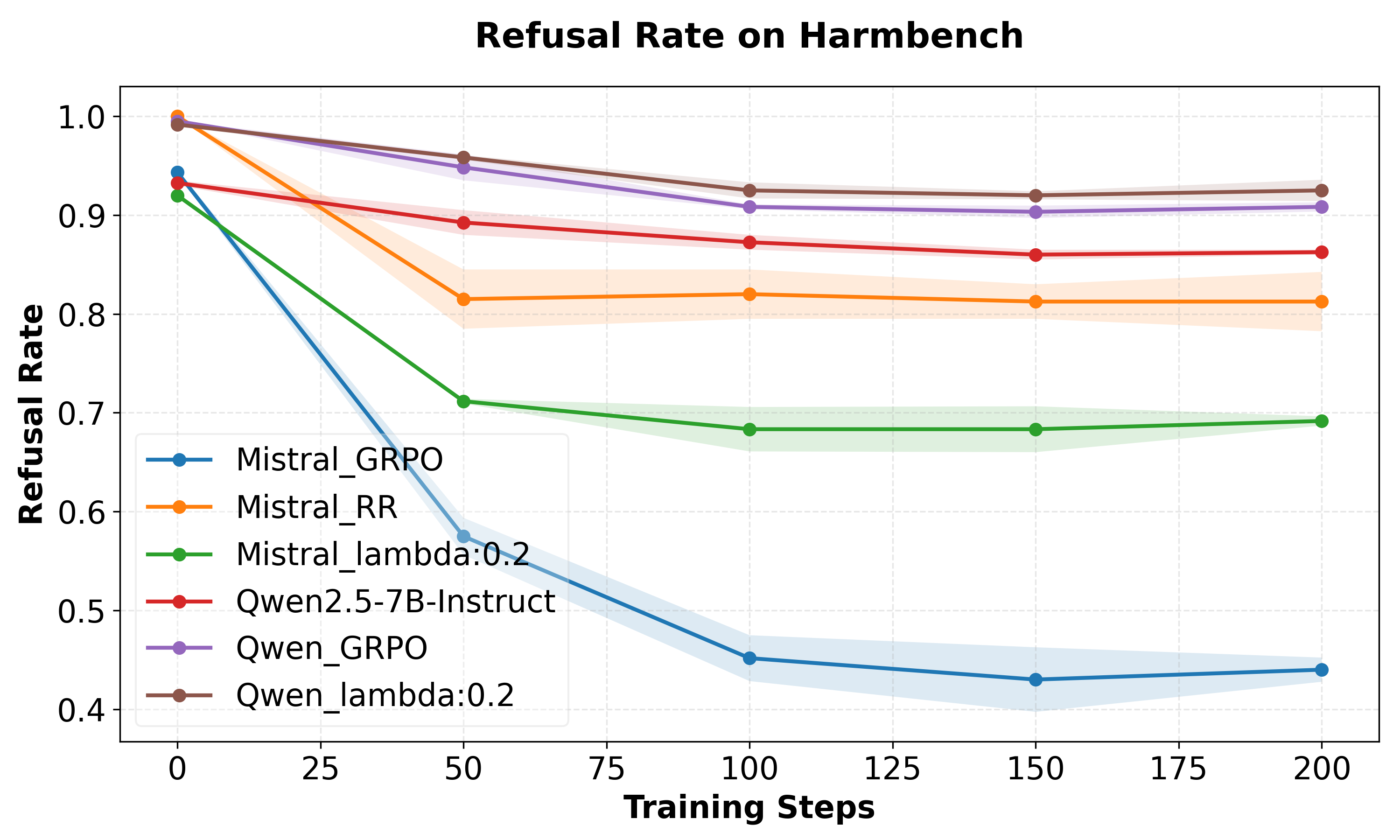}
    \end{subfigure}
    \hspace{0.03\textwidth}
    \begin{subfigure}{0.41 \textwidth}
        \includegraphics[width=\linewidth]{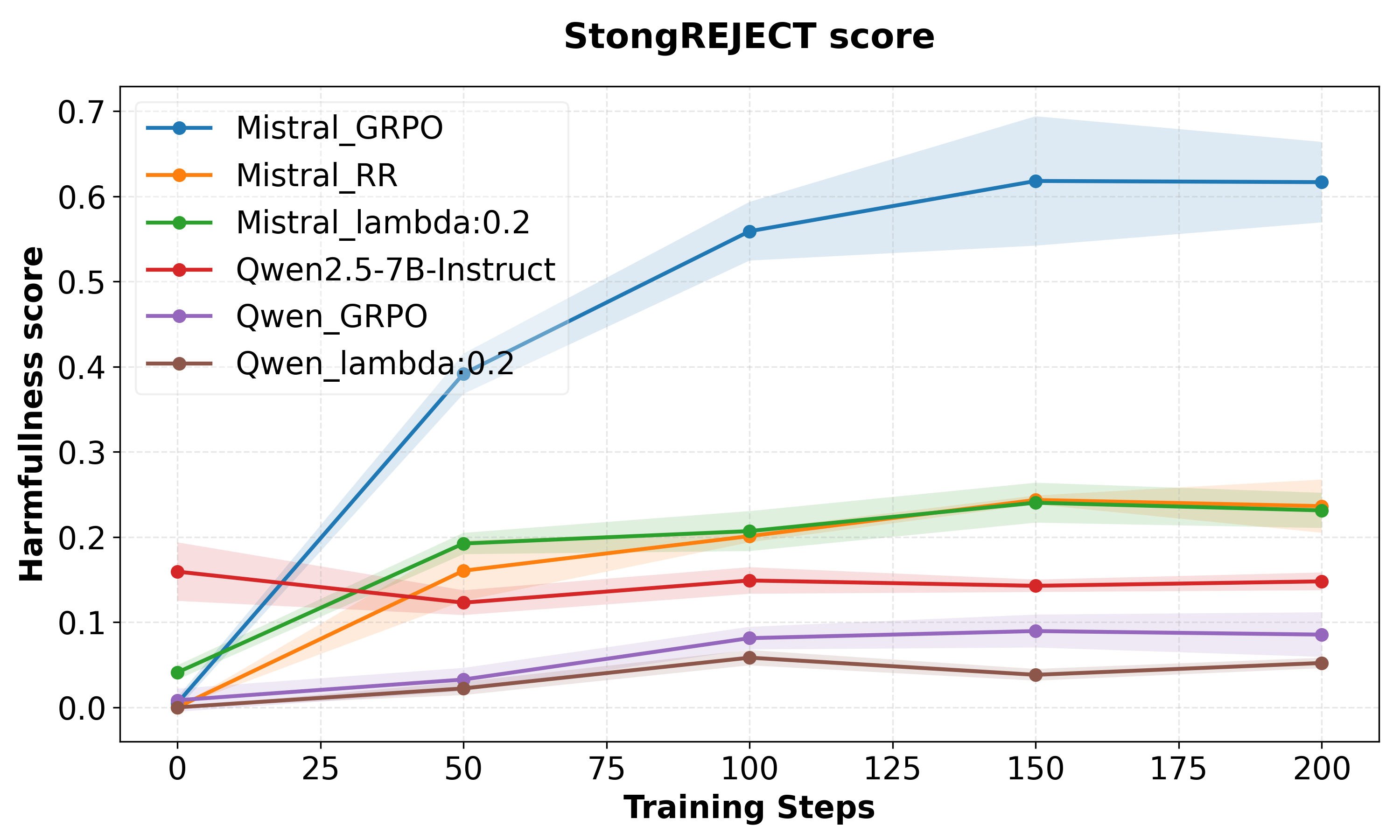}
    \end{subfigure}
    \vspace{-6pt}
     \caption{Safety metrics during GSM8k SFT for Mistral and Qwen models. Our method maintains higher refusal rates (left, $\uparrow$ is better) and better StrongREJECT scores (right, $\downarrow$ is better).}
     \label{fig:GSM8K_lim}
\end{figure}

For math fine-tuning on GSM8K (\Cref{fig:GSM8K_lim}), we focus on Mistral- and Qwen-based baselines, deferring a broader comparison to \Cref{sec:GSM8K_moreresults}. Qwen models trained with $\lambda=0.2$ better preserve safety on both benchmarks, whereas Mistral exhibits a larger gap with GRPO as our method significantly slows its safety degradation. The gap with GRPO is larger for Mistral, and our method significantly slows the safety degradation.
In general, we observe a sharper drop in safety for Mistral models than for Qwen and other baselines under GSM8K SFT. 

\vspace{-0.1in}

\paragraph{Is improved safety an artifact of reduced downstream adaptation?}  
We investigate whether the preservation of safety guardrails is simply a byproduct of weaker downstream adaptation—i.e., the model changes less and therefore forgets less. To this end, we evaluate models before and after GSM8K fine-tuning on several benchmarks: GSM8K, MMLU (general capabilities), IFEval (instruction following), and HumanEval (coding). As shown in \Cref{table:GSM8K_fine-tune}, post–fine-tuning performance on these benchmarks is comparable between our method and GRPO. 

These results suggest that the observed safety improvements are not due to a failure to learn the downstream task, but rather stem from updates that better preserve safety while maintaining competitive downstream performance.


\begin{table*}
  \centering

  \vspace{-4pt}
    \begin{tabular}{@{}lcccccccc@{}}
      \toprule
      & \multicolumn{2}{c}{GSM8K (maj@8)} 
      & \multicolumn{2}{c}{MMLU (5-shot)} 
      & \multicolumn{2}{c}{IFEval (pass@1)} 
      & \multicolumn{2}{c}{HumanEval} \\
      \cmidrule(lr){2-3} \cmidrule(lr){4-5} \cmidrule(lr){6-7} \cmidrule(l){8-9}
      Base Models & Base & FT & Base & FT & Base & FT & Base & FT \\
      \midrule
      Mistral-v0.1-Instruct & 50.5 & 57.4 & 55.0 & 53.7 & 39.6 & 37.2 & 28.7 & 23.2 \\
      Mistral-GRPO & 48.4 & 55.5 & 54.9 & 54.1 & 34.4 & 38.0 & 32.3 & 24.4 \\
      Mistral-FRPO($\lambda$=0.2) & 48.7 & 56.2 & 55.2 & 54.2 & 36.6 & 37.0 & 31.1 & 24.4 \\
      \bottomrule
    \end{tabular}
    \vspace{-4pt}
    \caption{Downstream task performance before (Base) and after (FT) GSM8K fine-tuning. FRPO achieves similar scores to GRPO 
  across all benchmarks. This confirms that the improved safety retention shown in \Cref{fig:GSM8K_lim} is not due to weaker downstream adaptation.}
  \vspace{-5pt}
  \label{table:GSM8K_fine-tune}
\end{table*}

\vspace{-0.1in}
\subsubsection{More Fine-tuning Settings}
\begin{figure}
    \centering
    \begin{subfigure}[b]{0.48\textwidth}
        \includegraphics[width=\linewidth]{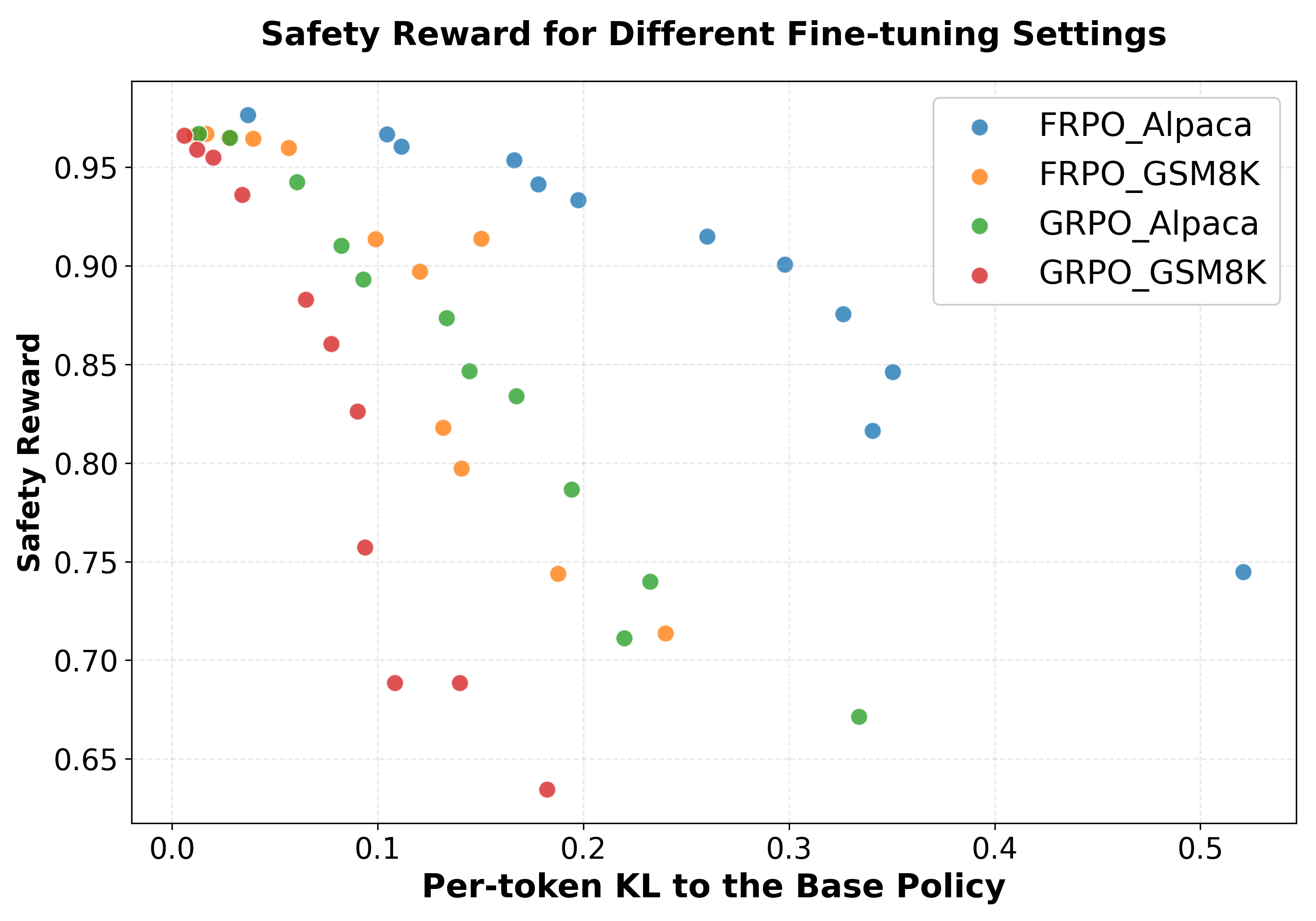}
    \end{subfigure}
    \hspace{0.02in}
    \begin{subfigure}{0.48\linewidth}
        \includegraphics[width=\linewidth]{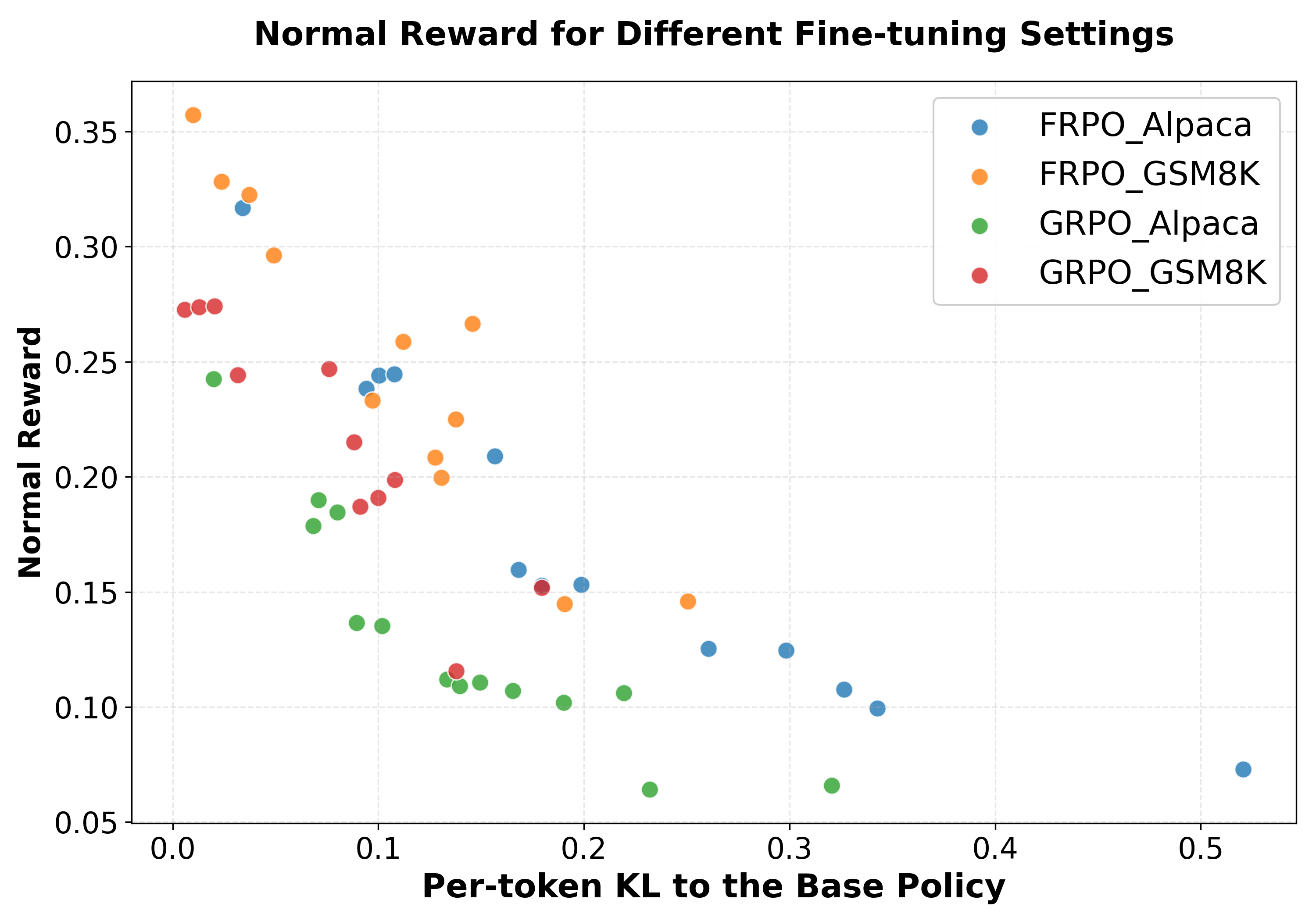}
    \end{subfigure}
     \caption{The safety reward \textbf{(left)} and the helpfulness reward \textbf{(right)} for Mistral trained with FRPO and GRPO and evaluated after fine-tuned with full SFT or LoRA (ranks $\in \{16, 32\}$), $\mathrm{lr} \in \{\mathrm{1e-6}, \mathrm{3e-5},  \mathrm{6e-5}, \mathrm{1e-5, 3e-5, 1e-4} \}$, for 1 or 2 epochs, and on Alpaca or GSM8K. Each point represent one of the 48 sampled settings from their combinations.  \textbf{Takeaway:} The helpfulness reward completely collapses after per-token-KL = 0.3 (i.e., overfitting); even a full-parameter SFT with a relatively large learning-rate that still does not overfit falls in the considered KL-ball.} 
     \label{fig:48_settings}
\end{figure}

We conduct another experiment in which we include 48 fine-tuning settings (e.g., with or without LoRA, different learning rates, different datasets, etc.) to examine what KL ball to the base model is reasonable so as to not become fully overfit and lose the helpfulness score completely. This figure is shown in \Cref{fig:48_settings}. Our conclusions are: \textbf{(i)} Most settings fall between 0.02 and 0.3 in per-token-KL. Beyond this KL ball, the model’s standard capabilities collapse and overfit, as shown on the right. Therefore, optimizing for this range achieves near-optimality. \textbf{(ii)} Across this wide range of settings, FRPO outperforms GRPO; even for smaller KL distances (e.g., LoRA with lr=1e-6, 1 epoch), FRPO stays higher.

\vspace{-5pt}
\subsubsection{RL on Helpfulness}\label{sec:helpfulness}
\vspace{-5pt}
We also study whether the same robustness holds under RL fine-tuning. In \Cref{sec:app_helpfulness}, we fine-tune on UltraFeedback with GRPO and evaluate the resulting helpfulness--safety trade-off. Consistent with our SFT results, smaller $\lambda$ better preserves safety while larger $\lambda$ slightly improves helpfulness but induces more safety degradation.

\subsection{Continual Learning for Math Training}\label{sec:math}
We test whether our method reduces forgetting of math capabilities after further fine-tuning. 
We start from Qwen2.5-Math-7B and train on MATH \citep{hendrycks2021measuring} (levels 3–5) using the Qwen-Math template, following \citep{liu2025understanding}. We use an outcome-only 0–1 reward that verifies the final answer, 
with a small bonus when a final answer is provided (details in \Cref{app:math_training}). To show the behavior in term of $\lambda$ again, we train with GRPO and FRPO for $\lambda \in [0.1, 10]$. We first show that all models reach a similar accuracy on MATH500. We then fine-tune all math-trained models with SFT on 25k samples from "nvidia/OpenCodeInstruct" \citep{ahmad2025opencodeinstruct}, an instruction-tuning dataset for code generation. The goal is to improve performance on coding benchmarks such as MBPP+ \citep{austin2021program,liu2023your} while avoiding an accuracy drop on MATH500, thereby demonstrating improved continual learning across math and coding.

\vspace{-0.1in}
\paragraph{Math training.} Training results are in \Cref{app:math_training}. All the models behave similarly during training and converge to a similar accuracy: the top row of \Cref{table:math_ft} shows that all models reach roughly $73\%$ on MATH500, improving over Qwen2.5-Math-7B (56.6\%). We also report MBPP+ results after math training in the third row of \Cref{table:math_ft}. All models remain close to the base MBPP+ accuracy (57.4\%).

\vspace{-0.1in}
\paragraph{Code fine-tuning.}
We fine-tune all math-trained models on "OpenCodeInstruct" using SFT. Since full-parameter fine-tuning without LoRA caused a substantial drop in MATH500 accuracy, we use LoRA  with rank $r=16$ and $\mathrm{lr = 5 \times 10^{-5}}$ for all models. Results on MATH500 and MBPP+ are summarized in \Cref{table:math_ft}. FRPO with $\lambda=2.0$ preserves MATH500 accuracy best across models, outperforming GRPO by 22\%. On MBPP+, all the models improve similarly, with a $4$--$5\%$ gain.

The optimal $\lambda$ here differs from the optimal $\lambda$ from safety (where $\lambda$ =0.2). This is because in the math setting, the final standard deviation is roughly $10\times$ larger than in safety training ($\sim 0.5$ vs.\ $\sim0.05$). Per our discussion in \Cref{sec:alpaca}, this leads to $\lambda =2$, which exactly matches our experiments. 

\begin{table*}
\centering
\vspace{-4pt}
\resizebox{\linewidth}{!}{%
\begin{tabular}{@{}ll cccccccc@{}}
\toprule
& &  \makecell{GRPO\\($\lambda {\scriptstyle \to} \infty$)} &  \makecell{FRPO\\($\lambda$=10)} &   \makecell{FRPO\\($\lambda$=4)} &  \makecell{FRPO\\($\lambda$=2)} &  \makecell{FRPO\\($\lambda$=1)} &  \makecell{FRPO\\($\lambda$=0.5)} &  \makecell{FRPO\\($\lambda$=0.2)} &  \makecell{FRPO\\($\lambda$=0.1)} \\
\midrule
\multirow{2}{*}{MATH500} 
& Base & 73.0 & 73.0 & 73.2 & 73.0 & 73.0 & 73.4  & 73.4 & 71.6 \\
& FT   & 42.3 ($\pm$ 2) & 44.5 ($\pm$ 3) & 61.0 ($\pm$ 3) & \textbf{64.5} ($\pm$ 2)& 63.1($\pm$ 2) & 53.6($\pm$ 2) & 59.3 ($\pm$ 4) & 54.6 ($\pm$ 3) \\
\midrule
\multirow{2}{*}{MBPP+} 
& Base & 57.1 & 57.9 & 57.4 & 57.9 & 58.2 & 57.9 & 57.1 & 57.7 \\
& FT   & 62.5 & 62.2 & 61.2 & 62.4 & 62.0  & 61.8 & 61.7 & 61.7 \\
\bottomrule
\end{tabular}%
}
\vspace{-4pt}
\caption{Performance before (Base) and after (FT) code fine-tuning all the math-trained model on "OpenCodeInstruct". FRPO with $\lambda=2.0$ best preserves math accuracy while achieving comparable coding performance. \uline{Results are averaged over 3 fine-tuning seeds.}}
\vspace{-5pt}
\label{table:math_ft}
\end{table*}

\subsection{Ablation: Comparison with SAM and other methods}\label{sec:SAM}
\begin{wrapfigure}{r}{0.38\linewidth}
  \vspace{-8pt}
  \centering
  \includegraphics[width=1.0\linewidth]{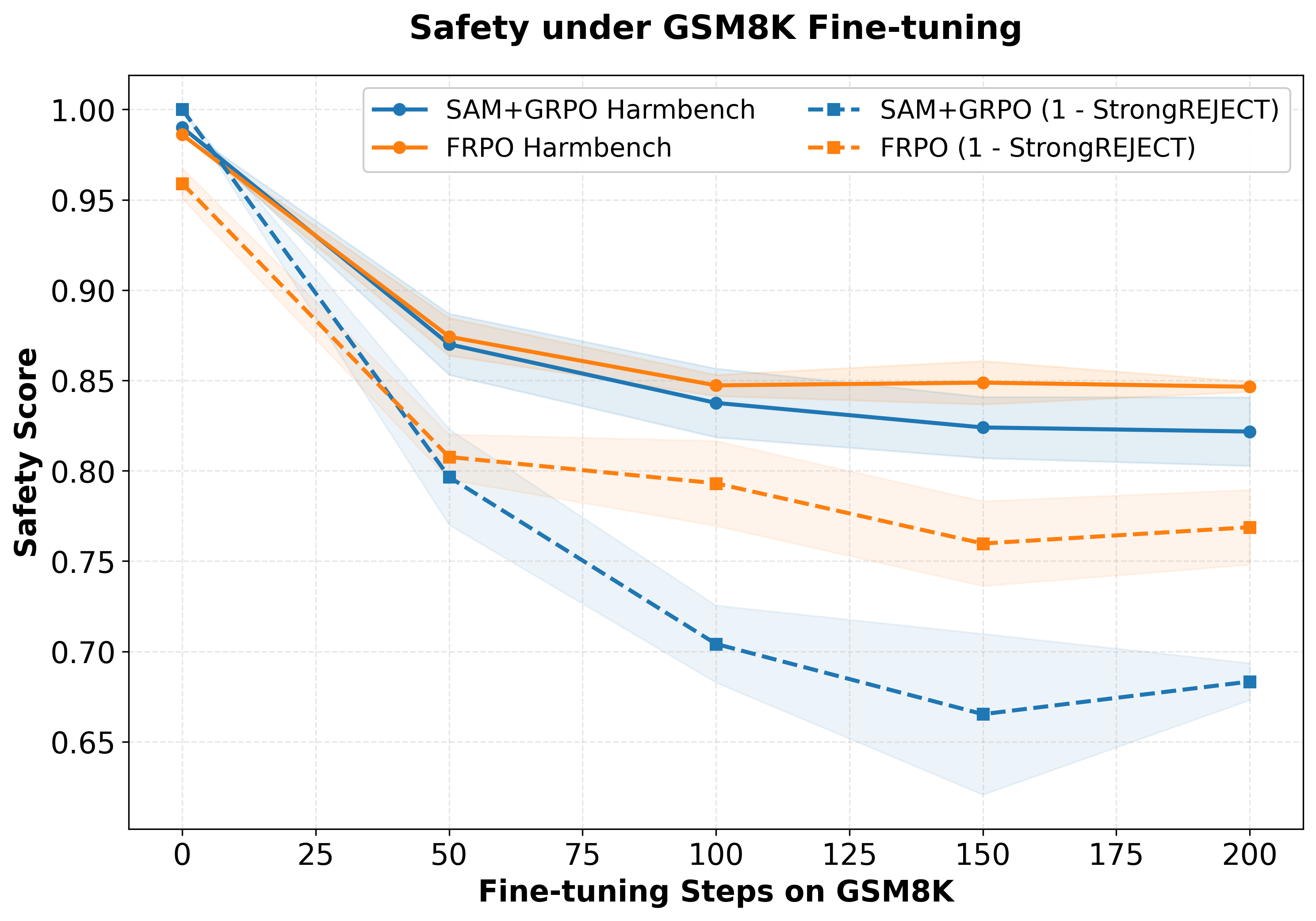}
  \vspace{-9pt}
  \caption{FRPO preserves the safety reward higher than GRPO+SAM after fine-tuning on GSM8K.}
  \label{fig:sam}
  \vspace{-15pt}
\end{wrapfigure}
Similar to FRPO, SAM also considers a max-min optimization and seeks flat regions \citep{foret2020sharpness,watts2026sharpness}, but in the parameter space regarding the L2 distance. We implemented SAM  on top of GRPO ($P=2$, $\rho=0.05$ as their default). \Cref{fig:sam} shows that at downstream time on GSM8K, GRPO+SAM improves over GRPO in maintaining the safety reward but still falls below FRPO with $\lambda =0.2$. 
SAM's L2 ball in the parameter space primarily concerns generalization, not continual learning, and does not cover all downstream policy shifts, while the performance is fragile w.r.t the perturbation radius and degrades above the optimal $\rho$ \citep{bahri2022sharpness}. I.e., the L2 ball can include irrelevant parameter perturbations, whereas the KL ball avoids this by directly constraining the output distribution shift.

We further compare with Replay methods in \Cref{sec:replay}. 

%% file: files_arxiv/conclusion_arxiv.tex
\section{Conclusion}
We proposed that robustness to downstream fine-tuning should be incorporated directly into the base policy during RLHF, rather than relying on later interventions. Our approach begins by optimizing reward stability within a KL-bounded neighborhood of policies. Solving this resulting max–min formulation yields FRPO, a robust policy gradient method that identifies reward-flat regions in policy space. Experiments demonstrate that this form of robustness transfers across diverse domains, including safety alignment and mathematical reasoning. Notably, FRPO maintained up to {\bf 22\% higher mathematical accuracy} under code fine-tuning.

This work opens several promising directions. The principle of optimizing for robustness to future adaptation may extend beyond RLHF to pretraining and supervised fine-tuning. Moreover, understanding which capabilities are inherently easier, or harder, to make robust remains an open question, with implications for alignment and continual learning.

%% file: files_arxiv/appendix_arxiv.tex
\section{Additional Related Work}\label{additional}

\paragraph{Catastrophic forgetting.}
Catastrophic forgetting---the abrupt loss of previously learned knowledge when training on new tasks---has been studied since early neural network research \citep{mccloskey1989catastrophic,ratcliff1990connectionist,french1999catastrophic}. The phenomenon arises because gradient updates for new objectives overwrite parameters critical to earlier tasks \citep{goodfellow2013empirical,kirkpatrick2017overcoming}. In LLMs, fine-tuning degrades capabilities acquired during pretraining or alignment \citep{kotha2023understanding,luo2025empirical}. Recent work shows that even benign fine-tuning can remove safety guardrails \citep{qi2023fine,yang2023shadow}, and task-specific adaptation (e.g., math) degrades safety \citep{qi2024safety,zhan2023removing}. This motivates building robustness into the pre-fine-tuning policy rather than relying solely on downstream interventions.

\vspace{-0.1in}
\paragraph{Continual learning.}
Continual learning methods aim to learn new tasks while preserving prior knowledge \citep{de2021continual,wang2024comprehensive}. Regularization-based approaches constrain updates: EWC \citep{kirkpatrick2017overcoming} uses Fisher information, synaptic intelligence (SI) \citep{zenke2017continual} accumulates importance online, and LwF \citep{li2017learning} applies knowledge distillation. For LLMs, MixOut \citep{lee2019mixout} stochastically resets toward pretrained weights, while other methods prevent the policy from drifting too far \citep{li2017learning,schulman2017proximal,lee2019mixout}, or ensure that the model's latent representations are preserved \citep{kirkpatrick2017overcoming, aghajanyan2020better, pan2024lisa}. Rehearsal methods augment downstream training with a subset of previous data or synthetic data generated by the model \citep{rolnick2019experience,sun2019lamol,scialom2022fine,huang2024mitigating}. Parameter-efficient fine-tuning (PEFT) constrain downstream updates to separate modules (e.g., LoRA \citep{hu2022lora} and progressive networks \citep{rusu2016progressive}), thereby reducing parameter interference between the objectives \citep{hsu2024safe, wang2024learn},
or aiming to keep the features orthogonal \citep{wang2023orthogonal,wang2024learn}. 
Finally, recent model merging methods aim to mathematically combine task-specific models post-hoc to retain all capabilities without expensive retraining \citep{ilharco2022editing, wortsman2022model, yi2024safety, djuhera2025safemerge,nobari2025activationinformedmerginglargelanguage}.
All these methods operate at downstream time; our approach instead builds robustness into the upstream policy, making it agnostic to the downstream fine-tuning protocol.

\vspace{-0.1in}
\paragraph{Limitations of downstream methods.}
Downstream methods have several limitations. Replay-based methods require curating samples that cover all capabilities prone to being forgotten. Also, buffer size and sample selection impact both effectiveness and training efficiency, and poor choices lead to continued forgetting \citep{hayes2019memory,prabhu2020gdumb}. Regularization approaches like EWC and SI have been shown to fail when task similarity is low \citep{hsu2018re}. Parameter-efficient methods, despite widespread adoption, do not reliably prevent forgetting---LoRA fine-tuning still degrades prior capabilities, sometimes comparably to full fine-tuning \citep{biderman2024lora}. Most critically, all these methods assume the downstream optimization follows a prescribed recipe. Our upstream approach avoids these issues by building robustness directly into the base policy, making it agnostic to how downstream adaptation is performed.

\section{Additional Experiments}
\subsection{Evaluations on Alpaca}\label{sec:normal_alpaca}

\begin{figure*}
  \centering
  \captionsetup[subfigure]{skip=0pt}
  \begin{subfigure}[c]{0.32\textwidth}
    \centering
    \includegraphics[width=\linewidth]{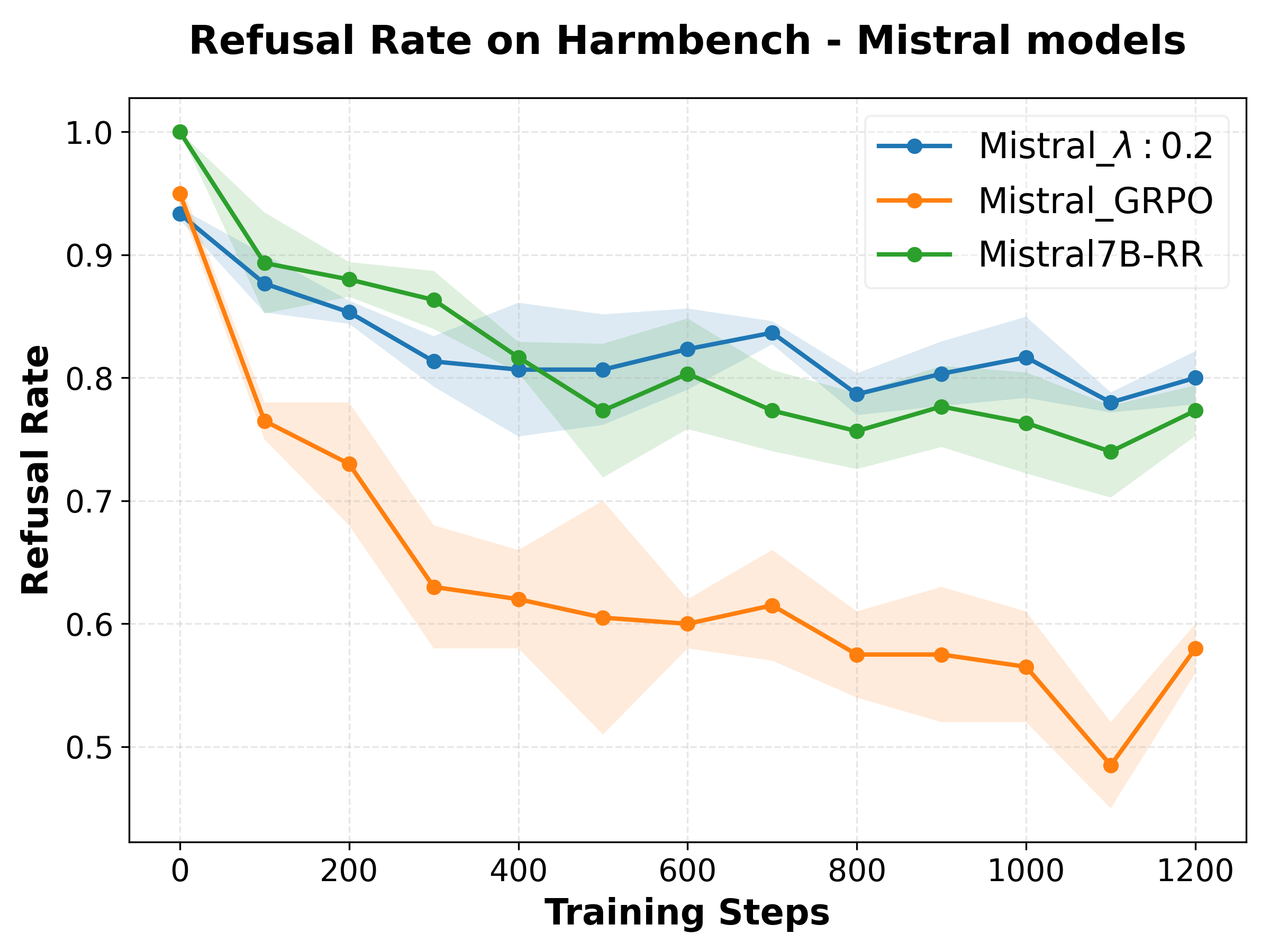}
    \caption{}
  \end{subfigure}%
  \begin{subfigure}[c]{0.32\textwidth}
    \centering
    \includegraphics[width=\linewidth]{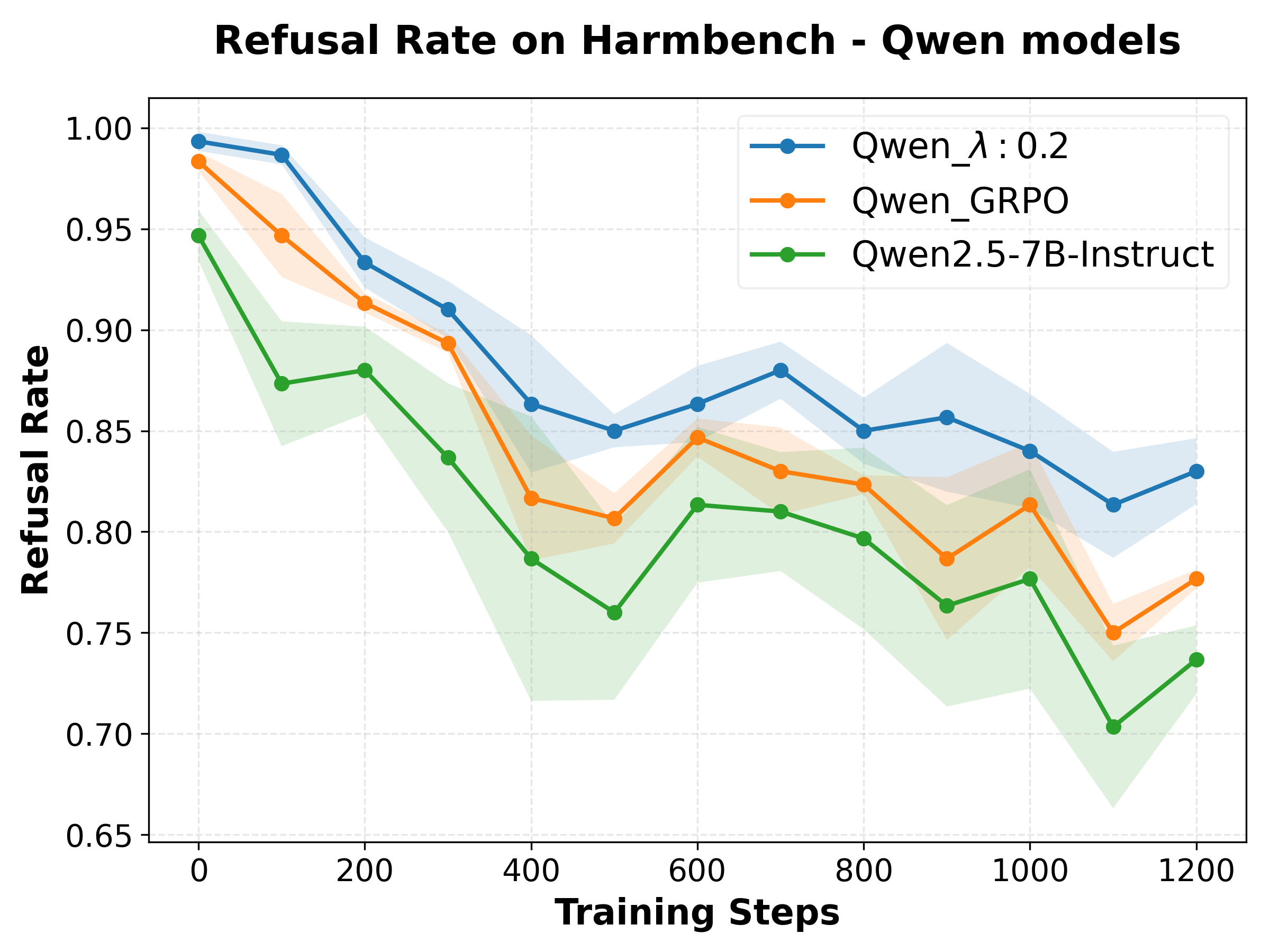}
    \caption{}
  \end{subfigure}%
  \begin{subfigure}[c]{0.32\textwidth}
    \centering
    \includegraphics[width=\linewidth]{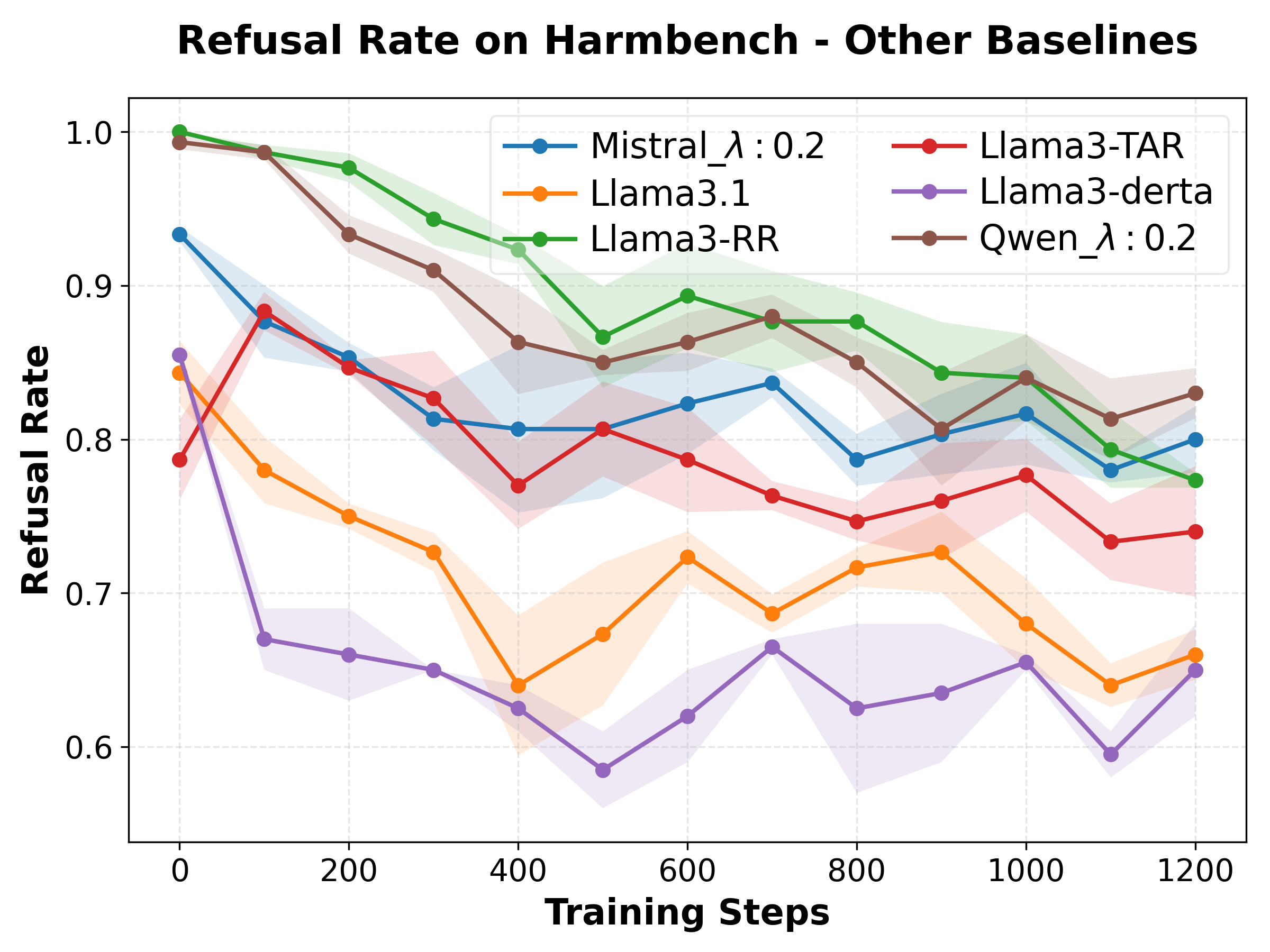}
    \caption{}
  \end{subfigure}
  \\
  \begin{subfigure}[c]{0.32\textwidth}
    \centering
    \includegraphics[width=\linewidth]{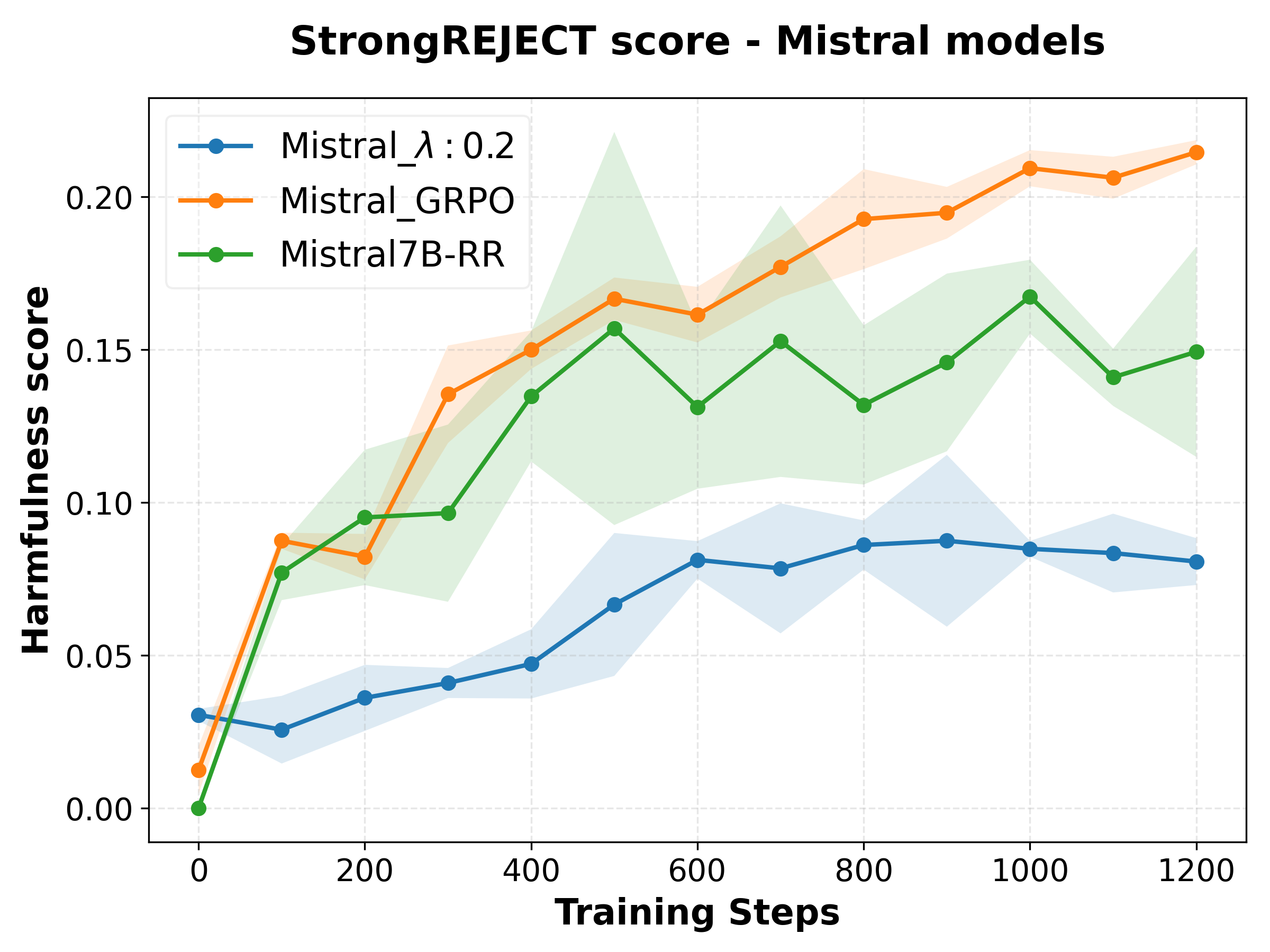}
    \caption{}
  \end{subfigure}%
  \begin{subfigure}[c]{0.32\textwidth}
    \centering
    \includegraphics[width=\linewidth]{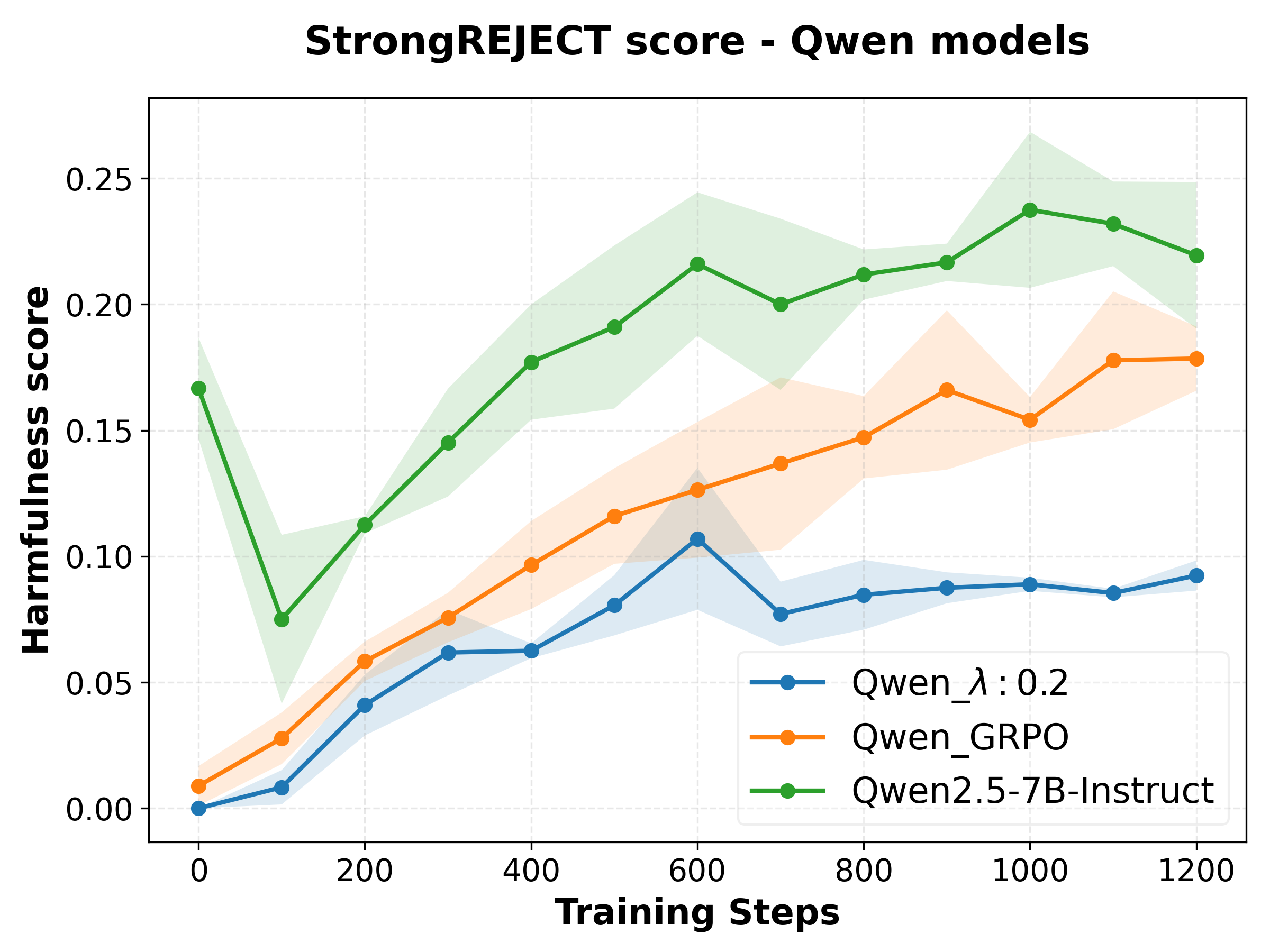}
    \caption{}
  \end{subfigure}%
  \begin{subfigure}[c]{0.32\textwidth}
    \centering
    \includegraphics[width=\linewidth]{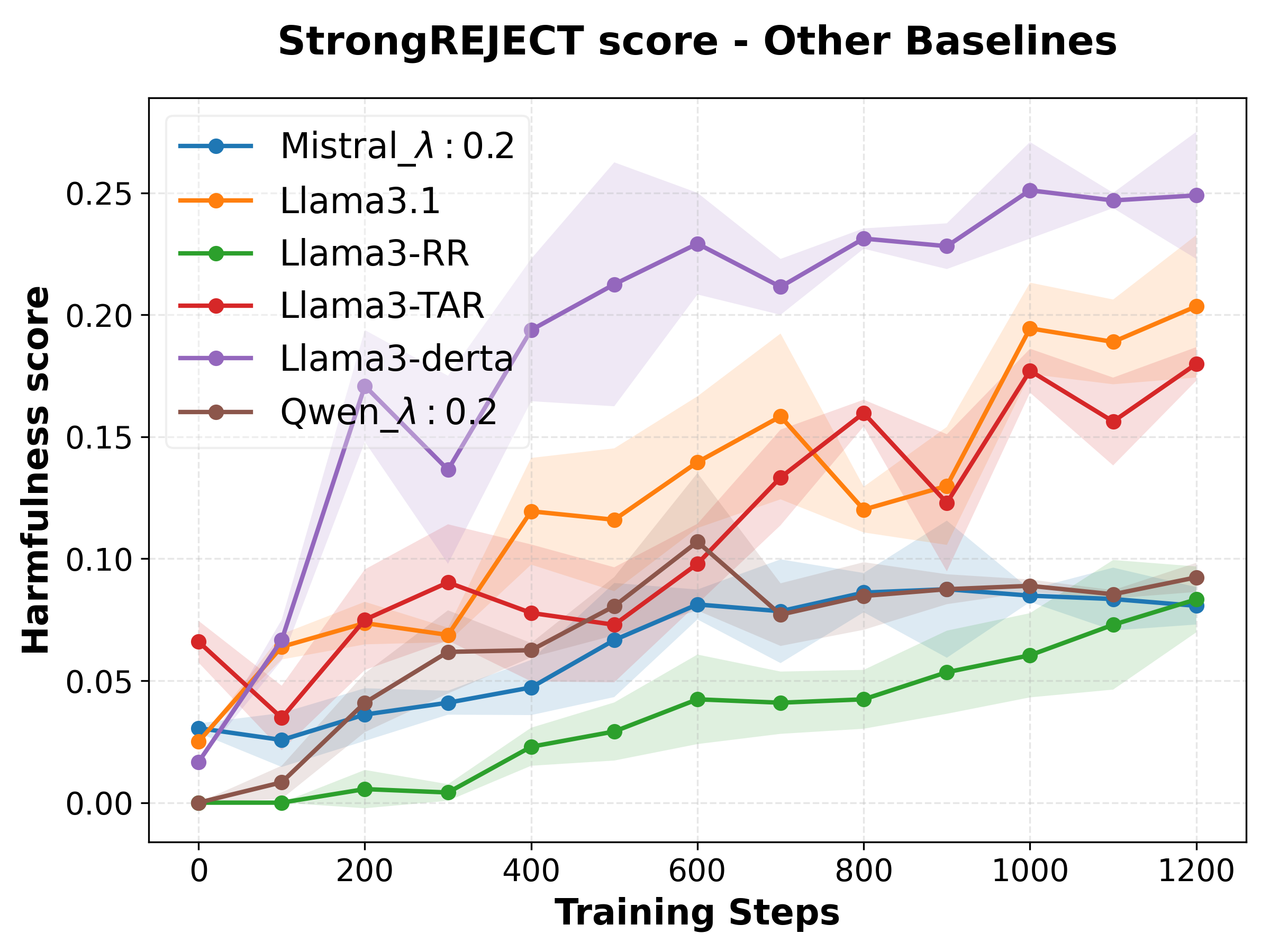}
    \caption{}
  \end{subfigure}
  \vspace{-7pt}
  \caption{Safety evaluation after Alpaca SFT on HarmBench ($\uparrow$ is better) and StrongREJECT ($\downarrow$ is better). \textbf{(a,d)} and \textbf{(b,e)} compare models from the same reference (Mistral and Qwen), showing consistent improvements of our method over GRPO baselines. \textbf{(c,f)} Broader comparison with other safety-focused methods; our approach achieves the highest refusal rates on HarmBench and competitive StrongREJECT scores.}
    \label{fig:alpaca}
  \vspace{-10pt}
\end{figure*}

It is important to demonstrate that the safety robustness conferred by FRPO does not come at the expense of downstream performance. \Cref{fig:alpaca} (left) shows that the SFT loss of the FRPO-trained model aligns closely with that of GRPO and Llama-3.1. In contrast, models such as Llama3-TAR \citep{tamirisa2024tamper} and Llama3-derta \citep{yuan2025refuse} exhibit an initial jump in loss. These models are adversarially trained to resist adaptation, preventing them from fitting the downstream data as effectively as models trained with other methods.

\begin{figure}[t]
    \centering
    \begin{subfigure}[b]{0.48\textwidth}
        \includegraphics[width=\linewidth]{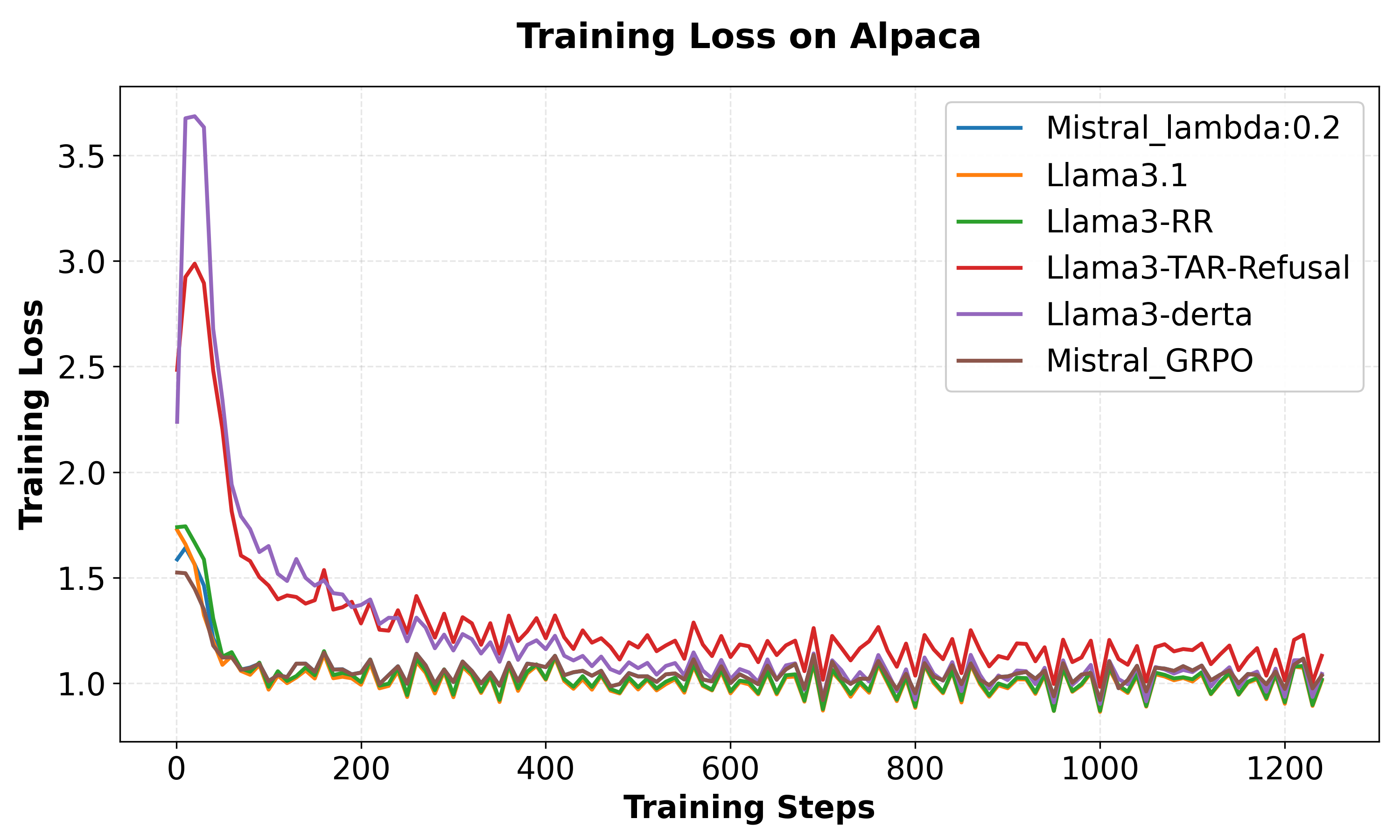}
    \end{subfigure}
    \hspace{0.02in}
    \begin{subfigure}{0.48\linewidth}
        \includegraphics[width=\linewidth]{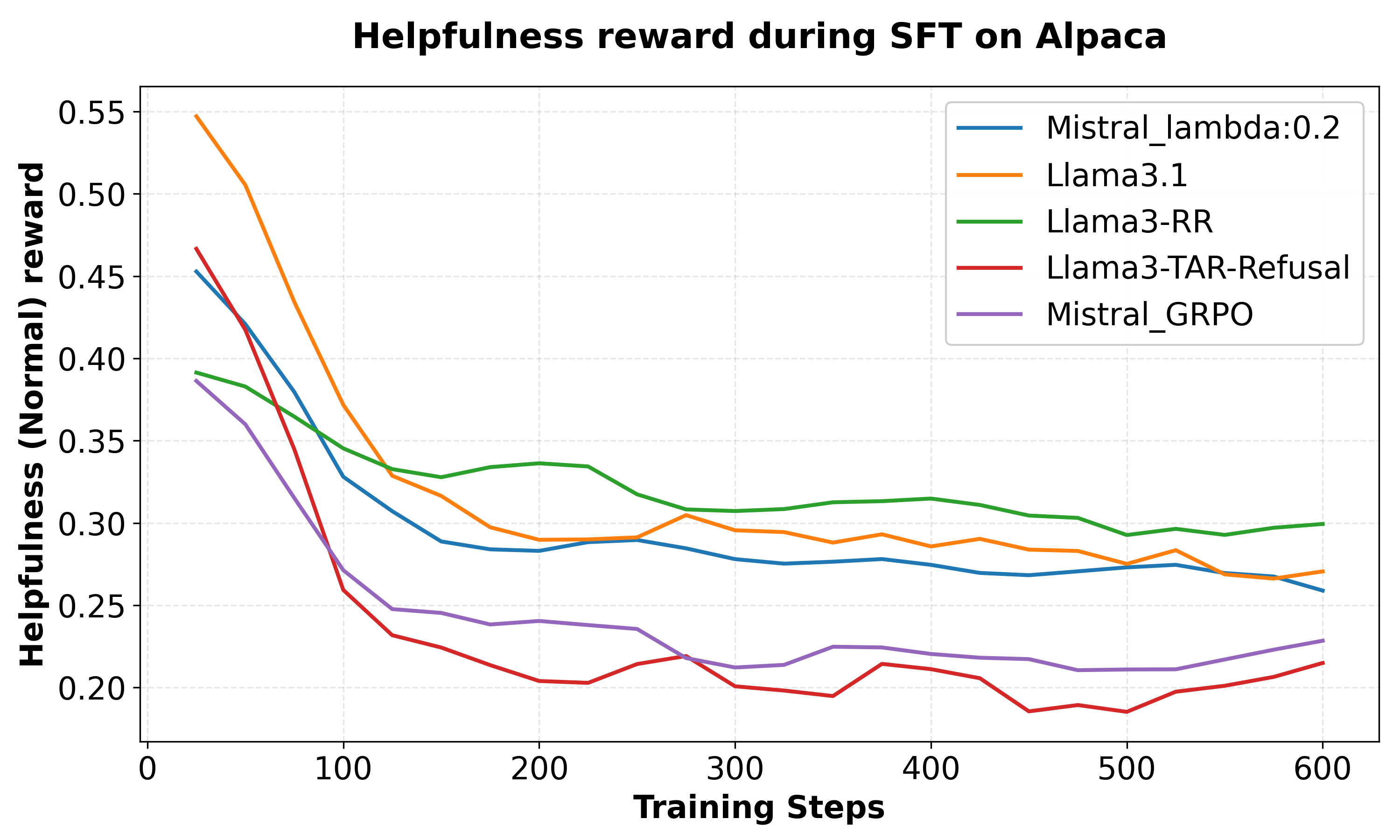}
    \end{subfigure}
     \caption{\textbf{(left)} We measure the SFT loss during fine-tuning to show that the FRPO-trained model fits the downstream task as well as other models. \textbf{(right)} FRPO also keeps the general capabilities higher that GRPO after fine-tuning.}
     \label{fig:alpaca_normal}
\end{figure}

We also track the helpfulness reward during SFT on Alpaca in \Cref{fig:alpaca_normal} (right). The results indicate that the general capabilities of the models, as measured by helpfulness score, follow similar trajectories. Notably, FRPO maintains a higher reward than GRPO by the end of fine-tuning. This suggests that the superior safety score is not an artifact of overfitting.

\subsection{Full Comparison on GSM8K}\label{sec:GSM8K_moreresults}
\begin{figure}[t]
    \centering
    \begin{subfigure}[b]{0.48\textwidth}
        \includegraphics[width=\linewidth]{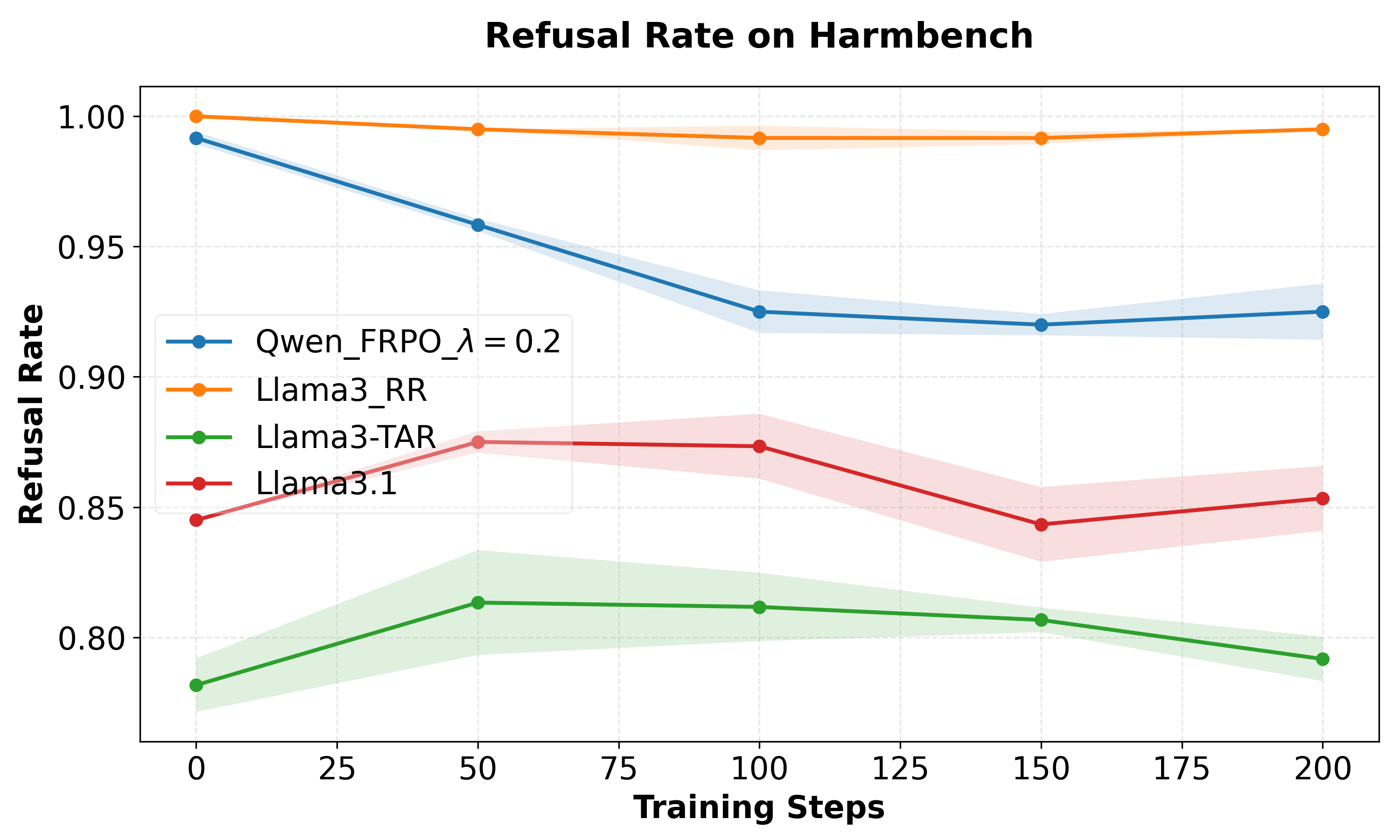}
    \end{subfigure}
    \hspace{0.02in}
    \begin{subfigure}{0.48\linewidth}
        \includegraphics[width=\linewidth]{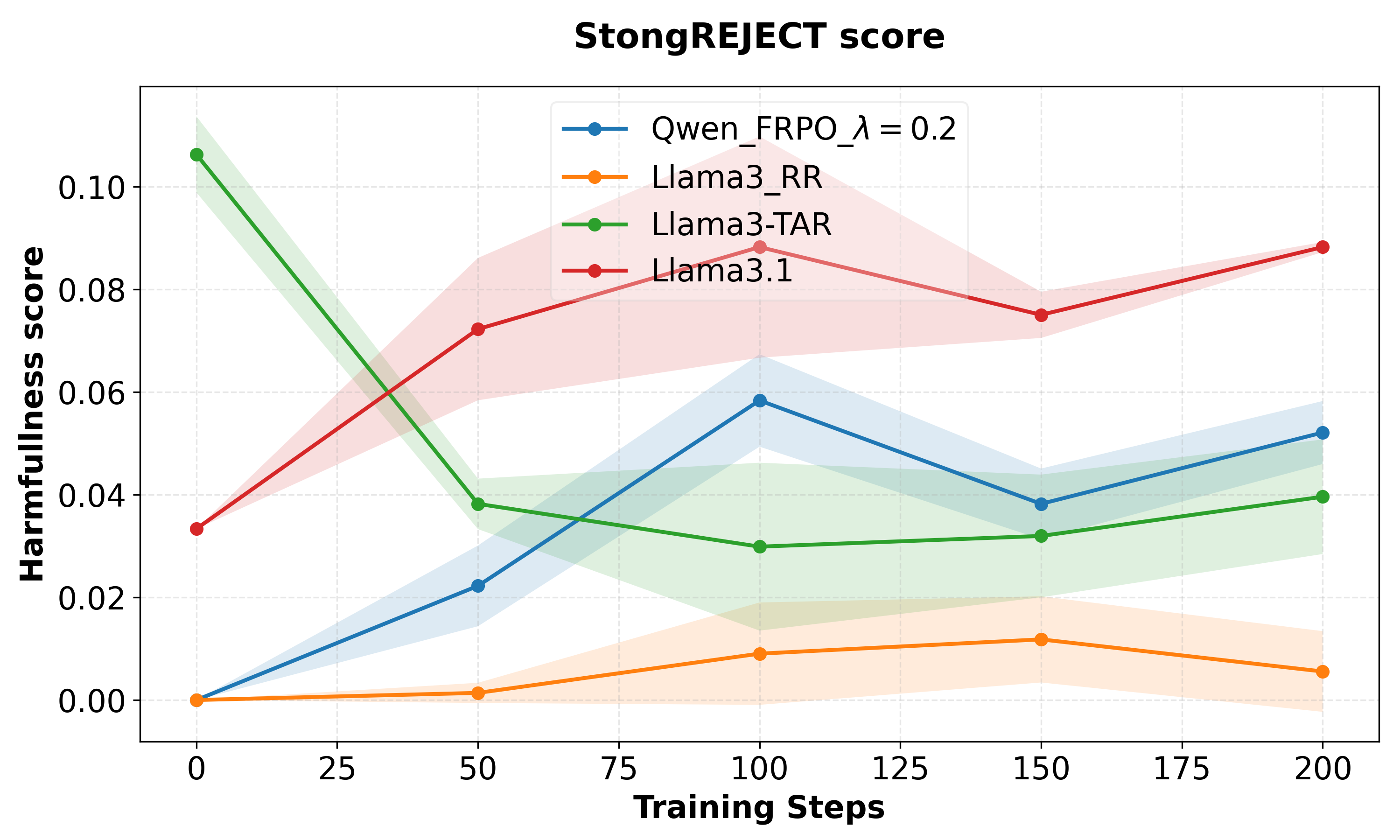}
    \end{subfigure}
     \caption{The refusal rate and the StrongREJECT score of the models during fine-tuning on GSM8K. Llama-3-TAR loses its general capabilities after fine-tuning, and thus the harmfulness score drops as well.}
     \label{fig:gsm8k_total}
\end{figure}

We compare the Qwen model trained with FRPO and $\lambda = 0.2$ against other baselines on HarmBench and StrongREJECT after fine-tuning on GSM8K (we already compared with Qwen and Mistral models in \Cref{sec:alpaca}). As \Cref{fig:gsm8k_total} shows, our model outperforms others except for Llama-3-RR \citep{zou2024improving}; this model has effectively unlearned unsafe responses and is not directly comparable to our robust RLHF method. Nevertheless, we show in \Cref{sec:alpaca} that the model trained with FRPO outperforms Llama-3-RR after fine-tuning on Alpaca. 

It must be noted that the drop in the StrongREJECT score of Llama-3-TAR is due to a significant drop in the helpfulness score (down to $\sim 0.1$). As mentioned earlier in \Cref{sec:normal_alpaca}, this model strictly resists fine-tuning, even if it comes at the cost of losing its general capabilities. In contrast, we show in \Cref{table:GSM8K_fine-tune} that our models maintain their general capabilities and improve on GSM8K.

\subsection{Details for RL on Helpfulness}\label{sec:app_helpfulness}
We consider RL fine-tuning and test if the robustness observed under SFT persists. Prior work shows that harmlessness and helpfulness often conflict \citep{bai2022training, tan2025equilibrate}. We study this tradeoff by fine-tuning on 15k UltraFeedback prompts \citep{cui2023ultrafeedback} with GRPO, using Llama-3.1-8B-Instruct-RM-RB2 \citep{malik2025rewardbench} as the reward model. We again sweep $\lambda$ and tune $\beta$ to keep all models at roughly the same KL distance from their base policy.

\begin{figure}
    \centering
    \begin{subfigure}{0.45\linewidth}
        \includegraphics[width=\linewidth]{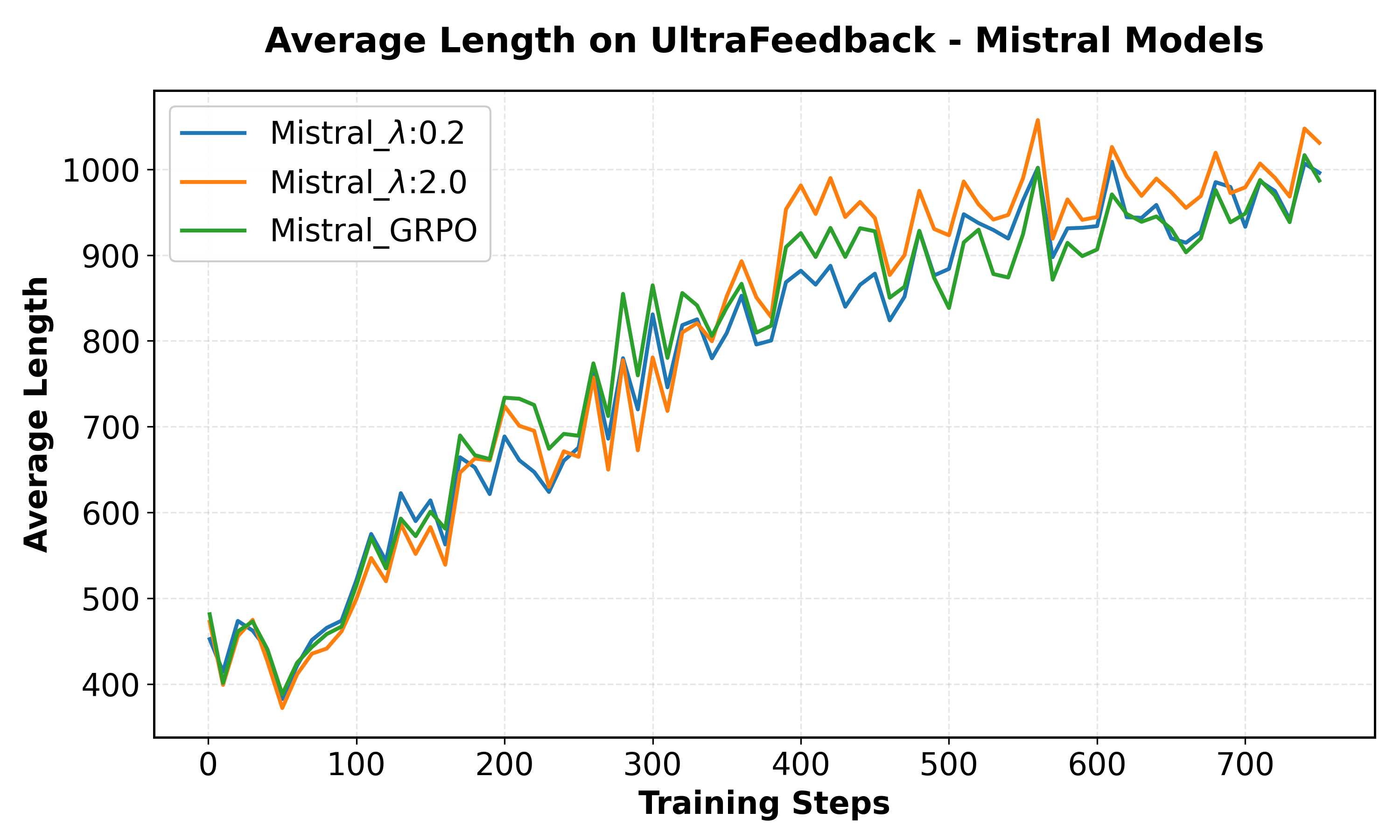}
    \end{subfigure}
    \hspace{0.03 \linewidth}
    \begin{subfigure}{0.36\linewidth}
        \includegraphics[width=\linewidth]{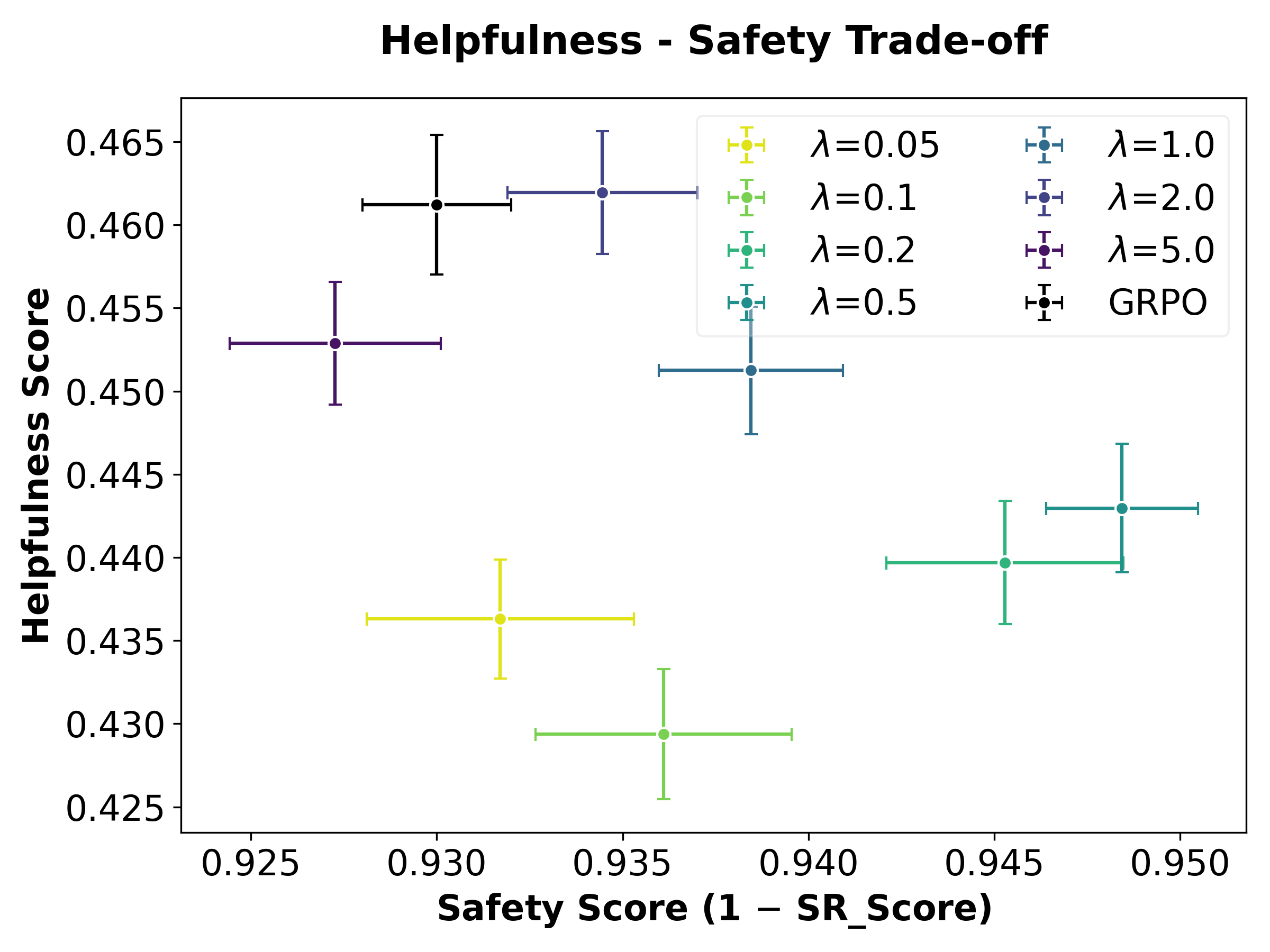}
    \end{subfigure}
    \hspace{0.01 \linewidth}
    \vspace{-6pt}
     \caption{\textbf{(left)} Fine-tuning the models on UltraFeedback with GRPO leads to a significant increase in the average response length, inducing more detailed answers to harmful demands. \textbf{(right)} Helpfulness vs. Safety score (1 $-$ StrongREJECT score) for Mistral models after GRPO on UltraFeedback. $\lambda = 0.5$ has better safety score but also lower helpfulness. $\lambda = 2.0$ and GRPO have the higher helpfulness score.}
     \vspace{-3pt}
     \label{fig:ultrafeedback}
\end{figure}

\vspace{-0.1in}
\paragraph{Helpfulness-safety trade-off under RL.} 
UltraFeedback consists of complex prompts (e.g., multi-step coding tasks) that require long responses. Accordingly, the average response length increases during training (see \Cref{fig:ultrafeedback}, left). This affects the responses to unsafe prompts, and models tend to generate more detailed harmful content. \Cref{fig:ultrafeedback} (right) shows StrongREJECT scores declining over training across all models (initial values shown in \Cref{fig:lambda}, left).  We observe a clear tradeoff: smaller $\lambda$ better preserves safety (StrongREJECT), while larger $\lambda$ slightly improves helpfulness at the cost of safety. Overall, GRPO and $\lambda \in \{2.0, 1.0, 0.5, 0.2\}$ lie on a Pareto frontier.
This behavior matches the intuition in \Cref{fig:intro}: within a specific KL ball, the best $\lambda$ yields a more reward-flat landscape, thus reducing drift toward helpfulness as the two objectives conflict.

\subsection{Comparison with Replay Methods}\label{sec:replay}
\begin{figure}
    \centering
    \begin{subfigure}[b]{0.48\textwidth}
        \includegraphics[width=\linewidth]{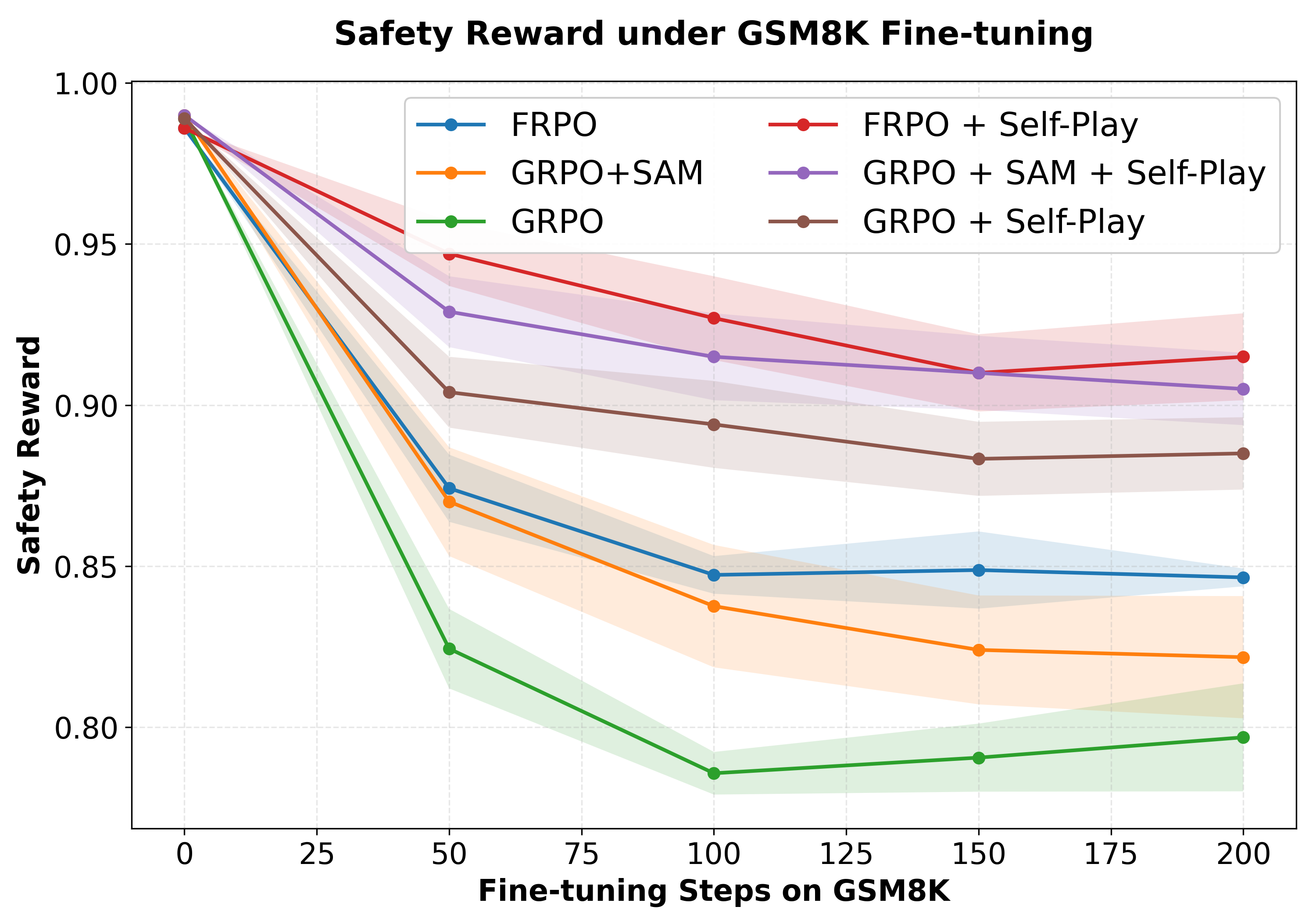}
    \end{subfigure}
     \caption{Self-play with a small portion of the safety training data helps to prevent catastrophic forgetting. FRPO + Self-play overperforms other methods.} 
     \label{fig:replay}
\end{figure}

We show that a self-play trick at downstream time can further help FRPO to prevent catastrophic forgetting. For a small portion of the safety training data (5\%) we generated new responses with the trained models, and added them to GSM8K supervised fine-tuning data. We then finetuned the modes with the same setting explained in \Cref{sec:alpaca}. \Cref{fig:replay} shows that FRPO + replay methods is more successful at preserving the safety rewards compared to both GRPO and GRPO +SAM, which was discussed in \Cref{sec:SAM}.

\subsection{High learning-rate SFT without LoRA Degrades the Capabilities}\label{sec:high_lr}
In this paper we posit that the KL constraint is implicitly satisfied in SFT when LoRA is deployed, or when the learning rate and training duration are moderate; we verified this empirically in \Cref{fig:alpaca} (right). Here, we show that in a high-learning-rate regime, even though safety degrades faster, general capabilities---measured by the general reward model ``Llama-3.1-8B-Instruct-RM-RB2'' \citep{malik2025rewardbench}---deteriorate significantly as well.

\begin{figure}[t]
    \centering
    \begin{subfigure}[b]{0.333\textwidth}
        \includegraphics[width=\linewidth]{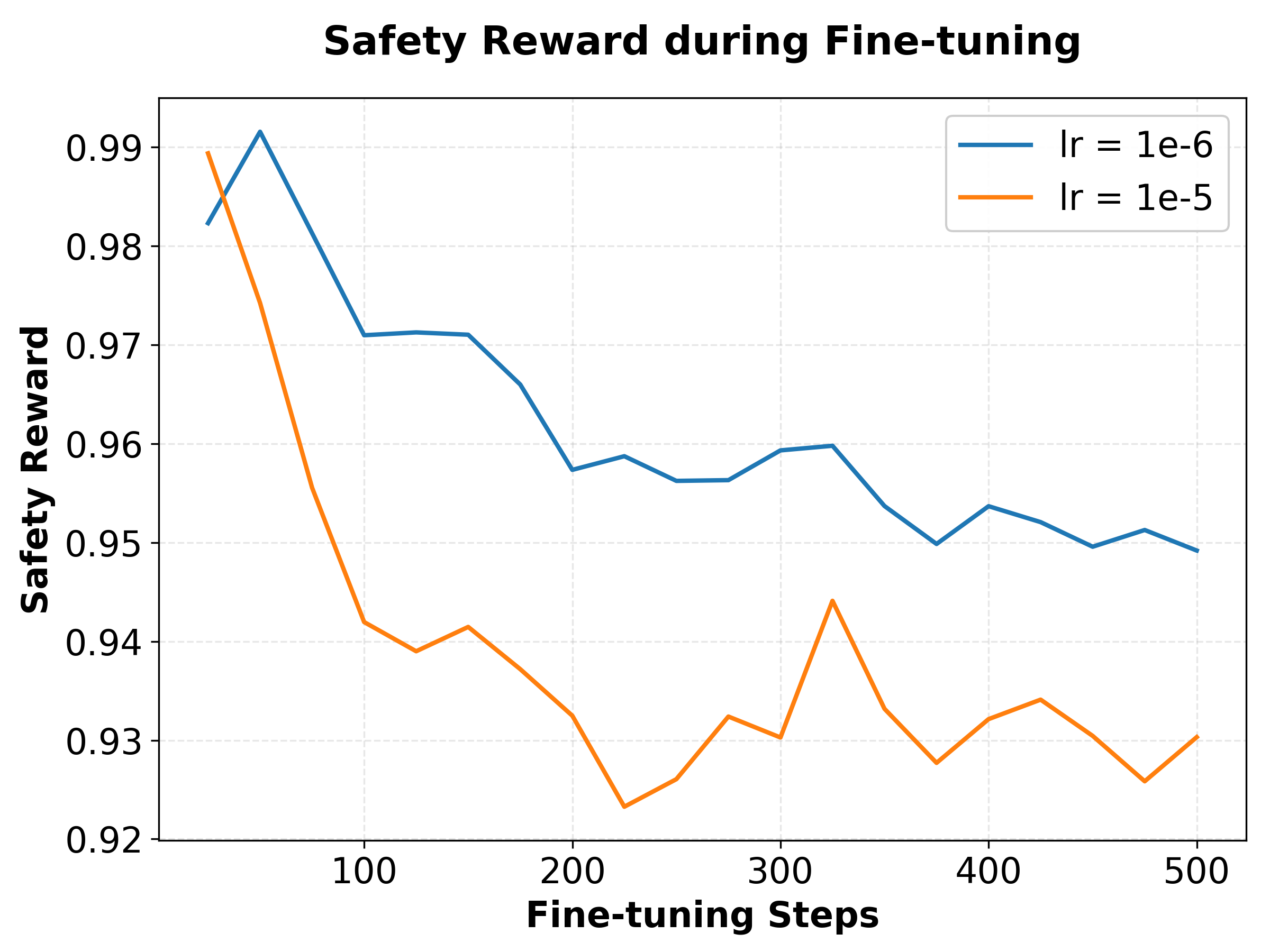}
    \end{subfigure}%
    \begin{subfigure}{0.333\linewidth}
        \includegraphics[width=\linewidth]{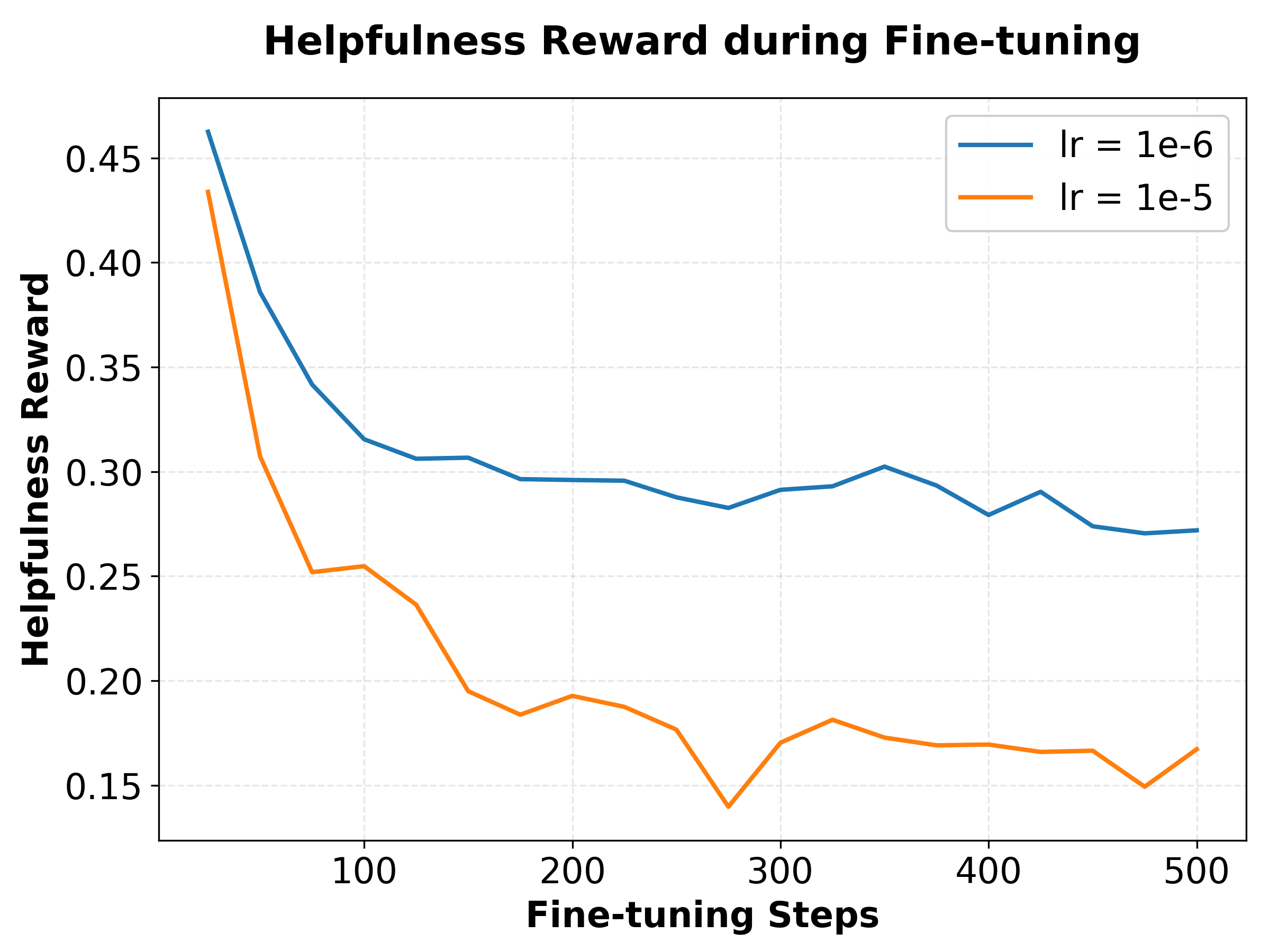}
    \end{subfigure}%
    \begin{subfigure}{0.333\linewidth}
        \includegraphics[width=\linewidth]{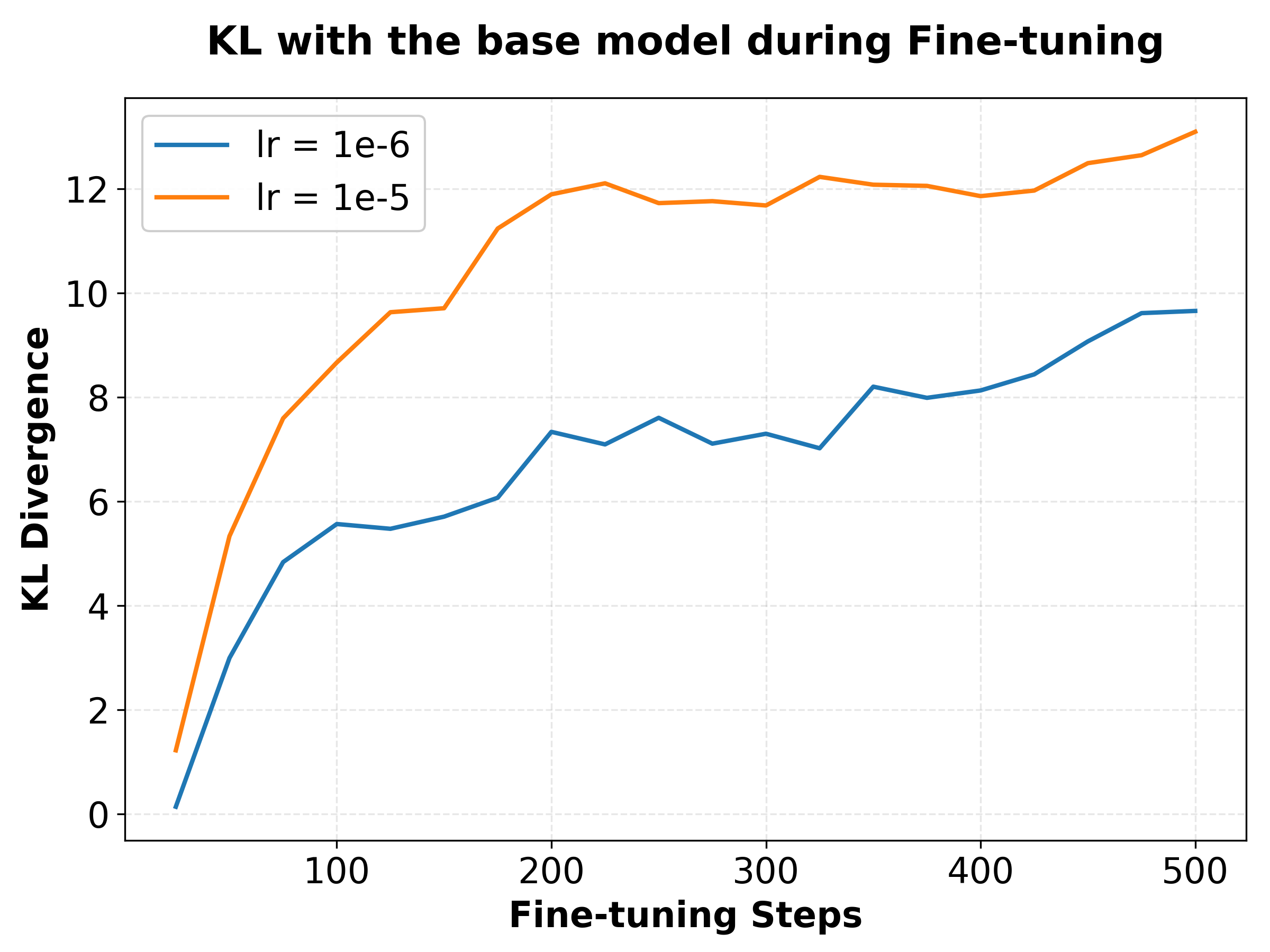}
    \end{subfigure}
     \caption{The model trained with FRPO and $\lambda =0.5$ fine-tuned on Alpaca; \textbf{(left/middle)} we consider a higher choice of learning rate for SFT on Alpaca (described in \Cref{sec:alpaca}) to show that it slightly degrades the safety but the helpfulness score is highly impacted. \textbf{(right)} This shows that a higher learning rate still changes the KL controllably but with higher slope, confirming our constraint in \Cref{sec:formulation}.}
     \label{fig:highlr}
\end{figure}

We repeat the SFT on Alpaca experiment for Mistral model as described in \Cref{sec:alpaca}, but with $\mathrm{lr = 1e-5}$ rather than $\mathrm{lr = 1e-6}$. \Cref{fig:highlr} (left) shows that the higher learning rate results in a slightly lower safety reward. This corresponds to a slightly higher KL with the base model in \Cref{fig:highlr} (right), a level that the lower learning rate would eventually reach given more training steps. However, as \Cref{fig:highlr} (middle) shows, the helpfulness score is highly compromised. This indicates that higher learning rates are more prone to overfitting in general, suggesting that our KL constraint remains applicable as long as the fine-tuning scheme avoids overfitting.

\section{Training Details}

\subsection{Safety Training}\label{app:training}
In order to fine-tune the models with our algorithm, we modified TRL's implementation of GRPO \citep{vonwerra2022trl}. We used 8xH200s for training our models. 

\vspace{-0.1in}
\paragraph{Dataset.} The safety dataset, adapted from \citep{kim2025reasoning}, contains 1000 harmful and 1000 harmless samples. Harmful prompts are from WildJailbreak, Aegis AI Content Safety Dataset 2.0, and RainbowTeaming jailbreaking prompts \citep{jiang2024wildteaming,ghosh2025aegis2,samvelyan2024rainbow}. Harmless prompts come from OR-Bench \citep{cui2024or} and are used to preserve general instruction-following behavior and mitigate over-refusal.

\vspace{-0.1in}
\paragraph{Reward models.} We use separate rewards for harmful vs.\ harmless subsets. For harmful prompts, the reward is
$ 1 - (s_1+s_2)/2,$
where $s_1, s_2$ are scores from OpenAI Moderation API and StrongREJECT judge \citep{markov2023holistic,souly2024strongreject}. 
For harmless prompts, we use an off-the-shelf helpfulness reward model: ``Llama-3.1-8B-Instruct-RM-RB2'' \citep{malik2025rewardbench}. This component is only added to avoid over-refusal, and our focus is not on improving helpfulness.

\vspace{-0.1in}
\paragraph{Hyper-parameters.}
To avoid overfitting in both GRPO and our algorithm, \underline{we use a relatively large $\beta$} and tune it to keep the ``per-token KL'' near $0.1$ for all models. This makes comparisons meaningful: each method is effectively searching for the best policy within a similar KL ball around the same reference model. We used group size $G = 8$ for all the experiments. As noted before, we compute the advantages by subtracting the group average reward in \Cref{eq:algorithm_GRPO} but do not divide them by the standard deviation. Instead, we ensure the rewards are scaled between 0 and 1---the safety reward is naturally between 0 and 1, the normal reward model's output is passed through a sigmoid function---to keep both the safety and normal signals relevant. We use LoRA \citep{hu2022lora} with $ r= 64$ and $\alpha = 64$ for safety training of all the models. The learning-rate is $ \mathrm{lr  = 10^{-5}}$ for the Mistral models and $ \mathrm{lr  = 3 \times 10^{-5}}$ for Qwen models. In order to keep the gradient norm consistent across values of $\lambda$, we omitted the $\lambda$ factor in \Cref{eq:algorithm_GRPO} and tuned $\beta$ to keep the final KL bounded, rather than changing the learning-rate for each $\lambda$. We used 2 epochs on 2000 samples of the training data for training the Mistral models and 3 epochs for the Qwen models. We found that the Qwen models need more steps for convergence as they begin with higher rewards and a smaller standard deviation, leading to smaller gradient signals. 

\begin{figure}[t]
  \centering
  \begin{subfigure}[c]{0.333\columnwidth}
    \centering
    \includegraphics[width=\linewidth]{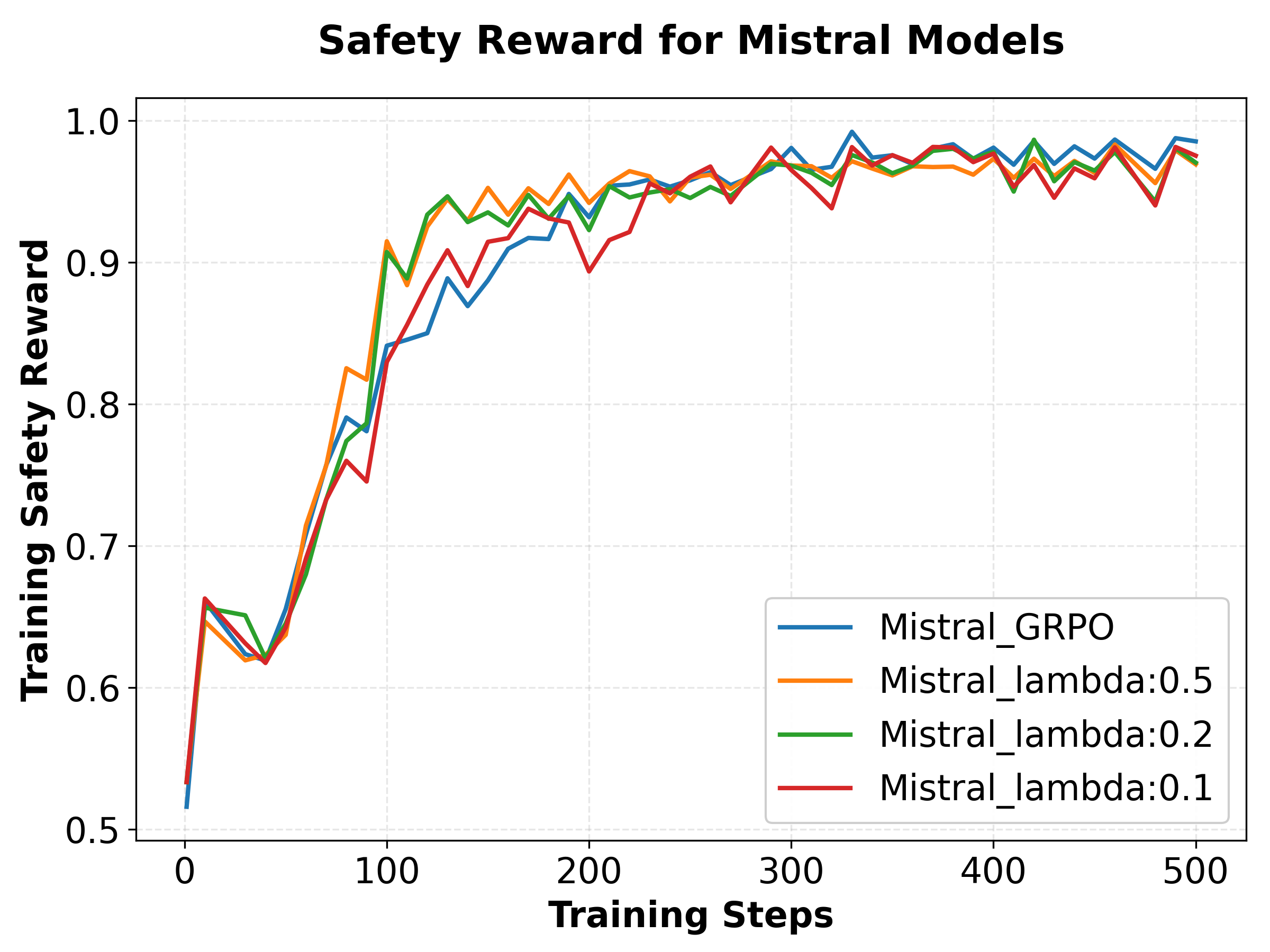}
    \caption{}
  \end{subfigure}%
  \begin{subfigure}[c]{0.333\columnwidth}
    \centering
    \includegraphics[width=\linewidth]{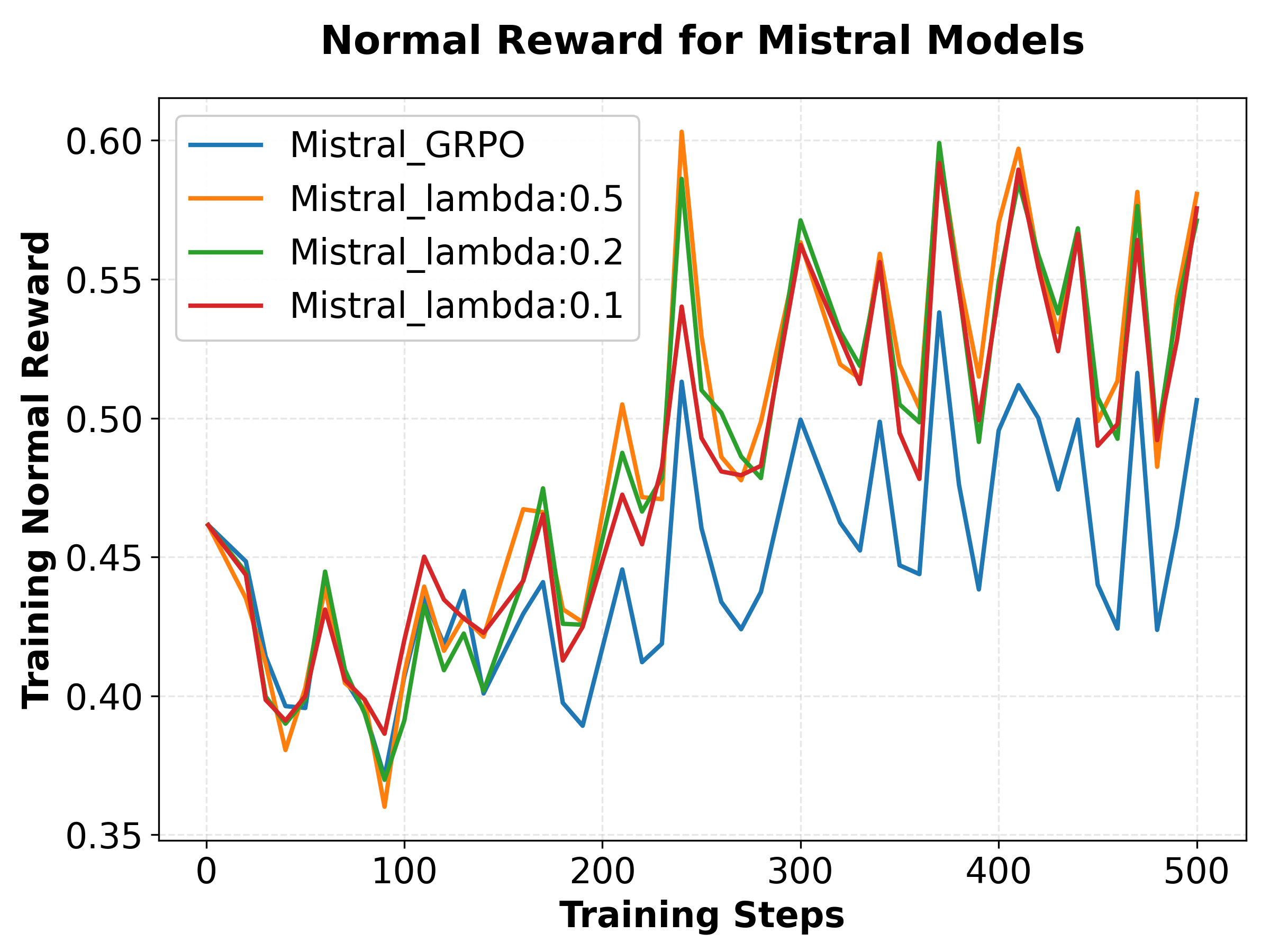}
    \caption{}
  \end{subfigure}%
  \begin{subfigure}[c]{0.333\columnwidth}
    \centering
    \includegraphics[width=\linewidth]{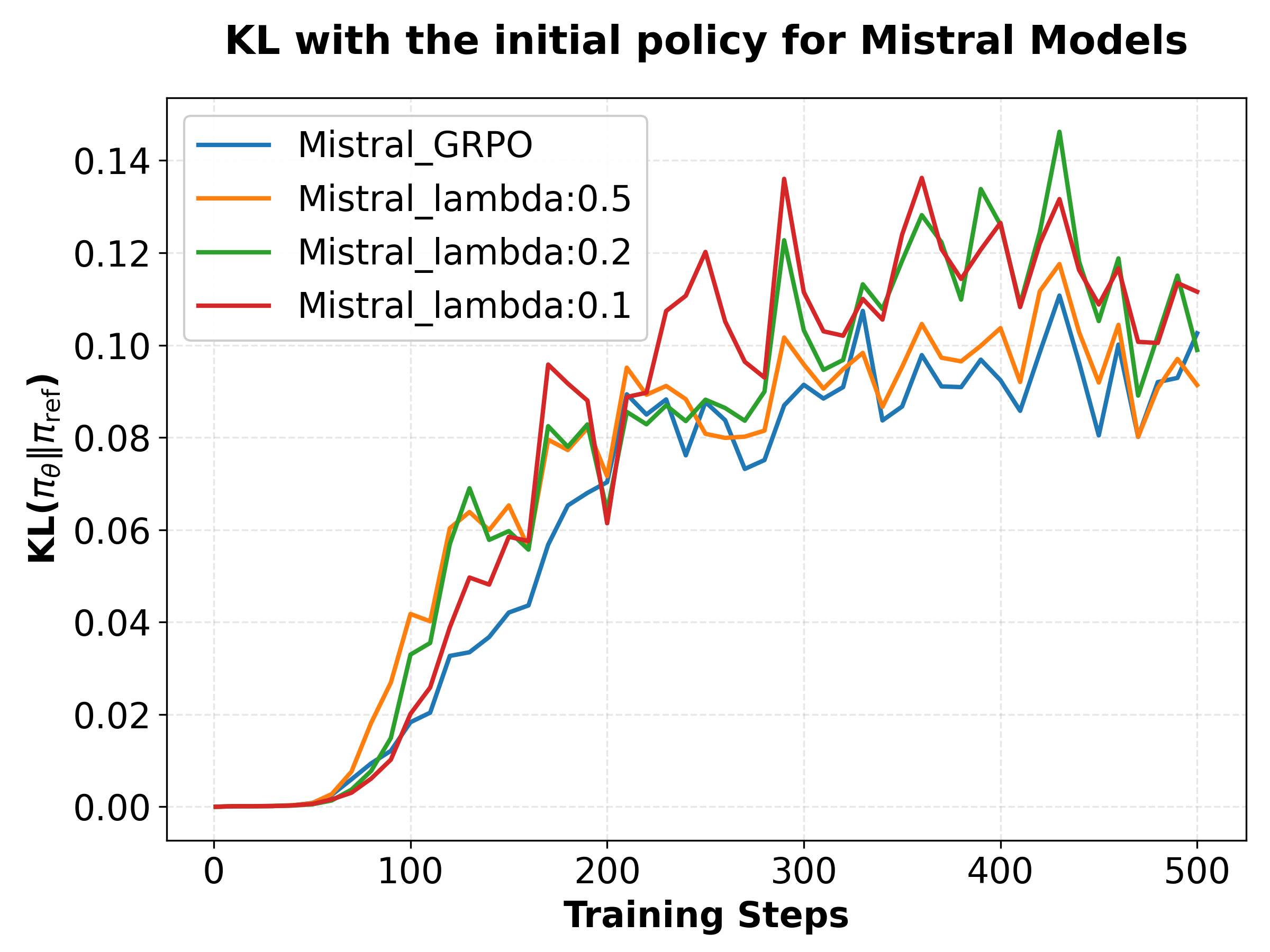}
    \caption{}
    
  \end{subfigure}
  \\
  \begin{subfigure}[c]{0.333\columnwidth}
    \centering
    \includegraphics[width=\linewidth]{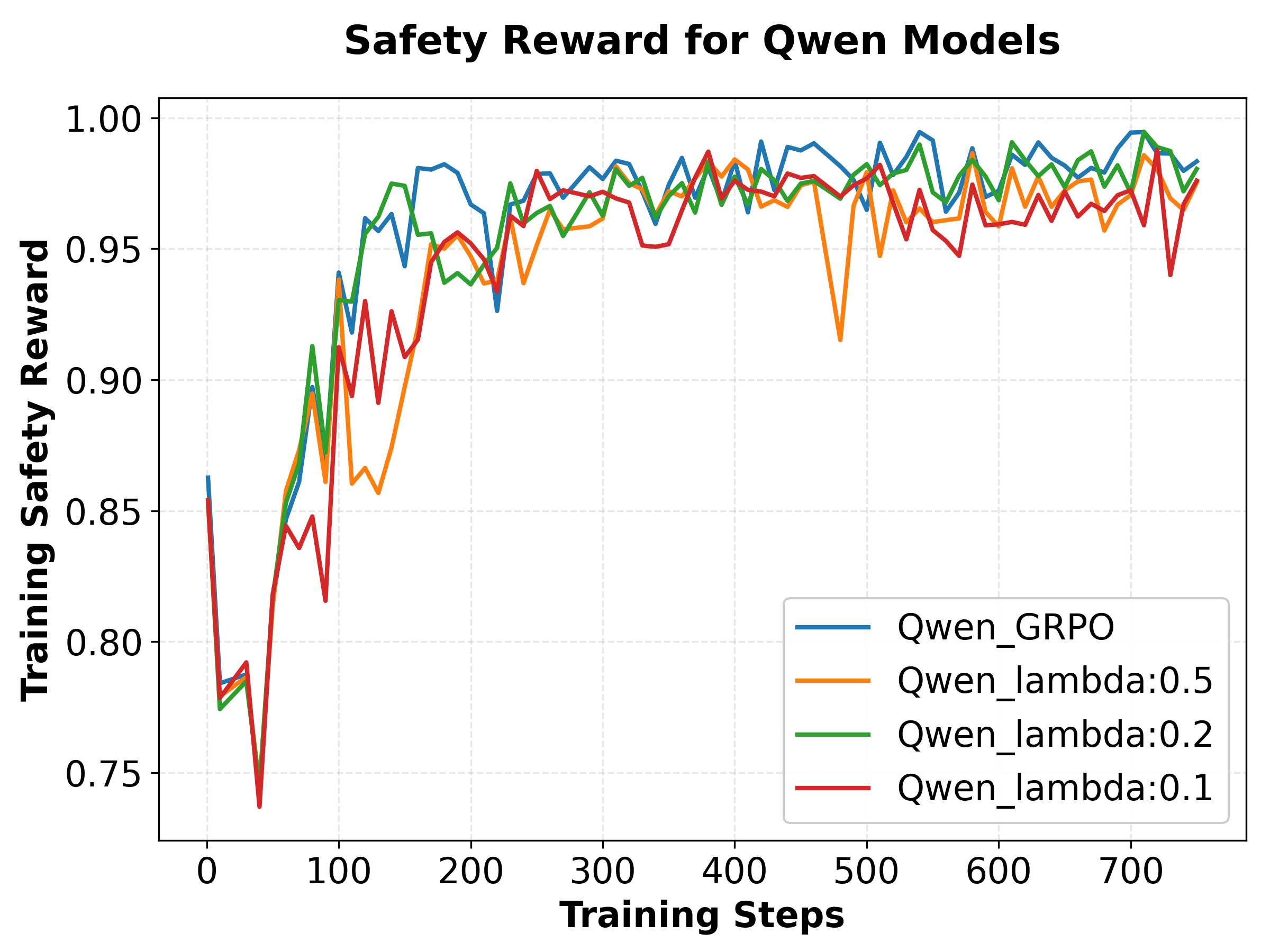}
    \caption{}
  \end{subfigure}%
  \begin{subfigure}[c]{0.333\columnwidth}
    \centering
    \includegraphics[width=\linewidth]{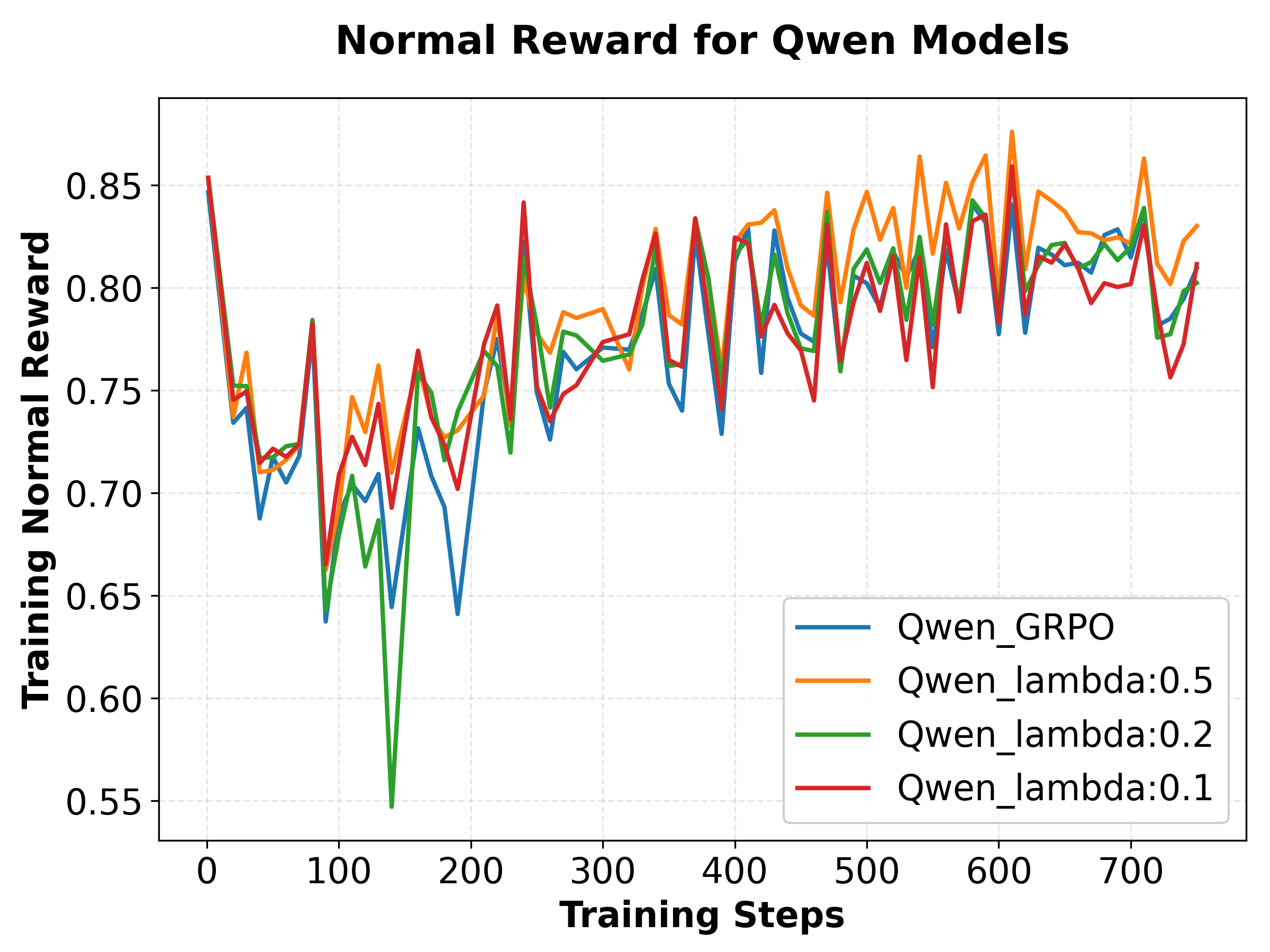}
    \caption{}
  \end{subfigure}%
  \begin{subfigure}[c]{0.333\columnwidth}
    \centering
    \includegraphics[width=\linewidth]{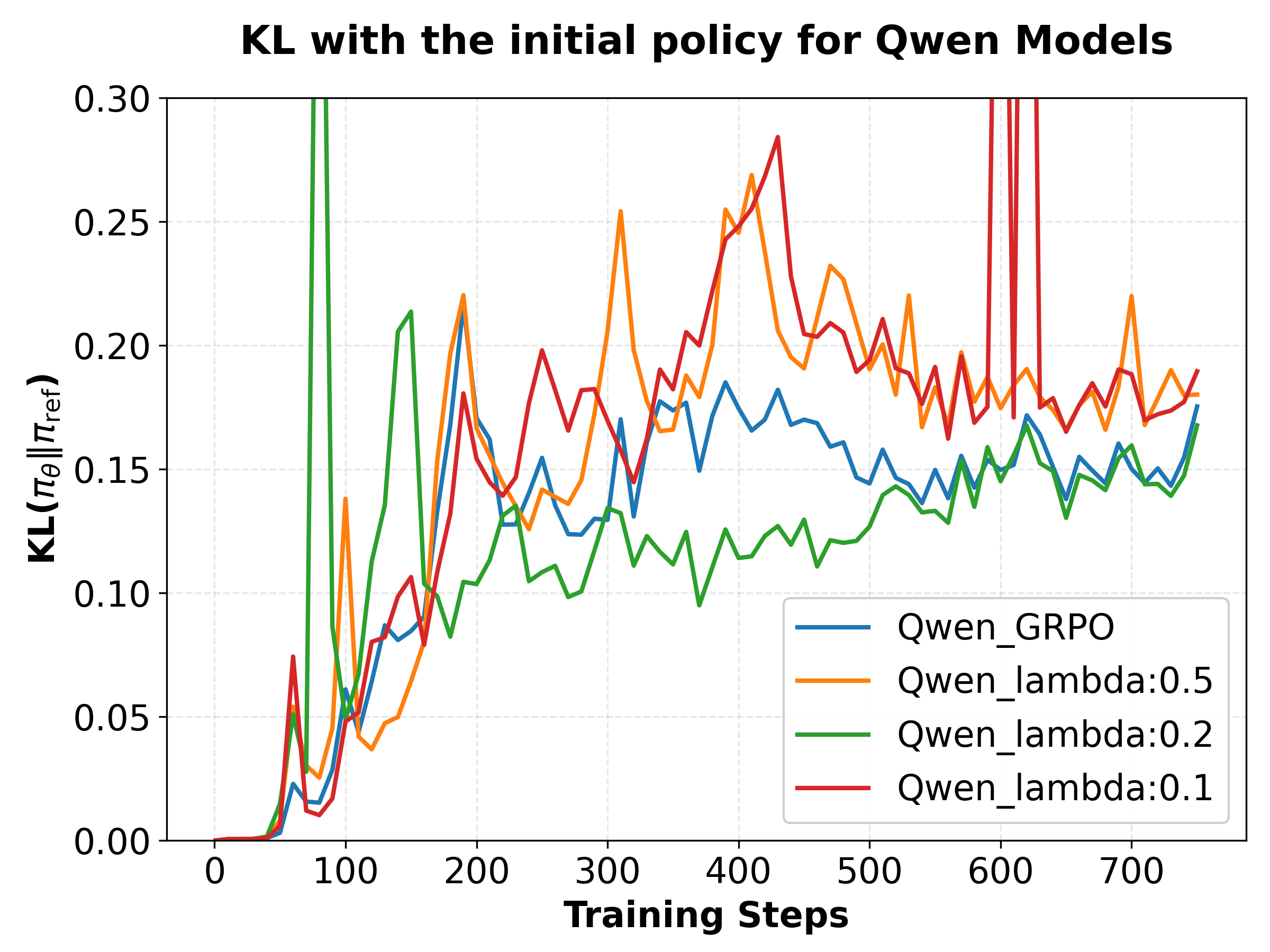}
    \caption{}
  \end{subfigure}
  \\
  \begin{subfigure}[c]{0.333\columnwidth}
    \centering
    \includegraphics[width=\linewidth]{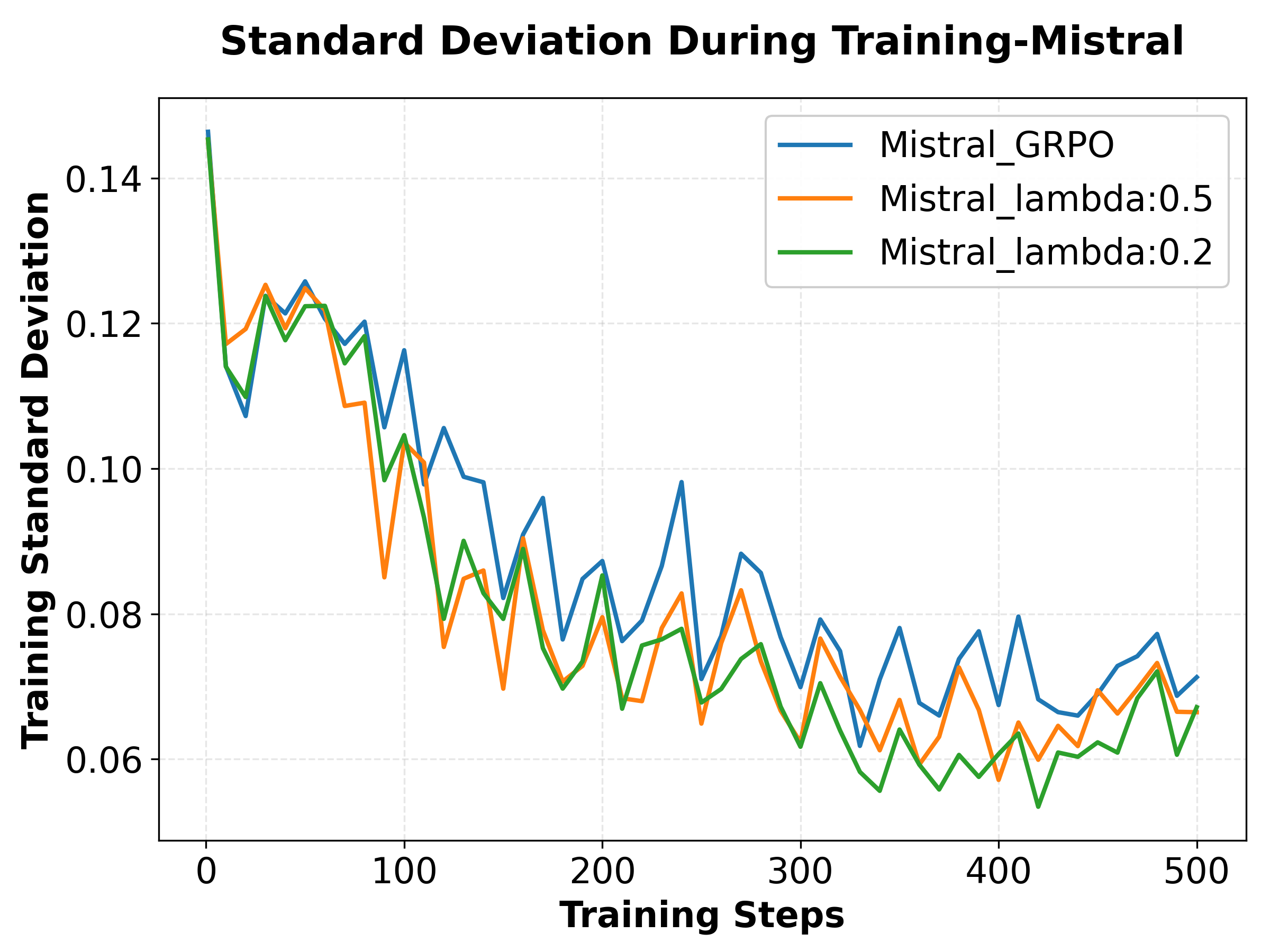}
    \caption{}
  \end{subfigure}%
  \hspace{0.2in}
  \begin{subfigure}[c]{0.333\columnwidth}
    \centering
    \includegraphics[width=\linewidth]{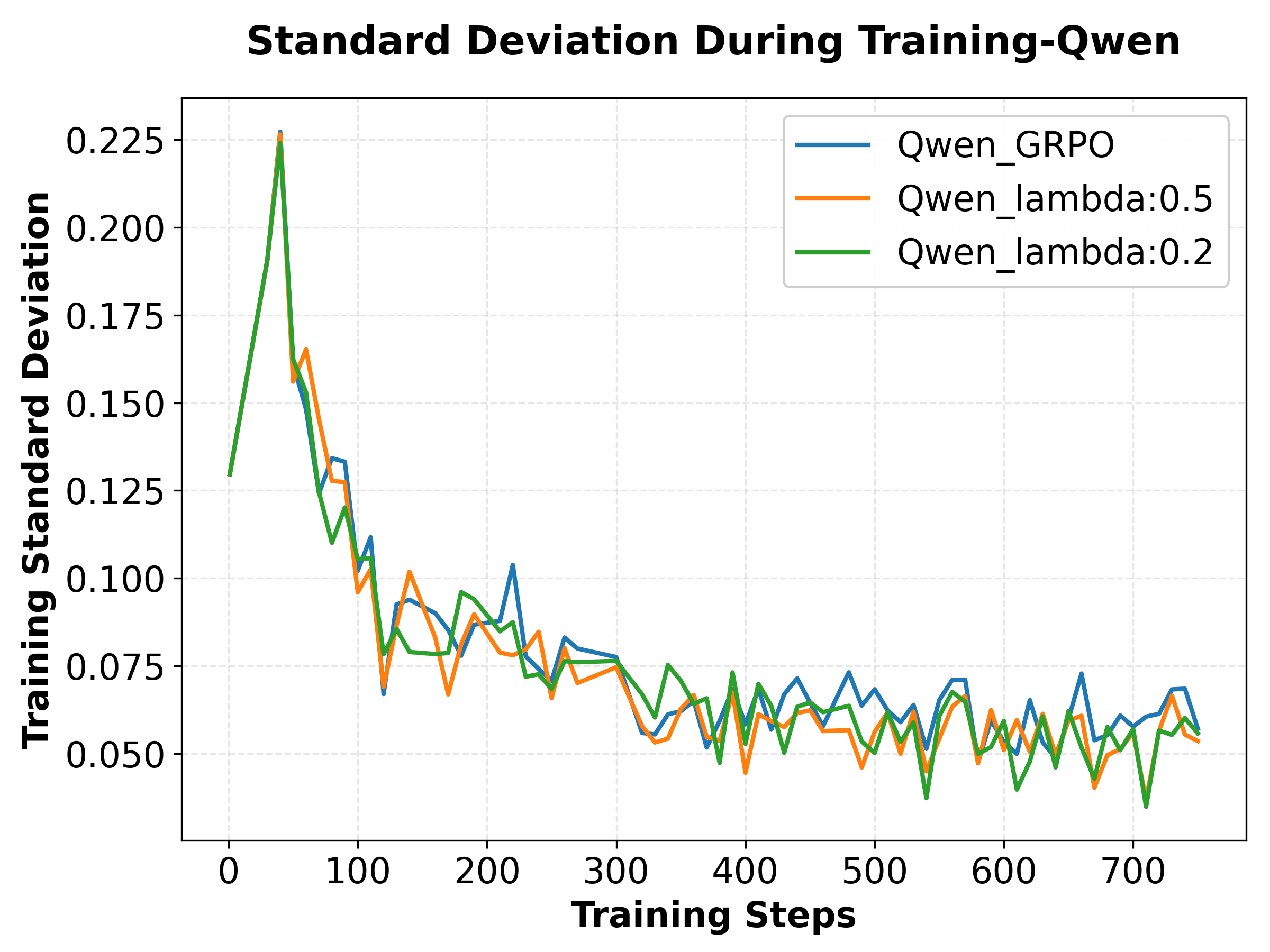}
    \caption{}
  \end{subfigure}%
  \caption{Safety training curves for Mistral and Qwen with GRPO and FRPO for $\lambda \in \{0.5, 0.2, 0.1\}$. \textbf{(a, d)} show that all Mistral and Qwen models converge in the safety score. \textbf{(b,e)} show that the Helpfulness (Normal) reward for the models slightly improve on OR-Bench, meaning that over-refusals are avoided. \textbf{(c, f)} We tune $\beta$ such that all models converge to roughly the same \textit{per-token-KL} to ensure a controlled comparison. \textbf{(g, h)} The standard deviation of the total reward for both Mistral and Qwen.}
    \label{fig:training}
\end{figure}

\vspace{-0.1in}
\paragraph{Results.} The training results are shown in \Cref{fig:training}, where the safety rewards converge for all the models with negligibly higher rewards for GRPO. As demonstrated in \Cref{fig:training}(b) and (e), the models trained with our algorithm achieve higher rewards on harmless prompts; as $\lambda$ decreases, our algorithm becomes more sensitive to a response with a low reward (near-zero reward for over-refusal) when the group average is high, leading to a large signal in \Cref{eq:algorithm_GRPO}. Finally, \Cref{fig:training}(c) and (f) show that we keep the KL to the reference models bounded after training. The values of the allowed KL are determined so that the normal reward and the policy's entropy are not affected.

\subsection{Math Training}\label{app:math_training}
As described in \Cref{sec:math}, we use a final answer verifier as the 0-1 reward for the RL training. We also add a small format reward if there is only one final answer within "\textbackslash boxed\{ \}". We average the two rewards with weights $[0.8, 0.2]$. The Qwen-Math system prompt is: \textit{"Please reason step by step, and put your final answer within \textbackslash boxed\{\}."}. We use $\mathrm{lr = 6 \times 10^{-6}}$ for both GRPO and FRPO with $\lambda \in \{10.0, 4.0, 2.0, 1.0\}$ and $\mathrm{lr = 5 \times 10^{-6}}$ for $\lambda \in \{0.5, 0.2, 0.1\}$ due to larger gradient norms for smaller $\lambda$. We choose a small $\beta = \mathrm{10^{-4}}$ for all the models, consistent with other math training settings \citep{liu2025understanding}. We use LoRA adapters with $r= 64$ and $\alpha = 128$. We train all the models for 3 epochs on the MATH training dataset (Levels 3-5).   

\begin{figure}
    \centering
    \begin{subfigure}{0.4\linewidth}
        \includegraphics[width=\linewidth]{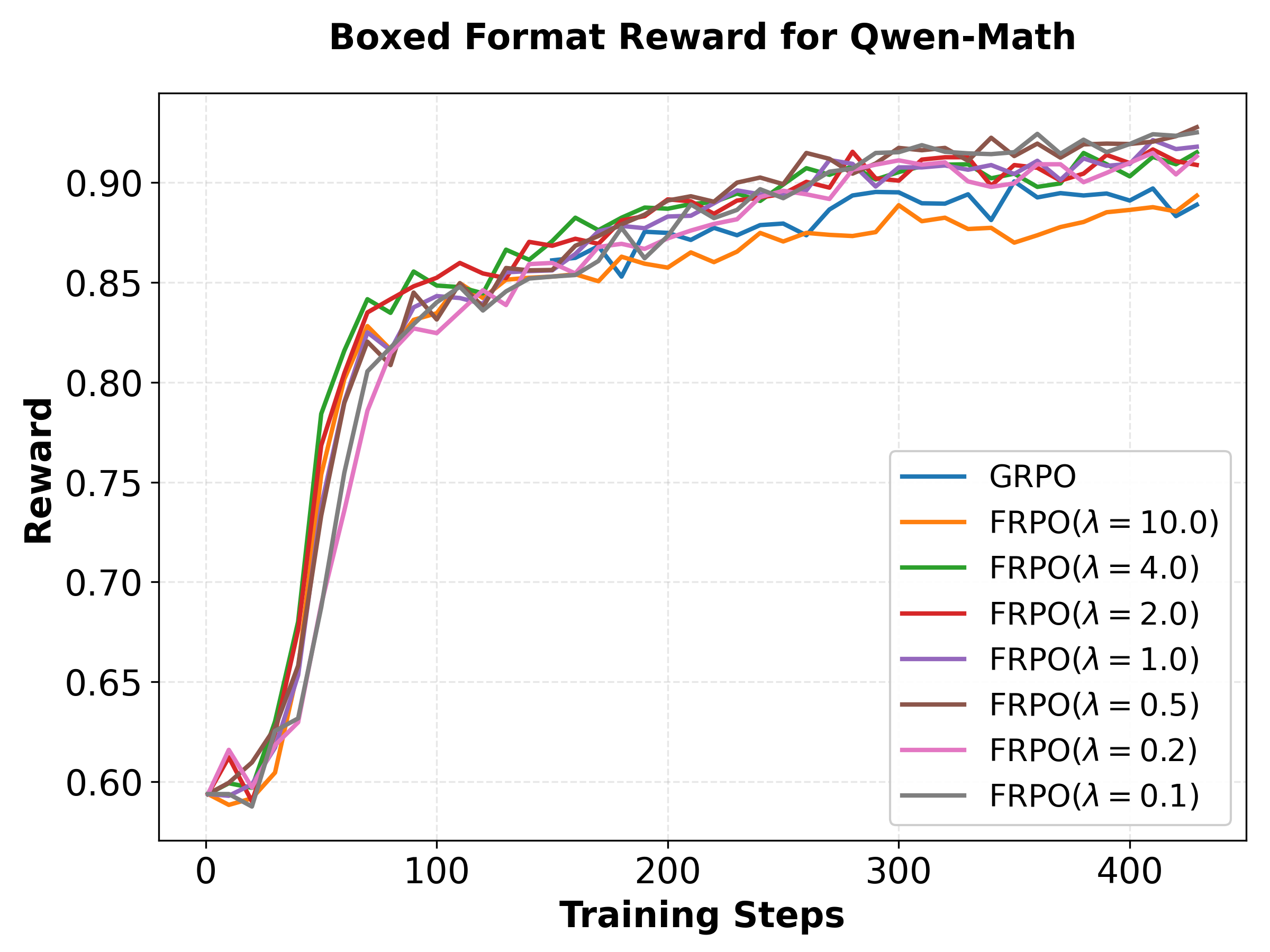}
    \end{subfigure}
    \hspace{0.03 \linewidth}
    \begin{subfigure}{0.4\linewidth}
        \includegraphics[width=\linewidth]{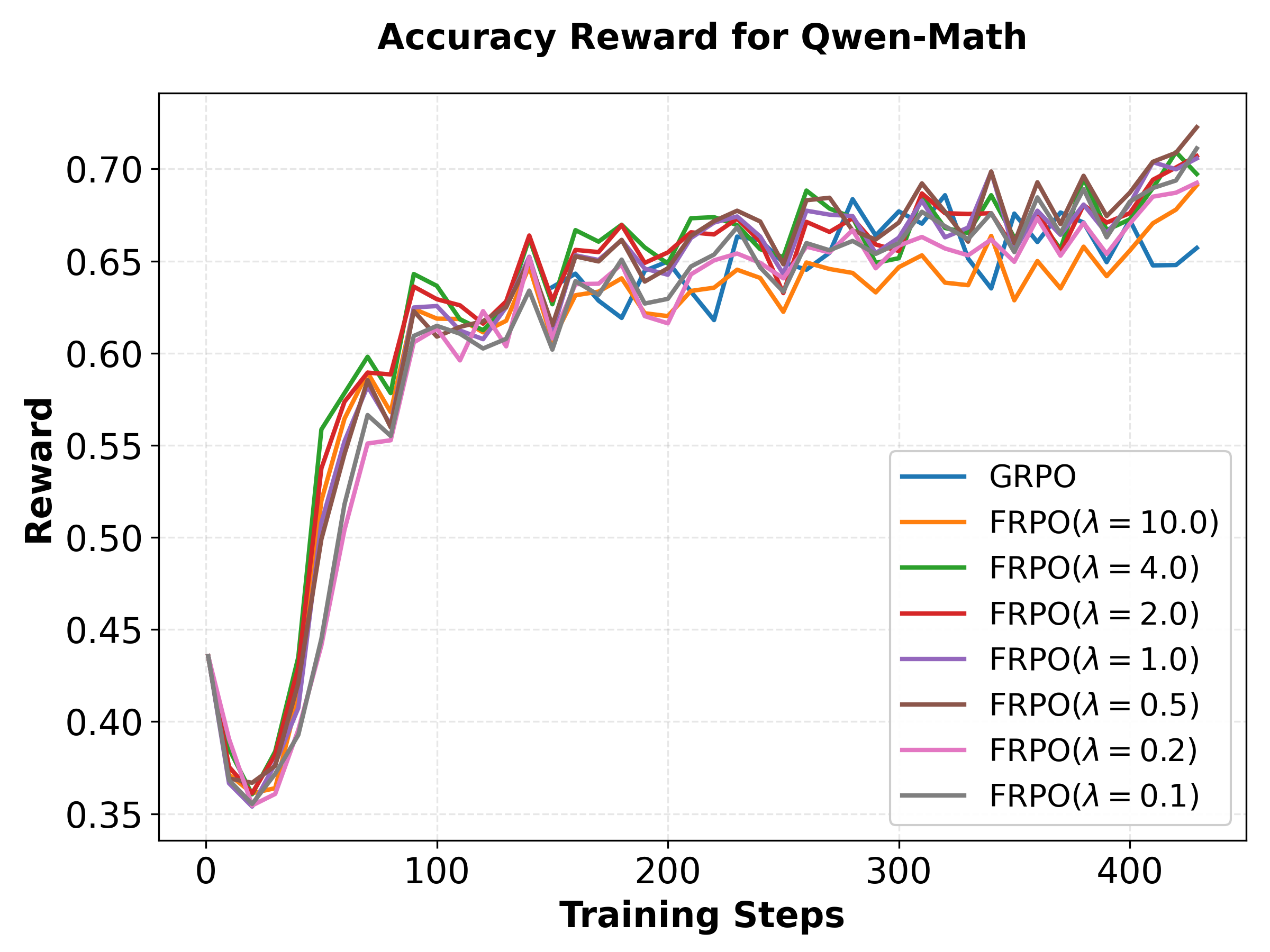}
    \end{subfigure}
     \caption{We have two rewards for math training: \textbf{(left)} shows that the format reward (whether the response contains any final answer) increases to $\sim$0.9 for all the training models;  \textbf{(right)} shows that the correctness reward increases similarly for all model, and roughly reaches 0.7.}
     \label{fig:math_training}
\end{figure}

\vspace{-0.1in}
\paragraph{Results.} The results of the training are presented in \Cref{fig:math_training}, where all the models roughly converge to the same point. \Cref{fig:math_training} (left) shows that the format reward converges to $\approx 0.9$ for all models. \Cref{fig:math_training} (right) shows that the initial training accuracy is 44\% for all the models, and all of them reach around 70\% by the end of training. The first row of \Cref{table:math_ft} shows the final accuracy on MATH500 for all the models.

\subsection{Ablation: Effect of Training KL Budget on Downstream Robustness}\label{ablation:higher_KL}

\begin{figure}
    \centering
    \begin{subfigure}{0.4\linewidth}
        \includegraphics[width=\linewidth]{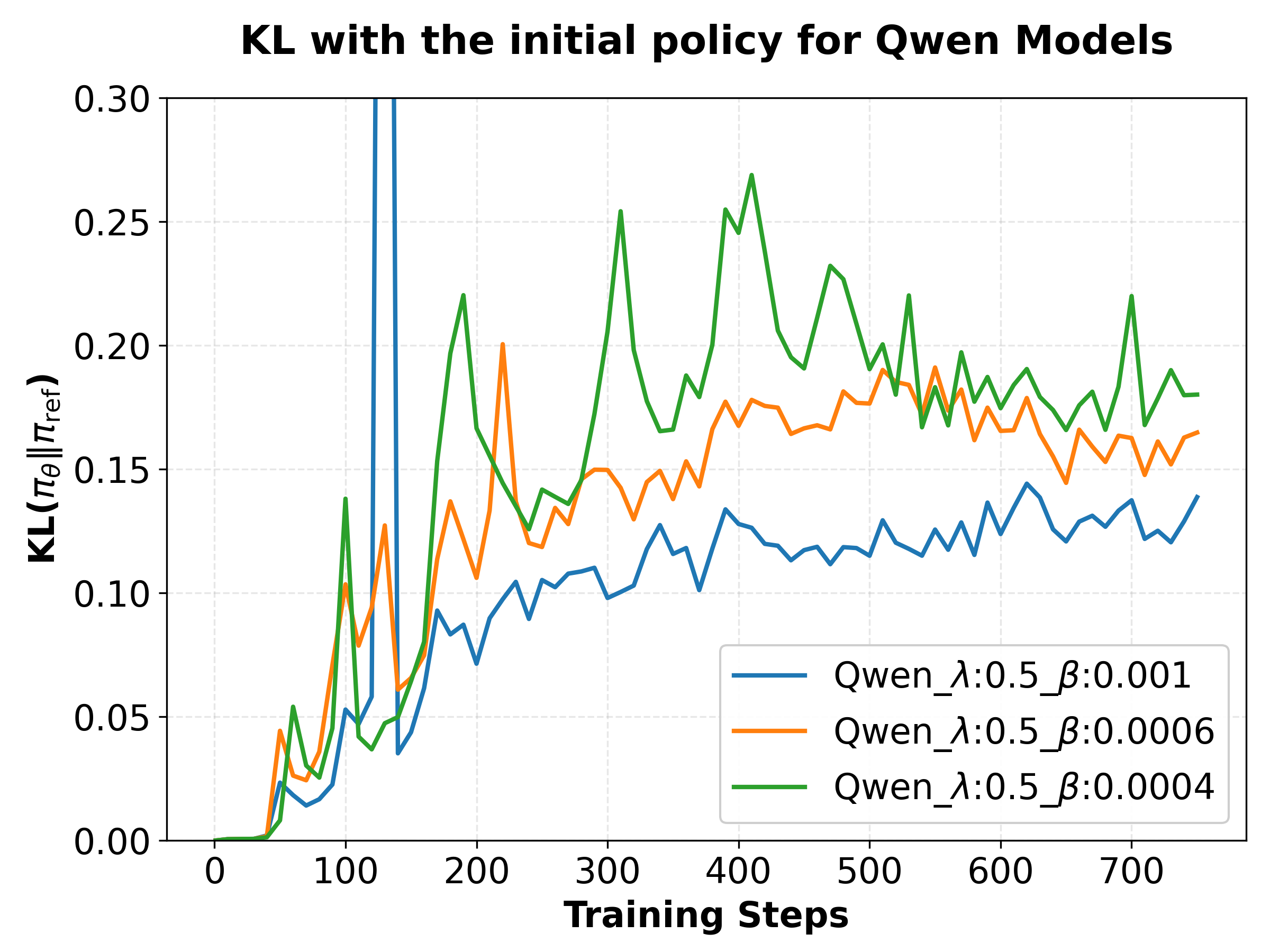}
    \end{subfigure}
    \hspace{0.03 \linewidth}
    \begin{subfigure}{0.4\linewidth}
        \includegraphics[width=\linewidth]{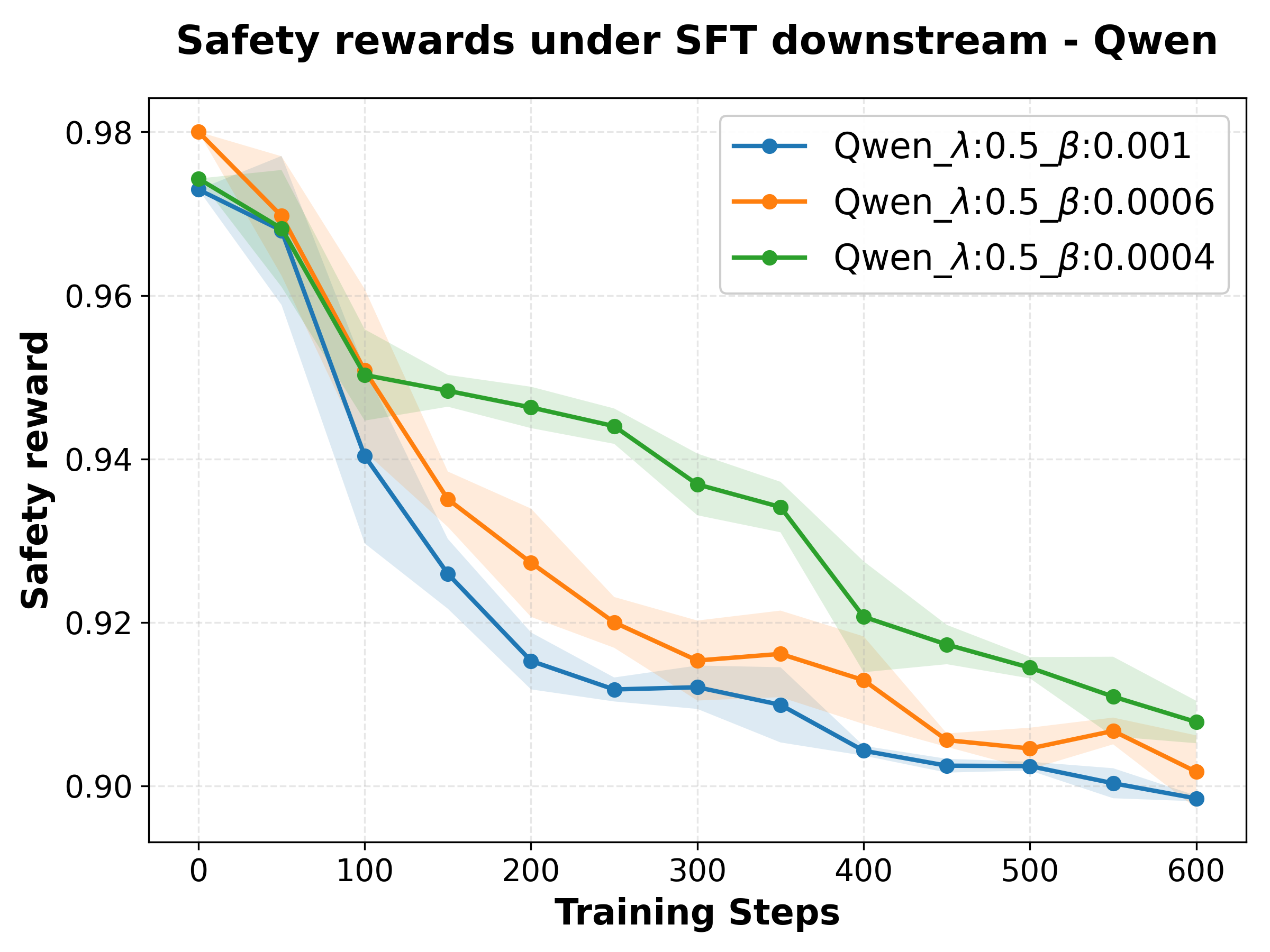}
    \end{subfigure}
     \caption{During training, among KL values that preserve the helpfulness reward and avoid reducing the policy entropy, using a larger KL (i.e., smaller $\beta$) yields a more robust solution with flatter rewards.}
     \label{fig:ablation_KL}
\end{figure}

During safety training, we tune $\beta$ to control the KL divergence to the reference model. Specifically, $\beta$ is chosen such that the resulting KL preserves the helpfulness reward while maintaining roughly constant policy entropy (noting that smaller $\beta$ typically reduces output entropy). Among such KL values below a given threshold, \Cref{fig:ablation_KL} shows that larger KL (smaller $\beta$) yields policies with higher safety reward after downstream fine-tuning. This occurs because \Cref{eq:final_opt} defines optimization over a KL ball around \(\pi_{\mathrm{ref}}\) whose radius shrinks with larger \(\beta\); a smaller ball overly constrains the reward-flatness objective in the first term \(-\E_{x\sim p}\!\left[\lambda\log\!\left(\E_{y \sim \pi_\theta(\cdot\mid x)}e^{-r(x,y)/\lambda}\right)\right]\), hindering its optimization.

\section{Proof of Lemma \ref{lem:optQ_fdiv}}\label{app:lemma_proof}
Denoting the likelihood ratio of $Q$ over $\pi_\theta$ as $ L(y\mid x)\ :=\ \frac{Q(y\mid x)}{\pi_\theta(y\mid x)}$, we can rewrite the KL term as:
$$\KL\!\big(Q(\cdot\mid x)\,\big\|\,\pi(\cdot\mid x)\big) = \int f \big( \frac{Q(y\mid x)}{\pi(y\mid x)} \big ) \pi(dy\mid x) = \E_{y\sim \pi(\cdot\mid x)}\big[f(L(y\mid x))\big]$$
where $f(x) = x \log (x)$. Moreover, using the Radon–Nikodym derivative,  we parametrize the Lagrangian for solving the infimum over $Q$ as a function of $L(y \mid x)$, a global multiplier $\lambda\geq 0$ for the average KL constraint and a
per-$x$ multiplier $\eta(x)$ for normalization ($\E_{\pi}[L\mid x]=1$):
\begin{align*}
\calL(L,\lambda,\eta)
=&\E_{x\sim p}\E_{y\sim\pi_\theta(\cdot\mid x)}
\big[\, L\,r(x,y) \big] +\lambda \big( \E_{x\sim p}\E_{y\sim\pi_\theta(\cdot\mid x)}[f(L(y \mid x)] - \rho \big) \\
& + \int \eta(x)\big(L (y \mid x) -1\big) \pi(dy \mid x) dx \\ 
\end{align*}
If we redefine $\eta(x) \leftarrow \frac{\eta(x)}{p(x)}$ as the new multiplier: 
\begin{align*}
\calL(L,\lambda,\eta)
&=\E_{x\sim p}\E_{y\sim\pi_\theta(\cdot\mid x)}
\Big[\, L\,r(x,y)+\lambda f(L)+\eta(x)\big(L-1\big)\,\Big]-\lambda\rho
\end{align*}
It is easy to see that the primal problem is convex since $f(x)$ is a convex function. Plus, Slater's condition hold because $\rho >0$ and $Q = \pi_\theta$ gives $\E[KL(Q\|\pi_\theta] = 0 < \rho$. Therefore, strictly feasible point exists, which results in strong duality $\min_L \max_{\lambda \geq 0, \eta} \calL = \max_{\lambda \geq 0, \eta} \min_L \calL$. Thus, minimizing over $L$ in terms of the Fenchel conjugate function gives:
\begin{align}
\min_L \calL =& \min_L \Big \{ \E_{x\sim p}\E_{y\sim\pi_\theta(\cdot\mid x)}
\Big[\, L\,r(x,y)+\lambda f(L)+\eta(x)\big(L \big)\,\Big] \Big \} - \E \eta(x) - \lambda\rho \nonumber\\
=& \E_{x\sim p}\E_{y\sim\pi_\theta(\cdot\mid x)}
\Big[\, \min_L \Big \{ (r(x,y) + \eta(x))\, L +\lambda f(L) \Big \} \Big]  - \E \eta(x) - \lambda\rho \nonumber\\
=& \E_{x\sim p}\E_{y\sim\pi_\theta(\cdot\mid x)}
\Big[ -\lambda \,f^*\big(\frac{-r(x,y) - \eta(x)}{\lambda} \big) - \eta(x) \Big] - \lambda\rho 
\end{align}\label{eq:afterL}

\section{Gradient of the Objective}\label{app:baseline}
\begin{figure}
    \centering
    \begin{subfigure}{0.4\linewidth}
        \includegraphics[width=\linewidth]{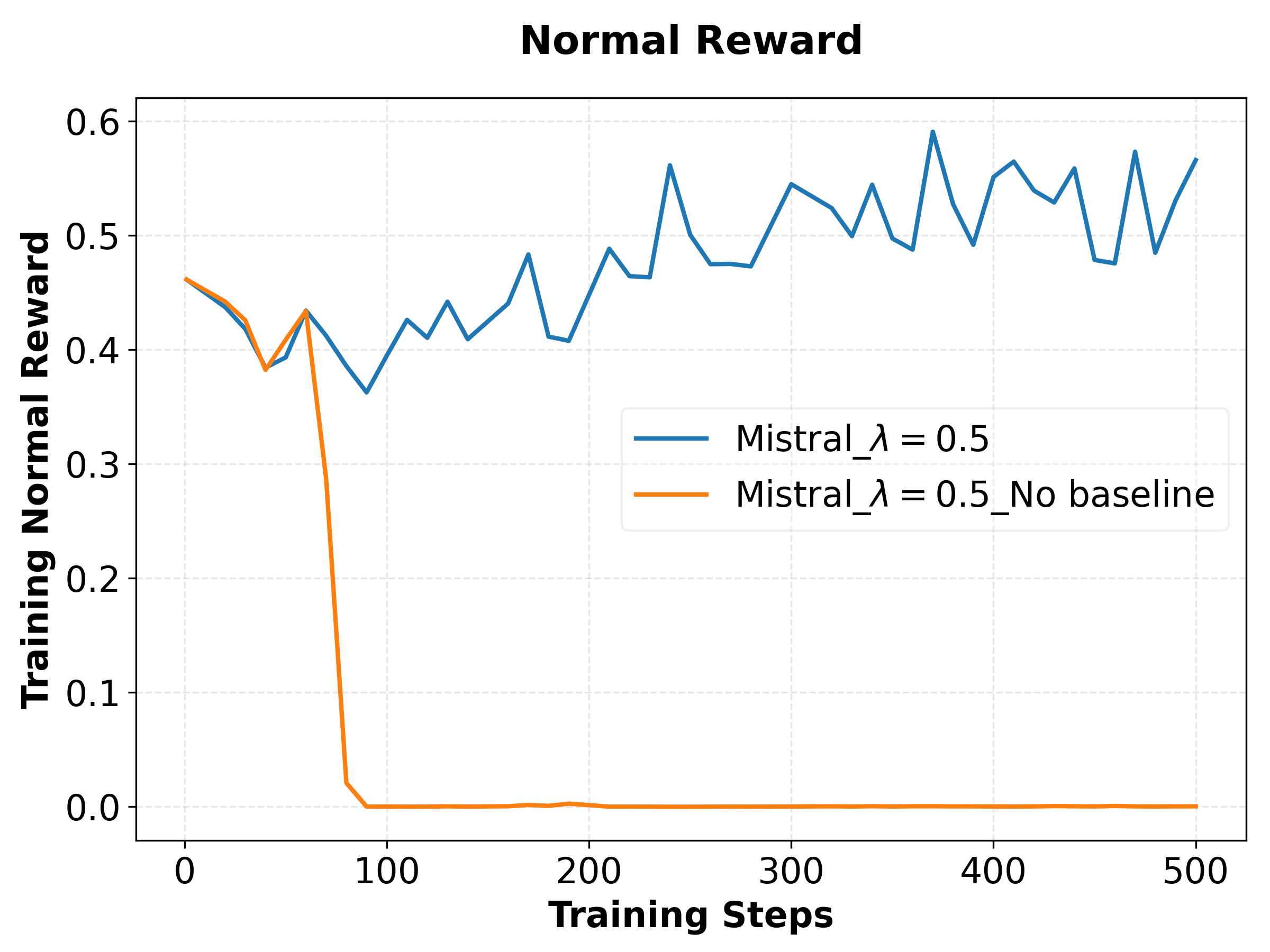}
    \end{subfigure}
    \hspace{0.03 \linewidth}
    \begin{subfigure}{0.4\linewidth}
        \includegraphics[width=\linewidth]{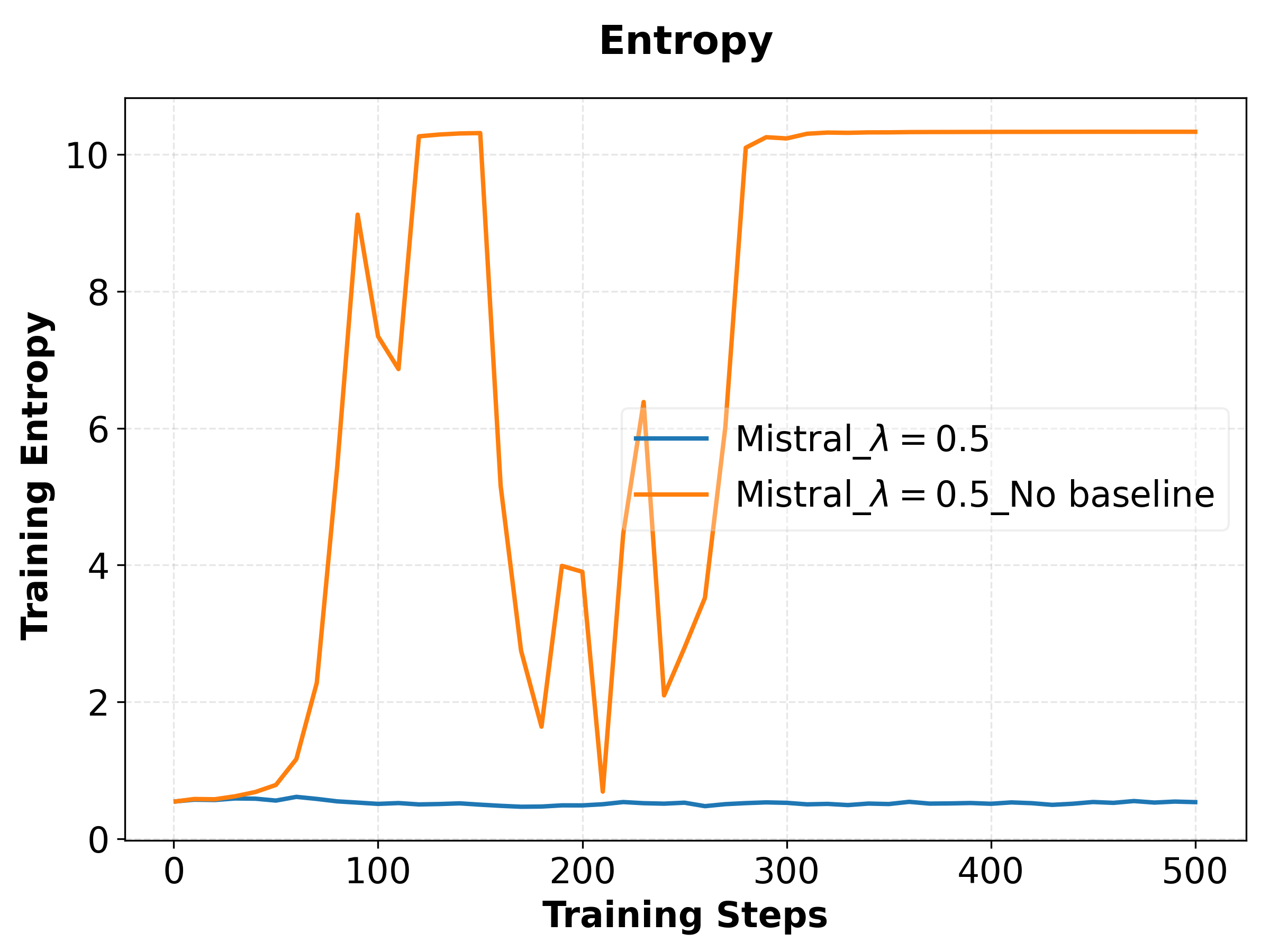}
    \end{subfigure}
     \caption{The policy collapses in the absence of the derived baseline in \Cref{eq:low_var} for FRPO; i.e., the token distribution becomes random and the policy entropy explodes, and the helpfulness (Normal) reward collapses as the responses become gibberish.}
     \label{fig:nooffset}
\end{figure}

\Cref{fig:nooffset} shows that omitting the baseline in FRPO causes the policy to collapse, resulting in a completely random token distribution (maximum entropy in \Cref{fig:nooffset}, right). This occurs because without a baseline, the variance is high. Furthermore, all trajectories are suppressed because the gradient assigns negative coefficients to all of them as we will show in \Cref{eq:grad}. This convergence issue is discussed in prior work \citep{chung2021beyond,mei2022role}, which demonstrates that the baseline's role extends beyond variance reduction and can determine the convergence point.

We compute the gradient of $J(\theta)$ defined in \Cref{eq:algorithm_GRPO}. We drop the thresholding for simplicity. Define:
\[
u(\theta):= \frac{1}{G}\sum_{i=1}^G \frac{1}{|y_i|}\sum_{t=1}^{|y_i|}\frac{\pi_{\theta}(y_{i,t}|x,y_{i,<t})}{\pi_{\mathrm{old}}(y_{i,t}|x,y_{i,<t})}e^{-A_{i,t}/\lambda}\,.
\]
So,
\begin{align*}
\nabla J(\theta) &= \E_{x\sim p}\!\left[-\lambda \frac{\nabla u(\theta)}{u(\theta)} - \beta \frac{1}{G}\sum_{i=1}^G \frac{1}{|y_i|}\sum_{t=1}^{|y_i|}\left(-\frac{\pi_{\mathrm{ref}}(y_{i,t}|x, y_{i,<t})}{\pi_{\theta}(y_{i,t}|x, y_{i,<t})^2}\nabla \pi_{\theta}(y_{i,t}|x, y_{i,<t}) + \nabla \log \pi_{\theta}(y_{i,t}|x, y_{i,<t}) 
\right) \right]\\
&= \E_{x\sim p}\!\left[-\lambda \frac{\nabla u(\theta)}{u(\theta)} - \beta \frac{1}{G}\sum_{i=1}^G \frac{1}{|y_i|}\sum_{t=1}^{|y_i|}\left(-\frac{\pi_{\mathrm{ref}}(y_{i,t}|x, y_{i,<t})}{\pi_{\theta}(y_{i,t}|x, y_{i,<t})}\nabla \log \pi_{\theta}(y_{i,t}|x, y_{i,<t}) + \nabla \log \pi_{\theta}(y_{i,t}|x, y_{i,<t}) 
\right) \right]
\end{align*}
In addition,
\[
\nabla u(\theta) = \frac{1}{G}\sum_{i=1}^G \frac{1}{|y_i|}\sum_{t=1}^{|y_i|}\frac{\nabla \pi_{\theta}(y_{i,t}|x,y_{i,<t})}{\pi_{\mathrm{old}}(y_{i,t}|x,y_{i,<t})}e^{-A_{i,t}/\lambda} =
\frac{1}{G}\sum_{i=1}^G \frac{1}{|y_i|}\sum_{t=1}^{|y_i|}\frac{\pi_{\theta}(y_{i,t}|x,y_{i,<t})}{\pi_{\mathrm{old}}(y_{i,t}|x,y_{i,<t})}e^{-A_{i,t}/\lambda} \nabla \log\pi_{\theta}(y_{i,t}|x,y_{i,<t})
\]
where we use the log-derivative trick to derive the gradient as in other policy gradient algorithms. Putting it back into $\nabla J(\theta)$, we obtain:
\begin{align*}
\nabla J(\theta) &= \E_{x\sim p}\!\left[
\frac{1}{G}\sum_{i=1}^G \frac{1}{|y_i|}\sum_{t=1}^{|y_i|} \left\{
-\frac{\pi_{\theta}(y_{i,t}|x,y_{i,<t})}{\pi_{\mathrm{old}}(y_{i,t}|x,y_{i,<t})}\frac{\lambda e^{-A_{i,t}/\lambda}}{u(\theta)} + \beta \frac{\pi_{\mathrm{ref}}(y_{i,t}|x, y_{i,<t})}{\pi_{\theta}(y_{i,t}|x, y_{i,<t})} -\beta
\right\}\nabla \log\pi_{\theta}(y_{i,t}|x,y_{i,<t}) 
\right]
\end{align*}
Now, similar to GRPO \citep{shao2024deepseekmath}, we simplify the analysis by assuming that the model only has a single update following each
exploration stage, thereby ensuring that $\pi_{\mathrm{old}} = \pi_{\theta}$.
Doing this, we can write
\[
u(\theta) = \frac{1}{G}\sum_{i=1}^G \frac{1}{|y_i|}\sum_{t=1}^{|y_i|}e^{-A_{i,t}/\lambda}\,,
\]
and
\begin{align}\label{eq:grad}
\nabla J(\theta) &= \E_{x\sim p}\!\left[
\frac{1}{G}\sum_{i=1}^G \frac{1}{|y_i|}\sum_{t=1}^{|y_i|} \left\{
-\frac{\lambda e^{-A_{i,t}/\lambda}}{u(\theta)} + \beta \frac{\pi_{\mathrm{ref}}(y_{i,t}|x, y_{i,<t})}{\pi_{\theta}(y_{i,t}|x, y_{i,<t})} -\beta
\right\}\nabla \log\pi_{\theta}(y_{i,t}|x,y_{i,<t}) 
\right]
\end{align}
The term above shows that \uline{adding a constant to the advantages of a group does not change the gradient}, and it cancels out from the numerator and denominator of the first term in \Cref{eq:grad}. Therefore, rewards can be replaced with advantages in \Cref{eq:algorithm_GRPO}.

\vspace{-0.1in}
\paragraph{Baseline derivation.} Note that if $\lambda \gg 1$, we can apply the Taylor expansion for the terms inside \Cref{eq:grad}:

\begin{align*}
\nabla J(\theta) &\approx \E_{x\sim p}\!\left[
\frac{1}{G}\sum_{i=1}^G \frac{1}{|y_i|}\sum_{t=1}^{|y_i|} \left\{
-\lambda\, \frac{\big (1 - A_{i,t}/\lambda + O(\frac{1}{\lambda^2}) \big)}{1- \langle A_{i, t} \rangle/\lambda + O(\frac{1}{\lambda^2})} + \beta \frac{\pi_{\mathrm{ref}}(y_{i,t}|x, y_{i,<t})}{\pi_{\theta}(y_{i,t}|x, y_{i,<t})} -\beta
\right\}\nabla \log\pi_{\theta}(y_{i,t}|x,y_{i,<t}) 
\right] \\
&= \E_{x\sim p}\!\left[
\frac{1}{G}\sum_{i=1}^G \frac{1}{|y_i|}\sum_{t=1}^{|y_i|} \left\{
 \big ( -\lambda + A_{i,t} + O(\frac{1}{\lambda}) \big) + \beta \frac{\pi_{\mathrm{ref}}(y_{i,t}|x, y_{i,<t})}{\pi_{\theta}(y_{i,t}|x, y_{i,<t})} -\beta
\right\}\nabla \log\pi_{\theta}(y_{i,t}|x,y_{i,<t}) 
\right]
\end{align*}
In the second line, we note that $ \langle A_{i, t} \rangle =0$ because we subtract the group average from each reward, similar to GRPO. Then, the $\lambda$ term can be removed since $\E_{y\sim\pi_\theta(\cdot|x)}\big[\nabla_\theta\log\pi_\theta(y|x)\big]=0$: any baseline $b(x)$ that does not depend on $y$ can be subtracted without changing the expectation of the gradient. This recovers the gradient from GRPO. This term would cause a large variance in cases where $\lambda$ is large. 

\section{Bias of the Partition Function}\label{app:jackknife}

\begin{figure}
    \centering
    \begin{subfigure}{0.4\linewidth}
        \includegraphics[width=\linewidth]{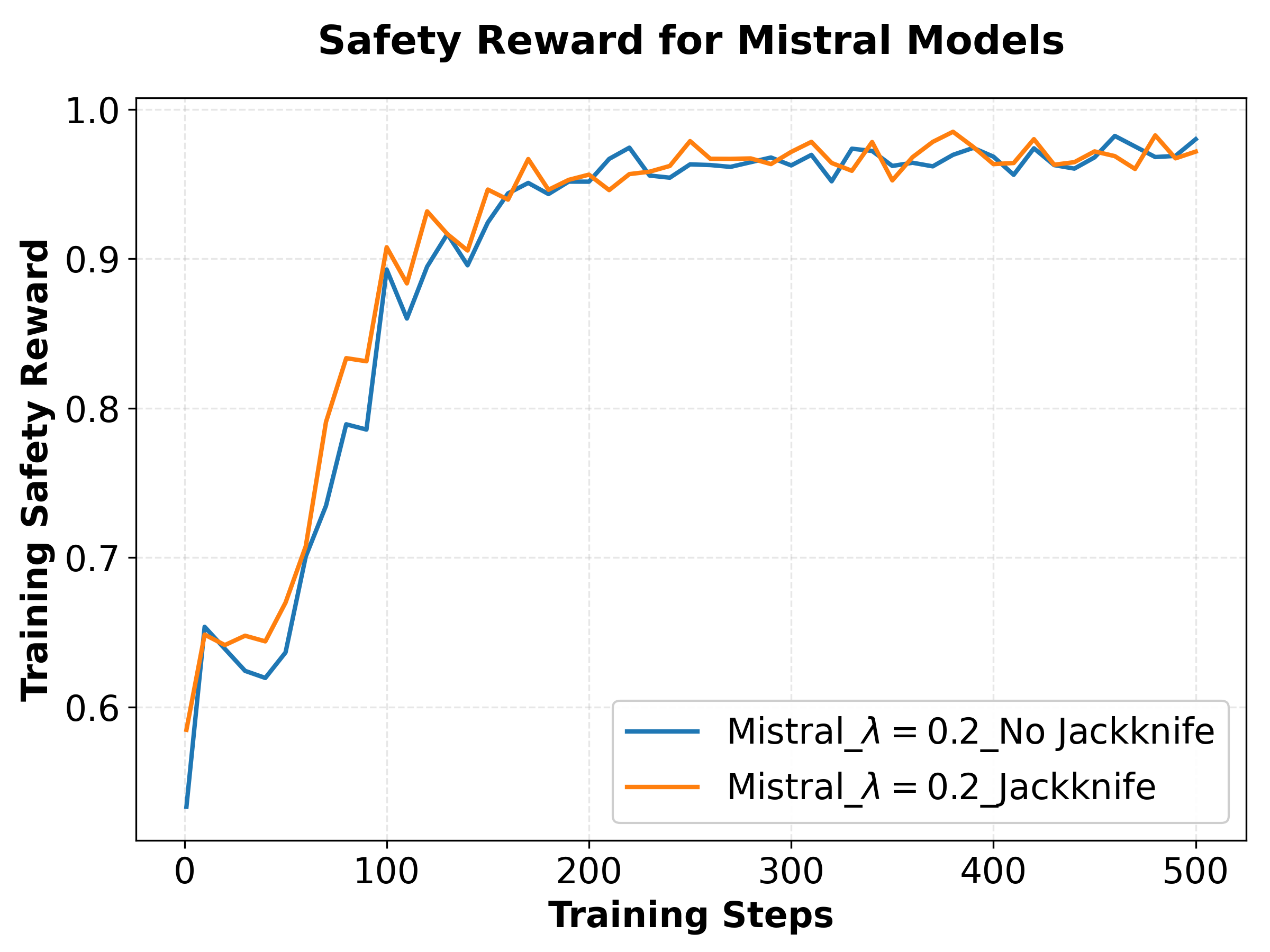}
    \end{subfigure}
    \hspace{0.03 \linewidth}
    \begin{subfigure}{0.4\linewidth}
        \includegraphics[width=\linewidth]{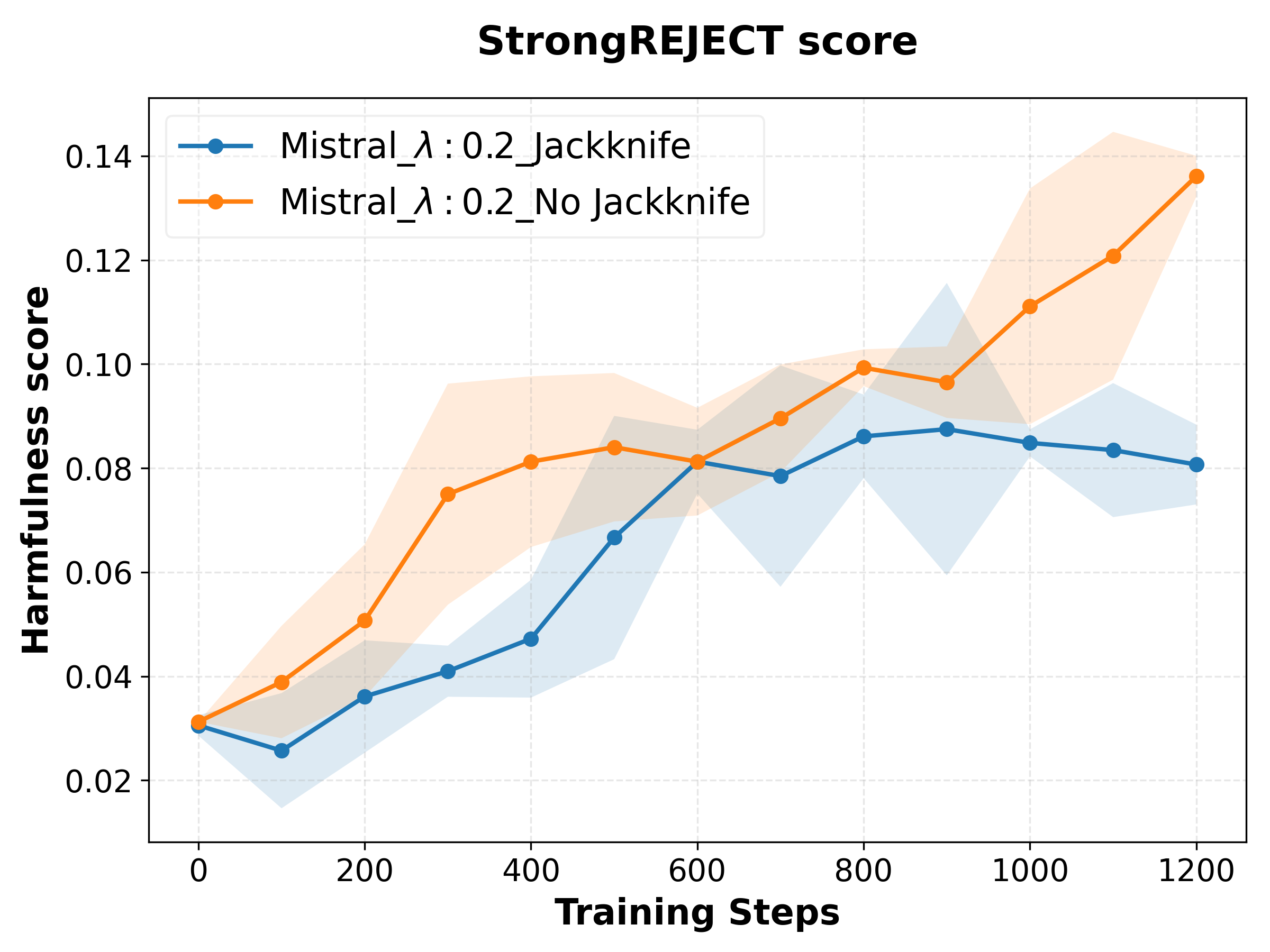}
    \end{subfigure}
     \caption{\textbf{(left)} The bias in the gradient estimator does not show itself in the safety training curves and the training without the jackknife trick looks similar. \textbf{(right)} However, at the downstream time, the model that is trained without the jackknife trick is less robust and the StrongREJECT score ($\downarrow$ is better) grows more compared to the model with the Jackknife trick.} 
     \label{fig:bias}
\end{figure}

Unlike the baseline discussed in \Cref{app:baseline}, the gradient bias arising from the Monte Carlo estimation of the log partition function in \Cref{eq:bias} does not impact the training curves, as shown in \Cref{fig:bias} (left). However, \Cref{fig:bias} (right) reveals that when the model is fine-tuned downstream, the reward drops more sharply and the harmfulness score increases relative to the model trained with the jackknife trick. We now explain why this bias appears and how we address it. Starting from the gradient derived in \Cref{eq:grad}, consider the first term:

\begin{equation}\label{eq:bias}
    \nabla_\theta J = -\E_{x\sim p}\!\left[ \frac{\lambda}{\sum_i e^{-\,A(x,y_{i})/\lambda}} \sum_i e^{-\,A(x,y_{i})/\lambda} \nabla_\theta \log \pi_\theta (y_{i, j} \mid x_i) \right]
\end{equation}
While this estimator converges to the expected value for large $G$, it is a biased estimator. This term is in fact self-normalized importance sampling (SNIS). SNIS is known to have a bias of order $O(1/G)$ \cite[\S~9]{mcbook}. More explicitly, the gradient term has the form of a \emph{ratio estimator}:
\[
\hat g \;=\; \frac{\sum_{i=1}^G w_i\,\phi_i}{\sum_{i=1}^G w_i},
\qquad
w_i := e^{-A_i/\lambda},\;\; \phi_i := \nabla_\theta \log \pi_\theta(y_i\mid x),
\]
which is exactly SNIS for expectations under $Q(y\mid x)\propto \pi_\theta(y\mid x)e^{-A/\lambda}$.
Even if both $\sum_i w_i\phi_i$ and $\sum_i w_i$ are unbiased for their respective expectations, their \emph{ratio} is not: by a second-order Taylor (i.e., delta-method) expansion around $(\E[w\phi],\E[w])$, one obtains a bias term proportional to
$\mathrm{Cov}(w\phi, w)/(\E[w])^2$, yielding the standard $O(1/G)$ bias for SNIS.
This effect is most pronounced when $G$ is small or when the weights $w_i$ are heavy-tailed (which happens for small $\lambda$).

\vspace{-0.1in}
\paragraph{The jackknife technique.} Consider an estimator $\hat{g}(X)$ for the underlying parameter $g$, computed from a sample size $|X| = k$. Let $X_{-i}$ denote the sample vector in which the $i$-th sample is deleted, and $\hat{g}(X_{-i})$ denote the estimator in the absence of the $i$-th sample. Then, the Jackknife estimator is: 
$$\tilde{g}(X)  = k \hat{g}(X) - \frac{k-1}{k} \sum_i \hat{g}(X_{-i}) $$
It can be easily seen that if the bias of the original estimator is $e_{\hat{g}} = c/k + O({1}/{k^2})$, then the jackknife technique reduces this to $e_{\tilde{g}} =  O({1}/{k^2})$ \citep{jacknife,jiao2020bias}. The jackknife targets this bias by using the leave-one-out terms $\hat g(X_{-j})$ to estimate the leading $1/k$ term in the Taylor expansion of the bias, and subtracting it via the linear combination in $\tilde g(X)$. As a result, the $O(1/k)$ term cancels and the remaining bias is $O(1/k^2)$.

